\documentclass[lettersize,journal]{IEEEtran}
\usepackage{amsmath,amsfonts}
\usepackage{algorithmic}
\usepackage{array}
\usepackage[caption=false,font=normalsize,labelfont=sf,textfont=sf]{subfig}
\usepackage{textcomp}
\usepackage{stfloats}
\usepackage{url}
\usepackage{verbatim}
\usepackage{graphicx}
\def\BibTeX{{\rm B\kern-.05em{\sc i\kern-.025em b}\kern-.08em
    T\kern-.1667em\lower.7ex\hbox{E}\kern-.125emX}}
\usepackage{balance}% From ICCV template

\usepackage{times}
\usepackage{epsfig}
\usepackage{graphicx}
\usepackage{amsmath}
\usepackage{amssymb}

\usepackage{booktabs}
\usepackage{tabularx}
\usepackage{multirow}
\usepackage{placeins}
\usepackage{enumitem}
\usepackage{arydshln}
\usepackage[table,xcdraw]{xcolor}
\usepackage{caption}
\usepackage[accsupp]{axessibility}  % Improves PDF readability for those with disabilities.
\DeclareUnicodeCharacter{1EA5}{\^a}
\DeclareUnicodeCharacter{1EA7}{\^a\`{}}
\usepackage[pagebackref=true,breaklinks=true,letterpaper=true,colorlinks,bookmarks=false]{hyperref}

\usepackage[capitalize]{cleveref}
\usepackage{adjustbox}
\crefname{section}{Sec.}{Secs.}
\Crefname{section}{Section}{Sections}
\Crefname{table}{Table}{Tables}
\crefname{table}{Tab.}{Tabs.}

\newcommand{\vect}[1]{\ensuremath{\mathbf{#1}}}

\newcommand{\revised}[1]{{\color{magenta}#1}}

\newcommand{\xT}{\vect{x}_T}
\newcommand{\xt}{\vect{x}_t}
\newcommand{\xzero}{\vect{x}_0}
\newcommand{\xtone}{\vect{x}_{t-1}}

\usepackage{amsmath,amsfonts}
\usepackage{algorithmic}
\usepackage{array}
\usepackage[caption=false,font=normalsize,labelfont=sf,textfont=sf]{subfig}
\usepackage{textcomp}
\usepackage{stfloats}
\usepackage{url}
\usepackage{verbatim}
\usepackage{graphicx}

\begin{document}
      % index file is design for include
  % in case we have to change templete again, just \input this file
  %%%%%%%%% TITLE - PLEASE UPDATE
\title{DiFaReli++: Diffusion Face Relighting with Consistent Cast Shadows}

\author{Puntawat Ponglertnapakorn, Nontawat Tritrong, and Supasorn Suwajanakorn\textsuperscript{*}
\thanks{\textsuperscript{*}Corresponding author.}
\thanks{P. Ponglertnapakorn, N. Tritrong, and S. Suwajanakorn are with the School of Information Science and Technology, Vidyasirimedhi Institute of Science and Technology, Thailand. E-mail: \{puntawat.p\_s19, nontawat.t\_s19,  supasorn.s\}@vistec.ac.th}
}
% \markboth{Journal of \LaTeX\ Class Files,~Vol.~18, No.~9, September~2020}%
% \markboth{Under review}%
{}

\iffalse
\maketitle
\begin{figure*} % Force figure to the top of the page
\centering
% \includegraphics[scale=0.1]{./figures/Teaser.pdf} % Adjust scale 
\includegraphics[scale=0.1]{./figures/Teaser_new.pdf} 
% \includegraphics[scale=0.1]{./figures/ph.pdf} % Adjust scale as necessary

\caption{\re\textbf{Relighting with controllable cast shadows.} Our method can realistically relight an input image under different target lighting conditions and also control cast shadows. In Relit 1 and Relit 2, our method generates new cast shadow effects, with the intensity controlled by strengthening or attenuating the shadows (as shown in the small side images). Additionally, on the right, we present examples of controllable cast shadows, where a single input image (in the middle) is relit under 8 different target lighting conditions, producing 8 distinct cast shadows that align with the lighting direction and face geometry.}
\label{fig-intro}
\end{figure*}
\fi

% \iffalse
\twocolumn[{%
\renewcommand\twocolumn[1][]{#1}%
\maketitle
\vspace{-1.2cm}
\begin{center}
\centering
  \includegraphics[scale=0.405]{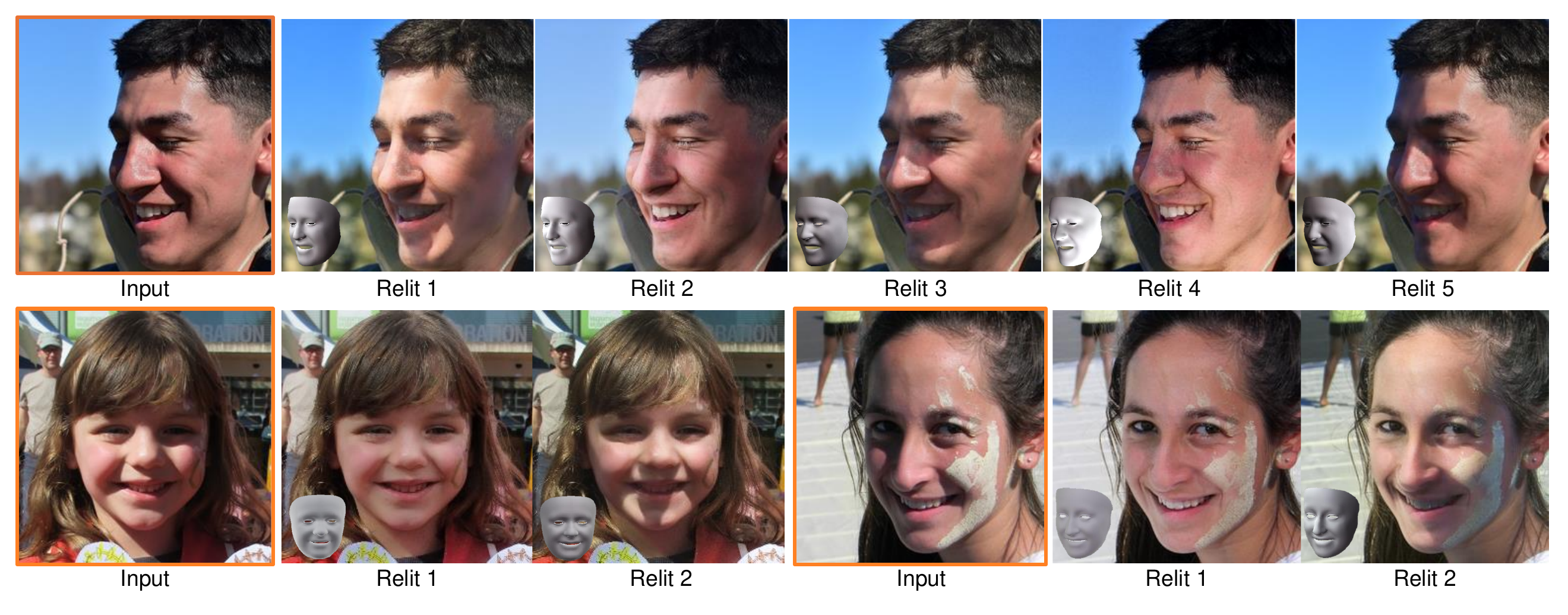}

  \vspace{-0.1cm}
  \captionof{figure}{
  %\textbf{Relighting with consistent cast shadows.} 
%Our method realistically relights input images with extreme head poses (top) and faces with makeup (bottom) under various target lighting conditions. It effectively addresses the hardest cases—removing strong input highlights and cast shadows while generating temporally consistent new ones—all in a single network pass, given pre-processed input data.
Our method addresses one of the most challenging relighting scenarios where input images contain strong highlights and cast shadows. It effectively removes these effects and generates convincing shading and temporally consistent new shadows---all in a single network pass, given pre-processed input data. It also works across varying head poses, identities, and facial makeup.
%It also works across head poses (top) and faces with makeup (bottom) under various lighting conditions.
  % In the first row (left), we compare our method to DiFaReli, which generates plausible cast shadows, whereas DiFaReli++ allows for freely controlling them. In the other rows, we present additional results demonstrating that our method can relight images and generate new cast shadows. Furthermore, our method (right) effectively handles extreme cast shadows covering the face, producing promising relighting results (input shown in the middle).
  }
  % In Relit 1 and Relit 2, our method generates new cast shadow effects, with the intensity controlled by strengthening or attenuating the shadows (as shown in the small side images). Additionally, on the right, we present examples of controllable cast shadows, where a single input image (in the middle) is relit under 8 different target lighting conditions, producing 8 distinct cast shadows that align with the lighting direction and face geometry.

  \label{fig:fig-intro}
\end{center}
}]

\setcounter{footnote}{1}
{\renewcommand*{\thefootnote}{\fnsymbol{footnote}}\stepcounter{footnote}%
  \footnotetext{P. Ponglertnapakorn, N. Tritrong, and S. Suwajanakorn are with the School of Information Science and Technology, Vidyasirimedhi Institute of Science and Technology, Thailand. E-mail: \{puntawat.p\_s19, nontawat.t\_s19,  supasorn.s\}@vistec.ac.th}}
\setcounter{footnote}{0}
% {\renewcommand*{\thefootnote}{\fnsymbol{footnote}}\stepcounter{footnote}%
  % \footnotetext{}}
  
\setcounter{footnote}{2}
{\renewcommand*{\thefootnote}{\fnsymbol{footnote}}\stepcounter{footnote}%
  \footnotetext{Published in IEEE Transactions on Pattern Analysis and Machine Intelligence (TPAMI), vol. 48, pp. 5068-5082, May 2026. DOI: 10.1109/TPAMI.2025.3648667}}
\setcounter{footnote}{0}
  
{\renewcommand*{\thefootnote}{\fnsymbol{footnote}}\stepcounter{footnote}%
  \footnotetext{Corresponding author}}
\setcounter{footnote}{0}

% Text\footnote{This works.}

%%%%%%%%% ABSTRACT (ICCV version)
\iffalse
\begin{abstract}
We present a novel approach to single-view face relighting in the wild. 
Handling non-diffuse effects, such as global illumination or cast shadows, has long been a challenge in face relighting.
Prior work often assumes Lambertian surfaces, simplified lighting models or involves estimating 3D shape, albedo, or a shadow map. This estimation, however, is error-prone and requires many training examples with lighting ground truth to generalize well.
Our work bypasses the need for accurate estimation of intrinsic components and can be trained solely on 2D images without any light stage data, multi-view images, or lighting ground truth. 
Our key idea is to leverage a conditional diffusion implicit model (DDIM) for decoding a disentangled light encoding along with other encodings related to 3D shape and facial identity inferred from off-the-shelf estimators.
We also propose a novel conditioning technique that eases the modeling of the complex interaction between light and geometry by using a rendered shading reference to spatially modulate the DDIM.
We achieve state-of-the-art performance on standard benchmark Multi-PIE and can photorealistically relight in-the-wild images. Please visit our page: \textit{\textcolor{pink}{\footnotesize \url{https://diffusion-face-relighting.github.io}}}.
\end{abstract}
\fi

%%%%%%%%% ABSTRACT (TPAMI version)
\begin{abstract}
We introduce a novel approach to single-view face relighting in the wild, addressing challenges such as global illumination and cast shadows.
A common scheme in recent methods involves intrinsically decomposing an input image into 3D shape, albedo, and lighting, then recomposing it with the target lighting.
However, estimating these components is error-prone and requires many training examples with ground-truth lighting to generalize well.
Our work bypasses the need for accurate intrinsic estimation and can be trained solely on 2D images without any light stage data, relit pairs, multi-view images, or lighting ground truth. 
Our key idea is to leverage a conditional diffusion implicit model (DDIM) for decoding a disentangled light encoding along with other encodings related to 3D shape and facial identity inferred from off-the-shelf estimators.
We propose a novel conditioning technique that simplifies modeling the complex interaction between light and geometry. It uses a rendered shading reference along with a shadow map, inferred using a simple and effective technique, to spatially modulate the DDIM. 
Moreover, we propose a single-shot relighting framework that requires just one network pass, given pre-processed data, and even outperforms the teacher model across all metrics.
Our method realistically relights in-the-wild images with temporally consistent cast shadows under varying lighting conditions. We achieve state-of-the-art performance on the standard benchmark Multi-PIE and rank highest in user studies.
Please visit our page: \textit{\textcolor{pink}{\footnotesize \url{https://diffusion-face-relighting-pp.github.io}}}

%We further improve upon DiFaReli by enabling consistent cast shadow generation through shadow maps estimated with a novel technique from 2D images.
%computed solely from a 2D image under our proposed scheme, without the need for light stage data, multi-view images, or precise light-geometry estimation.
% Additionally, we propose an alternative way to compute shadow maps solely from 2D images, which can be used to condition the network and give users control over cast shadow effects without requiring light stage data, multi-view images, or suffering from inaccuracies in light and face geometry estimation.
%Furthermore, we propose a single-shot framework for training a relighting network that requires only one network pass, significantly boosting speed while maintaining image quality.
%We also demonstrate that a simple modification to the background conditioning enables the relighting of other parts of the person such as clothes and hats.
%Our method can realistically relight in-the-wild images with temporally consistent cast shadows as the lighting changes. We achieve state-of-the-art performance on the standard benchmark Multi-PIE and are top-rated according to a user study. 
% Please visit our page: \textit{\textcolor{pink}{\footnotesize \url{https://diffusion-face-relighting.github.io}}}.
\end{abstract}


  \begin{figure*}
\centering
  \includegraphics[scale=0.45]{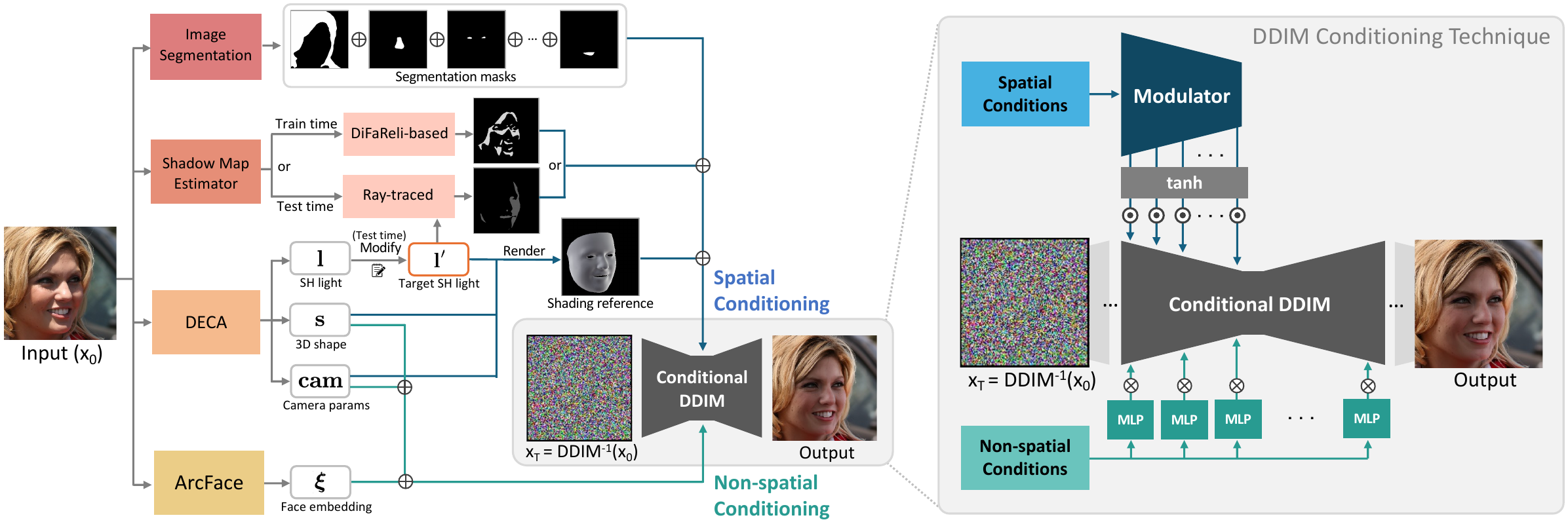}

  % \vspace{-0.5cm}
  \caption{\textbf{Overview of DiFaReli++.} We use off-the-shelf estimators to derive various encodings from the input image: segmentation masks, shadow map, (light, shape, camera) parameters, and face embedding. These encodings are then fed into a conditional DDIM via ``spatial''  and ``non-spatial'' conditioning techniques. 
  For spatial conditioning, a shading reference, shadow map, and segmentation masks are concatenated and fed into the Modulator to produce spatial modulation weights for the first half of the DDIM. Meanwhile, the 3D shape, camera, and face embedding are concatenated and processed by a set of MLPs, which modulate the DDIM using a modified version of adaptive group normalization (AdaGN). For DiFaReli, please see Figure 18.
  % \ref{fig:pipelinecompare}.
  %For spatial conditioning, we concatenate a shading reference, a shadow map, and segmentation masks together and feed it to the Modulator to produce spatial modulation weights for the DDIM's first half. The 3D shape, camera, and face embedding are concatenated and fed to a set of MLPs for modulating the DDIM with our modified version of adaptive group normalization (AdaGN).
  }
  \label{pipeline_overview}
  \vspace{-1.5em}
\end{figure*}

\section{Introduction}
%Face relighting is a long-standing problem 
%The ability to relight face images under any lighting condition has a wide range of applications, including in portrait photography, where one may seek to soften cast shadows for a more pleasing, diffuse appearance. Another is in Augmented Reality, where the lighting should be made consistent for every person in the scene to achieve realism. Yet, many relighting challenges remain unsolved.
\vspace{-0.1em}
% \IEEEPARstart{T}{he} ability to relight face images under any lighting condition has a wide range of applications, such as in Augmented Reality, where consistent lighting for all individuals in the scene is essential to achieve realism. Another use is in portrait photography, where one may aim to soften cast shadows to create a more pleasing, diffuse appearance. Yet, relighting single-view face images remains unsolved.
\IEEEPARstart{R}{elighting} face images under various lighting conditions is a key aspect of portrait photography and artistic image editing. It involves adjusting facial shading to new lighting while accounting for global illumination and subsurface scattering effects that produce soft lighting transitions and a realistic skin appearance, as well as self-occlusion for hard cast shadows.
%, which create softer lighting transitions and a realistic skin appearance, as well as self-occlusion, which produces hard cast shadows.
%appearance of the face to corres
%This ability includes adjusting light direction and intensity, along with manipulating complex lighting effects, such as strengthening or diffusing cast shadows, or even adding new ones to enhance the visual mood. 
These capabilities enable applications such as digital characters that adapt to virtual lighting, cinematic effects for filmmakers, and virtual try-on. However, achieving such realism and control in single-view face relighting is challenging, as it requires modeling light interactions with facial geometry and materials from just one image.
Earlier work \cite{barron2014shape, sengupta2018sfsnet, le2019illumination, wang2008face, shu2017neural} often assumes Lambertian surfaces and simplified lighting models, which struggle to capture complex light interactions, such as indirect lighting or cast shadows. Using multi-view, multi-illumination data from a light stage or a simulation, \cite{nestmeyer2020learning, hou2021towards, pandey2021total, kim2024switchlight, yeh2022learning, mei2024holo} proposed relighting pipelines that predict surface normals, albedo, and a set of diffuse and specular maps with neural networks given a target HDR map. 
% These approaches share a common scheme: they first decompose the face image into surface normals, albedo, and lighting parameters, then use these along with a shadow or visibility map to render the relit output. 
% While these methods have made progress in face relighting, which can handle complex lighting to some extent, challenges remain in effectively handling and controlling cast shadows. Although some pipelines can produce plausible cast shadows by implicitly inferring them from an HDR map or background, they are not specifically designed to directly control this effect.
% These method shows improvement handle complex lighting to some extent and produce plausible cast shadows
% While these methods have advanced face relighting research and can handle complex lighting to some extent, challenges remain in achieving photorealistic results and effectively controlling cast shadows.
% These demonstrate progress in handling complex lighting and producing plausible cast shadows. However, challenges remain, particularly in achieving photorealistic results while offering control over cast shadow placement. Since these pipelines are not designed for direct control of cast shadows, recent methods aim to address these limitations by modeling cast shadows explicitly, either by predicting a shadow map using a neural network \cite{hou2021towards, nestmeyer2020learning} or by rendering a shadow map through physical ray tracing with estimated geometry \cite{hou2022face}.
Recent methods explicitly model cast shadows through shadow maps, either by predicting them using a neural network \cite{hou2021towards, nestmeyer2020learning} or by rendering with physical ray tracing with estimated geometry \cite{hou2022face}.
% These methods demonstrate progress in handling complex lighting and producing plausible cast shadows \cite{pandey2021total, yeh2022learning}. 
% Recent methods have further focused on hard cast shadows and improved their consistency by explicitly modeling them through shadow maps, either by predicting a shadow map using a neural network \cite{hou2021towards, nestmeyer2020learning} or by rendering them through physical ray tracing with estimated geometry \cite{hou2022face}.

% Despite improvements, these methods still struggle with photorealistic relighting in challenging in-the-wild scenarios, such as removing existing hard cast shadows and strong highlights from the input image while realistically adding new ones. In such cases, state-of-the-art methods often fail to fully remove shadows or highlights, leaving residual artifacts in the output (Figure \ref{fig:ffhq_cs}) or produce less convincing results.
Despite progress, state-of-the-art methods still struggle with photorealistic relighting in challenging in-the-wild scenarios, often failing to remove or add strong highlights and cast shadows realistically, leading to residual artifacts (Figure \ref{fig:ffhq_cs}) or less convincing results.

%For example, when an input image contains cast shadows that need to be removed, some shadows often remain in the predicted albedo map, leading to unrealistic results (Figure \ref{fig:ffhq_cs}). Furthermore, this failure mode makes the relit outcomes appear even less realistic when new cast shadow effects are added.

% Despite these improvements, a key challenge remains in effectively handling inputs with cast shadows. Most prior works struggle to remove cast shadows from input images, often leaving residual shadows in the predicted albedo map, which results in unrealistic final outputs (Figure \ref{fig:ffhq_cs}). Furthermore, this limitation makes the relit outcomes appear even less realistic when new cast shadows are added.

These issues arise from a common strategy in these pipelines: decomposing the face into surface normals, albedo, and lighting parameters to render the relit output. 
%This process depends heavily on the accuracy of the estimated components, which are challenging to compute correctly in real-world scenarios and directly affect the realism of the final output. 
This process heavily depends on the accuracy of these estimated components, which is challenging to achieve in real-world scenarios and directly impacts the realism of the final result. 
%Moreover, estimating the geometry for other areas like hair and ears is also extremely challenging, and they are often omitted from relighting pipelines, resulting in unrealistic final composites (Figure \ref{fig:ffhq_cs} and \ref{fig:mutlipie_cs}).
Additionally, geometry estimation for other areas like hair and ears is extremely difficult and often omitted from relighting pipelines, leading to unrealistic composites (Figure \ref{fig:ffhq_cs} and \ref{fig:mutlipie_cs}).

This paper introduces an alternative approach that does not rely on accurate intrinsic face decomposition and is trained solely on 2D images---without 3D scans, multi-view images, relit pairs, or lighting ground truth---once given a few off-the-shelf estimators.
Our general idea is simple: we first encode the input image into a feature vector that disentangles lighting from other information about the input image. Then, we modify the light encoding and decode it. 
%The challenge, however, is how to disentangle the light encoding well enough so that the decoding will only affect the shading without altering the person's shape and identity. 
The challenge lies in disentangling lighting so that only shading changes while shape and identity remain intact.
Our key idea is to leverage a conditional diffusion implicit model \cite{song2021denoising} with a novel conditioning technique to learn complex light interactions implicitly via the generative model trained on real-world 2D face data.

% \revised{The working principle of our method relies on mechanisms recently introduced in Denoising Diffusion Implicit Models (DDIM) \cite{song2021denoising} and Diffusion Autoencoders (DiffAE) \cite{preechakul2022diffusion}, which are closely related to DDPMs \cite{pmlr-v37-sohl-dickstein15, ho2020denoising, NEURIPS2019_3001ef25}}. 
Our method builds on mechanisms from Denoising Diffusion Implicit Models (DDIM) \cite{song2021denoising} and Diffusion Autoencoders (DiffAE) \cite{preechakul2022diffusion}.
%By exploiting the deterministic reversal process of DDIM,
% proposed by Song et al.~\cite{song2021denoising}
Leveraging DDIM's deterministic reversal, 
DiffAE encodes an image into a meaningful semantic code disentangled from other information.
%DiffAE shows how one can encode an image into a meaningful semantic code and disentangle it from other information.
% , which includes stochastic variations. 
By modifying and decoding the semantic code, DiffAE can manipulate attributes in real images. Relighting can be viewed as manipulating the ``light'' attribute. But unlike DiffAE, which discovers semantic attributes automatically and encodes them in a \emph{latent} code, our method requires an explicit and interpretable light encoding that facilitates lighting manipulation by the user.
%semantic attributes are discovered automatically and encoded in a latent, non-human-readable code, our method requires explicit, meaningful, and well-disentangled light encoding that facilitates precise lighting manipulation.

% \revised{To solve this, we encode the light information explicitly into spherical harmonic (SH) coefficients using an off-the-shelf estimator, DECA \cite{DECA:Siggraph2021}}, 
To solve this without lighting ground truth, we use an off-the-shelf estimator, DECA \cite{DECA:Siggraph2021}, to encode lighting information as spherical harmonic (SH) coefficients and employ a conditional DDIM to decode and disentangle the light information in the process. 
%To solve this, we use an off-the-shelf estimator, DECA \cite{DECA:Siggraph2021}, to encode the lighting information as spherical harmonic (SH) coefficients and rely on a conditional DDIM to decode and learn to disentangle the light information in the process. 
%Unlike prior work, our use of SH lighting is not for direct rendering of the output shading, as the result would be restricted by the limited capacity of SH lighting to express complex illumination. Rather, it is used to condition a generative process that learns the complex shading prior to reproduce real-world 2D face images.
Unlike prior work, we do not use SH lighting to directly render the output, which is limited in capturing complex illumination. Instead, we use it to condition a generative process that learns complex shading priors to reproduce real-world 2D face images.
%Another practical reason for using SH lighting is the availability of SH light estimator from face images.
%Note that more complex lighting models can be used in this framework, but estimating such complex models from a single-view face image
%As we aim to learn purely from 2D data without lighting ground truth,
%We also choose SH for a practical reason that there exists a reliable SH estimator from a single-view face image. 
To help preserve identity during relighting, we also condition the DDIM on attributes such as face shape and Arcface embeddings~\cite{deng2019arcface}. 

%Another key component is our novel technique for conditioning the DDIM. Instead of flattening the light or shape parameters and use them as non-spatial condition vectors as in DiffAE or other diffusion models, we render a shading reference using the known SH equation and feed it to another network called \emph{Modulator}, which computes layer-wise spatial modulation weights for the DDIM. This conditioning scheme helps retain spatial information in the shading reference and provides an easy-to-learn conditioning signal as the pixel intensities in the shading reference correlate directly with the output RGB pixels. 
% Another key component is our novel technique for conditioning the DDIM. Instead of treating the SH lighting as a global, non-spatial condition vector as in DiffAE or other diffusion models, we render a shading reference using the known SH equation and feed it to another network called \emph{Modulator}, which computes layer-wise spatial modulation weights for the DDIM. This conditioning technique helps retain spatial information in the shading reference and provides an easy-to-learn conditioning signal as the pixel intensities in the shading reference correlate more directly with the output RGB pixels. 

Another key component is our conditioning technique for DDIM. Instead of using SH lighting as a global conditioning vector as in DiffAE, we render a shading reference via the known SH equation and feed it to a \emph{Modulator} network, which computes layer-wise spatial modulation weights for DDIM. 
This scheme helps preserve spatial information in the shading reference and provides an easy-to-learn signal, since its pixel intensities correlate directly with the output RGB values. 
Note that a similar spatial modulation idea, though not designed for relighting, was proposed in ControlNet \cite{zhang2023adding}, released concurrently with our conference version DiFaReli~\cite{ponglertnapakorn2023difareli}. 
%\revised{
%Our conditioning technique is similar to ControlNet \cite{zhang2023adding}, which was released concurrently during the same period as our work on DiFaReli \cite{ponglertnapakorn2023difareli}. This technique uses a spatially aligned conditioning signal to retain spatial information. By doing so, it provides an easy-to-learn conditioning signal as the pixel intensities in the shading reference correlate directly with the output RGB pixels.}
% \revised{refer to ControlNet}
% \cite{}
% Another key component is our novel technique for modeling the spatial relation between light and face shape. In addition to using the flattened light or shape parameters as non-spatial condition vectors as in DiffAE or other diffusion models, we render a shading reference using the known SH equation and feed it to another network called \emph{Modulator}, which computes layer-wise spatial modulation weights for the DDIM. This conditioning scheme helps retain spatial information in the shading reference and provides an easy-to-learn conditioning signal as the pixel intensities in the shading reference correlate directly with the output RGB pixels. 

In DiFaReli, cast shadows are controlled by a single scalar for ``intensity,'' allowing adjustment of hard shadows (Fig.~\ref{fig:cast_shadow}). However, these shadows may not always appear physically plausible as target lighting changes.
%In DiFaReli, cast shadows are modeled using a single conditioning scalar that represents their ``intensity,'' allowing for the strengthening or attenuation of hard cast shadows (Figure \ref{fig:cast_shadow}). However, DiFaReli's cast shadows may not always appear physically plausible when the target lights move around. 
% For instance, as the sun moves overhead, the hard shadows cast by the nose should sweep naturally across the face rather than fading in and out toward the target direction, as DiFaReli sometimes produces. 
For instance, as the sun moves overhead, the shadows cast by the nose should sweep naturally across the face rather than simply fading in and out, as DiFaReli sometimes produces.
%not sweep across the face but instead fade in and out toward the target direction.
%At test time, we can strengthen or attenuate cast shadows by modifying this scalar (Figure \ref{fig:cast_shadow}).  
% However, the appearance of cast shadows is not consistent, especially under moving lighting. For example, when the light moves from left to right, the shadows sometimes fade in and out inconsistently instead of sweeping smoothly as the light passes over the face, creating shadows under the nose or cheek. This effect is clearly observed when we move the light around, as illustrated in Figure \ref{fig:fig-intro},  \ref{fig:rotate_cs}, and the supplementary video.
%enables the generation of cast shadows that adapt to the target light direction in a temporally consistent and physically plausible way under moving lights.
%we augment the shadow conditioning signal to generate cast shadows that consistently adapt to the target light direction.
%guide the generation of consistent cast shadows that can change. 

Our extended DiFaReli++ addresses this by augmenting the shadow conditioning signal with information capturing the exact shape of cast shadows and using it to condition our DDIM. Unlike prior work ~\cite{hou2021towards, hou2022face}, which estimate shadow maps with error-prone lighting and geometry estimates~\cite{hou2022face}, often causing DDIM to disregard the conditioning signal entirely, we propose a simple yet effective method: use DiFaReli to generate stronger and weaker cast-shadow versions of the input image, then compute their pixel difference (Fig.~\ref{fig:compute_shadow_map}).
At test time, we use ray tracing solely to generate a new conditioning shadow map, which guides the network in synthesizing the final composite (Figure \ref{fig:rotate_cs}).

Additionally, we expand DeFaReli's relightable area to include clothing and hats by relaxing our previous hard constraints. Specifically, we use segmentation masks inferred from an off-the-shelf segmentation network~\cite{yu2018bisenet} instead of raw background pixels to condition the DDIM, enabling relighting while preserving scene and facial structures.
As our framework is based on diffusion models, it is slow due to the multiple steps required for image denoising and inversion. To address this, we propose a single-shot framework that includes dataset generation and training of a single-shot relighting network. Using the same architecture as DiFaReli++, the single-shot network can be trained with a simple supervised L2 loss, without requiring complicated distillation losses or training~\cite{sauer2023adversarial, xu2023ufogen, song2023consistency}.

%Our well-designed architecture facilitates simple and effective single-shot performance. This framework relies on a simple L2 supervised loss rather than any complex distillation loss.
%It convincingly generates new cast shadows and can strengthen or attenuate them as needed. Although our method is designed for single-image relighting, the generated cast shadows are temporally consistent under moving lights without relying on temporal constraints or previous frames as input.

We conduct qualitative and quantitative evaluations, along with user studies. Our method achieves state-of-the-art performance on the standard Multi-PIE benchmark~\cite{gross2010multi} and receives top ratings across all scenarios. Our method produces highly plausible and photorealistic results with consistent cast shadows.
%Moreover, it reproduces the original facial details with high fidelity and avoids leaving cast shadow residuals in the relit images, unlike competing methods that predict albedo maps.
Compared to recent work, our method better preserves details 
% (Figure \ref{fig_app:holo_res}) 
(Figure 19)
and handles inputs with cast shadows more effectively 
(Figure 20)
% (Figure \ref{fig_app:switch_res}).
Surprisingly, our distilled single-shot framework even outperforms the teacher model across all visual metrics and achieves a $1,000\times$ speedup once the input has been pre-processed (Figures \ref{fig:speedup} and \ref{fig:runtime}).
%with its 1,000$\times$ speedup even outperforms the teacher model DiFaReli across all metrics (see Figures \ref{fig:speedup} and \ref{fig:runtime}).
%not only matches the capabilities of the original DiFaReli++ but also outperforms it across all metrics, with a 1,000$\times$ speedup (see Figures \ref{fig:speedup} and \ref{fig:runtime}).
%, in a single network pass while maintaining the quality of the relit images
%Our ability to learn from unconstrained 2D face data also 
%including UNets designed to preserve high-frequency information. 

% Our method produces highly plausible and photorealistic results and can convincingly strengthen or attenuate cast shadows. Moreover, we can reproduce the original facial details with high fidelity, which is difficult for competing methods that predict an albedo map with neural networks. 
%Our ability to learn from unconstrained 2D face data also 
%including UNets designed to preserve high-frequency information. 
% \revised{This approach leads to a 100$\times$ faster inference while still preserving the quality of the relit images (see Figures \ref{fig:speedup} and \ref{fig:runtime}).}

To summarize, this paper introduces two key contributions, which are presented in its conference version \cite{ponglertnapakorn2023difareli}:

\begin{itemize}
    \item A state-of-the-art face relighting framework based on a conditional DDIM that produces photorealistic shading without requiring accurate intrinsic decomposition or 3D and lighting ground truth.
    \item A novel conditioning technique that converts a shading reference rendered from the estimated light and shape parameters into layer-wise spatial modulation weights.
\end{itemize}   

Extending the conference version, this paper introduces DiFaReli++ with the following additions:

\begin{itemize}
    % \item Single-shot inference face relighting framework (Section \ref{single_shot_inf}): We show that our framework can be used to generate high fidelity and quality training dataset. This can be used to train another single shot face relighting network that need only single inference per image and reduce the running time up to $100\times$ faster. We also conduct an experiment to show that not only the dataset, but our designed architecture is essential and necessary to acheive the best performance with least sacrifice the quality.
    % \item An additional conditioning technique (Section \ref{sec:cs}) to expand DiFaReli's relightable area to include non-facial parts such as clothes or hats.
    \item \emph{Relighting with consistent cast shadows (Section \ref{sec:cs}):} We enhance cast shadow consistency and realism under dynamic lighting by utilizing a conditioning shadow map inferred using a novel technique based on DiFaReli.
    % that specifies the cast shadow position. 
    % As opposed to using ray tracing to compute the shadow map for training, we proposed a different method utilizing DiFaReli, which avoids the noise conditions stemming from inaccurate light estimation or geometry. 
    %Our method produces highly photorealistic results with temporally consistent cast shadow movement.
    % \item An additional conditioning technique (Section \ref{sec:relit_bg}) to expand DiFaReli's relightable area to include non-facial parts such as clothes or hats.   
    
    \item \emph{Expanding relightable area (Section \ref{sec:relit_bg}):} We replace raw-pixel background conditioning in DiFaReli with segmentation masks, expanding relightable regions to non-facial parts like clothing and hats.
   % We modify DiFaReli's background conditioning and use segmentation masks instead of raw pixels to make conditioning more flexible and expand the relightable area to non-facial parts like clothing and hats.
    % \item A single-shot relighting framework (Section \ref{single_shot_inf}) that achieves a significant speedup from diffusion-based DiFaReli with minimal quality degradation.
    \item \emph{A single-shot relighting framework (Section \ref{single_shot_inf}):} A simple distillation technique that improves quality over the diffusion-based DiFaReli++ and significantly speeds up inference given all pre-processed data.
    \item \emph{User studies and concurrent work comparison (Section \ref{sec:user_study}, \ref{sec:LP}, \ref{sec:qual_eval} and 
    VI-A
    % \ref{sec_app:compare_more_hdr})
    :} 
    %We conducted user studies and received top ratings across all scenarios. These include relighting quality for facial and non-facial parts and consistent cast shadow movement. 
    We conducted additional experiments, user studies to evaluate relighting quality for facial / non-facial parts and cast shadow consistency, and comparisons with concurrent work (DiffusionRig \cite{ding2023diffusionrig}, IC-Light \cite{iclight}, HoloRelighting \cite{mei2024holo}, SwitchLight \cite{kim2024switchlight}, Neural Gaffer \cite{jin2024neural_gaffer}, and Guo et al. \cite{Guo_2025_CVPR}).
\end{itemize}

   \section{Related work}
%\textbf{Face Relighting}
A common approach to face relighting~\cite{barron2014shape, tewari2021monocular, le2019illumination, sengupta2018sfsnet, wang2008face, shu2017neural} is to decompose an input image into intrinsic components (e.g., lighting, albedo, surface normals) and recompose it back with modified light-related components. Decomposition can be done by regularized optimization \cite{barron2014shape}, morphable model fitting~\cite{blanz1999morphable, wang2008face}, or neural prediction~\cite{sengupta2018sfsnet,le2019illumination,tewari2021monocular,nestmeyer2020learning,wang2020single,pandey2021total, shu2017neural, Paraperas_2023_ICCV, ding2023diffusionrig, futschik2023controllable, ren2023relightful, kim2024switchlight, Guo_2025_CVPR}. Most earlier methods \cite{barron2014shape, sengupta2018sfsnet, le2019illumination, wang2008face, shu2017neural} assume Lambertian surfaces, simplified lighting (e.g., second-order spherical harmonics), and a physical image formation, making them unable to handle real-world effects such as specular highlights or cast shadows.
%non-diffuse effects, such as specular highlights or cast shadows, commonly occur in real-world scenarios.

 Instead of physical decomposition, some methods~\cite{zhou2019deep, sun2019single} use encoder-decoder networks with a latent lighting representation. Zhou et al. \cite{zhou2019deep} constrain this code to predict SH lighting and train another regressor that maps the SH lighting of a reference image to a latent code for relighting.
 Sun et al. \cite{sun2019single} use low-resolution illumination maps, obtained from a light stage, e.g., \cite{wenger2005performance}, instead of the SH lighting. In principle, these models can learn hard shadows and specularities with enough data, but in practice they struggle due to limited light stage data~\cite{sun2019single} or limited variations in synthetic datasets~\cite{zhou2019deep}.
% In contrast, our framework trains directly on diverse, readily available 2D face images.
 %can be trained on 2D face images, which are cheaply available and cover far more diverse scenarios.

\textbf{Handling non-diffuse components.} 
%Relighting non-diffuse components has long been a challenge.
Nestmeyer et al.~\cite{nestmeyer2020learning} use a two-stage framework, predicting non-diffuse components as residuals over a diffuse rendering and cast shadows via a visibility map.
%Nestmeyer et al. \cite{nestmeyer2020learning} use a two-stage framework to predict non-diffuse components as a residual correction of a diffuse rendering from their first stage. Cast shadows are predicted separately as a visibility map, which is multiplied to the output.
Wang et al.~\cite{wang2020single} instead use intrinsic decomposition to learn shadow and specular maps from a large-scale relighting dataset.
%Wang et al. \cite{wang2020single} propose a technique based on intrinsic decomposition that predicts shadow and specular maps by learning from their own large-scale relighting dataset.
Pandey et al.~\cite{pandey2021total} predict specular maps with varying Phong exponents from estimated normals and an HDR environment map. Combined with diffuse and albedo maps, these are used in a UNet to generate a relit image.
%Pandey et al. \cite{pandey2021total} introduce a pipeline that predicts a set of specular maps with varying degrees of Phong exponents using estimated surface normals and an input HDR environment map. These maps along with diffuse and albedo maps are used to predict a relit image with a UNet. 
Yeh et al. \cite{yeh2022learning} and Kim et al. \cite{kim2024switchlight} follow a pipeline similar to \cite{pandey2021total} with different modifications: Yeh et al. use synthetic light stage data from 3D scans and an albedo refinement step to reduce the synthetic-real gap, while Kim at al. adopt the Cook-Torrance model with an additional pre-training stage. 
Hou et al.~\cite{hou2021towards} fit a morphable model and use ray tracing to compute a shadow map for predicting pixel luminance ratios in relighting. Later, Hou et al.~\cite{hou2022face} predicted a shadow mask from estimated depth via ray tracing and rendered relit images using network-predicted albedo and shading maps.
%Hou et al. \cite{hou2021towards} compute a shadow map based on a morphable model fitted to the input and standard ray tracing, then use it to help predict the ratio of pixel luminance changes for relighting. 
%Hou et al. \cite{hou2022face} predict a shadow mask via ray tracing based on their estimated depth map and render a relit image with estimated albedo and shading maps from networks.

While these methods \cite{nestmeyer2020learning, hou2021towards, hou2022face} produce promising non-diffuse effects,
their reliance on physical image formation makes them sensitive to geometry errors. 
Thus, some \cite{hou2021towards, hou2022face} can only relight the face, not ears or hair, and still fail on in-the-wild cast shadows (Figure \ref{fig:ffhq_cs} and \ref{fig:rotate_cs}). 
Neural rendering methods~\cite{wang2020single, pandey2021total, kim2024switchlight} handle estimation errors better but often lose high-frequency details, even when predicted by a UNet \cite{wang2020single}.
Many methods~\cite{pandey2021total, kim2024switchlight} rely on high-quality, proprietary light stage data for training, but capturing large, diverse datasets remains challenging.
Synthetic light stage data~\cite{yeh2022learning} helps bridge the domain gap but depends on 3D face scans, which are harder to obtain than 2D images.
%shows great potential in bridging the domain gap but currently relies on 3D face scans for generation, which are more difficult to obtain compared to 2D images.
Fine-tuning strategies~\cite{ren2023relightful, kim2024switchlight} further improve results, yet dependence on light stage data remains unavoidable.
Other related studies \cite{he2024diffrelight, saito2024relightable} focus on subject-specific relighting using 3D Gaussian splatting but are unsuitable for in-the-wild relighting, as they require diffuse inputs \cite{he2024diffrelight} or multiview video data \cite{saito2024relightable}, which are not readily obtainable from in-the-wild scenarios.
Another research direction~\cite{zeng2024dilightnet, jin2024neural_gaffer} targets object relighting, but these methods typically do not generalize well to faces and require multi-view light-stage data for fine-tuning, which lacks the quality and diversity of object datasets.

\textbf{Style transfer-based methods.} Some approaches~\cite{li2018closed, luan2017deep, shih2014style} adapt style transfer to relighting by transferring lighting and shading styles between images, though they were not originally designed for face relighting.
However, these style-based methods capture information beyond lighting and fail to produce accurate relit results. Shu et al.~\cite{shu2017portrait} address relighting with spatially varying color histogram matching based on face geometry, formulated as a mass transport problem. Yet, it does not model self-occlusion for cast shadows and is sensitive to occlusion from hair or accessories.
%do not directly solve face relighting, they can be adapted for this task by transferring the lighting and shading styles from one image to another.
%However, the style representation used in these methods captures broad information beyond the lighting condition and cannot produce accurate relit results.
%Shu et al. \cite{shu2017portrait} solves relighting using color histogram matching that is redesigned be spatially varying and dependent on the face geometry and can be solved as a mass transport problem. However, this technique does not model self-occlusion required for handling cast shadows and can easily suffer from occlusion by hair or accessories.
Other methods~\cite{ren2023relightful,iclight,wang2023semi,Harmonizer} perform background-based harmonization, transferring lighting from background to subject. Since they require a background and focus on composites, they are not directly comparable to our method.
%use background-based harmonization for compositing, transferring lighting from the background to the subject. As they require a background image and target realistic composites, they are not directly comparable to our relighting method.
%Other methods~\cite{ren2023relightful, iclight, wang2023semi, Harmonizer} involve background-based harmonization techniques developed within a compositing context.  These approaches require a background image to transfer the lighting style onto the foreground subject and aim to achieve a realistic composite, which makes them not directly comparable to our relighting method.

\iffalse
\textbf{Style transfer-based methods.} Another class of relighting approaches is based on style transfer. Although some of these methods \cite{li2018closed, luan2017deep, shih2014style} do not directly solve face relighting, they can be adapted for this task by transferring the lighting and shading styles from one image to another.
However, the style representation used in these methods captures broad information beyond the lighting condition and cannot produce accurate relit results.
Shu et al. \cite{shu2017portrait} solves relighting using color histogram matching that is redesigned be spatially varying and dependent on the face geometry and can be solved as a mass transport problem. However, this technique does not model self-occlusion required for handling cast shadows and can easily suffer from occlusion by hair or accessories.
\fi

%  GAN

\textbf{GAN-based methods.} A few techniques use GANs \cite{goodfellow2020generative} for relighting \cite{tewari2020pie, mallikarjun2021photoapp, ranjan2023facelit}. 
Ranjan et al. \cite{ranjan2023facelit} condition a tri-plane generator in EG3D~\cite{chan2022efficient} on lighting estimates from DECA~\cite{DECA:Siggraph2021} to learn a disentangled 3D generative model with lighting and pose control. However, they do not demonstrate relighting results for real images.
Mei et al. \cite{mei2024holo} also build on EG3D and implicitly model lighting effects instead of using explicit reflectance models, improving upon Ranjan et al. \cite{ranjan2023facelit} and successfully handling real images.
Tewari et al.~\cite{tewari2020pie} build on StyleRig~\cite{tewari2020stylerig}, which provides semantic control of StyleGAN~\cite{karras2019style} by mapping morphable model parameters and a StyleGAN latent code to a new code representing target parameters. While StyleRig originally worked only on synthetic images, they extend it to real images by optimizing a latent code to reconstruct the input and then manipulating the lighting through StyleRig.
%Tewari et al.~\cite{tewari2020pie} extend StyleRig~\cite{tewari2020stylerig}, which enables semantic control of StyleGAN~\cite{karras2019style}, by mapping morphable model parameters and an initial StyleGAN latent code to target parameters.
%Specifically, they extend StyleRig, which only works on synthetic images, to real images by optimizing a latent code that reproduces the input image and use StyleRig to manipulate the lighting condition. 
Similarly, Mallikarjun et al.~\cite{mallikarjun2021photoapp} map target illumination and a pSp-predicted StyleGAN code~\cite{richardson2021encoding} to a new code for relighting. However, their imperfect GAN inversion often alters identity and facial details.
%Similarly, Mallikarjun et al. \cite{mallikarjun2021photoapp} map a target illumination and a StyleGAN latent code predicted from pSp network \cite{richardson2021encoding} to a new code that represents a relit image. However, these techniques tend to change the identity and facial details of the input person due to the imperfect GAN inversion. 
New GAN inversion methods~\cite{feng2022near, roich2022pivotal, xie2023high} are promising, but only \cite{mei2024holo} show real-image relighting. However, it requires fine-tuning EG3D for each input image. In contrast, our method leverages DDIM's near-perfect inversion to preserve details without person-specific fine-tuning.

\textbf{Acceleration of diffusion sampling.}
%Acceleration techniques fall into two groups: training-free and training-based. 
Training-based methods~\cite{watson2021learning, bao2022analytic, sauer2023adversarial, song2023consistency, xu2023ufogen} address this at the cost of additional training, often via distillation. Recent approaches such as UFOGen \cite{xu2023ufogen}, Consistency Models \cite{song2023consistency}, and Adversarial Diffusion Distillation \cite{sauer2023adversarial} achieve single-step sampling but require complex losses (e.g., adversarial) and still lag behind teacher performance.
Training-free methods \cite{lu2022dpm, lu2022dpm++, zhao2023unipc} use fast solvers applicable to any pretrained diffusion model, reducing sampling to 5–10 steps but still short of single-shot inference, with a notable quality–speed trade-off.
%However, they currently reduce the sampling process to 5-10 steps, which fall short of achieving single-shot inference. Additionally, all of the above techniques involve a significant trade-off between quality and sampling speed.

This paper demonstrates that training a network on relit pairs from pretrained DiFaReli++ with simple L2 loss enables single-step generation with quality surpassing DiFaReli++.
  \section{Approach}

% Given an input face image, we seek to relight this image under a target lighting condition, described by spherical harmonic coefficients and an additional scalar representing the ``degree'' of visible cast shadows.
Given an input face image, we aim to relight it under a target lighting condition, described by SH coefficients and a representation of cast shadows. 
% In DiFaReli, we use a scalar value to represent the ``intensity'' of visible cast shadows. In DiFaReli++, we represent both the intensity and shape of cast shadows with a shadow map.
In DiFaReli++, we represent cast shadows with a shadow map capturing both intensity and shape, unlike DiFaReli, which uses only a scalar for intensity.

To explain our method, we first cover the relevant background on DDIM \cite{song2021denoising} and a key finding from DiffAE \cite{preechakul2022diffusion} in Section \ref{sec:conddiff}, which demonstrates how a conditional DDIM can perform attribute manipulation on real images by functioning as both a decoder and a `stochastic' encoder.
In Sections \ref{sec:method_overview} to \ref{sec:relighting}, we review the core concepts of DiFaReli, including its conditioning technique, as well as the details of its training and relighting processes. Finally, in Section \ref{sec:difareli++}, we introduce DiFaReli++, an extension of DiFaReli that incorporates shadow map conditioning, non-facial part relighting, and a single-shot relighting framework.

\subsection{Background: Conditional DDIM \& DiffAE}
\label{sec:conddiff}
Our method relies on a conditional Denoising Diffusion Implicit Model (DDIM) \cite{song2021denoising}, which is a variant of diffusion models \cite{pmlr-v37-sohl-dickstein15, ho2020denoising, NEURIPS2019_3001ef25}. 
% (For a full review and notation convention, please refer to \cite{song2021denoising}.) 
Unlike standard diffusion models, DDIM uses a non-Markovian inference process that relies on the conditional distribution $q(\vect{x}_{t-1} \mid \xt, \xzero)$ that is conditioned on $\xzero$ (the original image) in addition to $\xt$.
One important implication is that the generative process can be made deterministic, allowing us to deterministically map $\xT \sim \mathcal{N}(\vect{0}, \vect{I})$ to $\xzero$ and vice versa. 
Here the mapping from $\xzero$ to $\xT$ can be viewed as the encoding of an input image $\xzero$ to a latent variable $\xT$.
% This mapping from $\xzero$ to $\xT$ can be viewed as the encoding of an input image $\xzero$ to a latent variable $\xT$.

Diffusion Autoencoders (DiffAE) \cite{preechakul2022diffusion} show that such image encoding yields $\xT$ that contains little semantic information about the input image $\xzero$ and propose to condition the DDIM also on a learnable latent variable $\vect{z}$ predicted from a separate image encoder. By jointly training the image encoder and the DDIM, the encoded $\vect{z}$ now captures meaningful semantics, while the encoded $\xT$, inferred by reversing the deterministic generative process of the DDIM, captures the rest of the information not encoded in $\vect{z}$, such as stochastic variations. 
%Like in DDIM, $\xT$ is inferred or encoded by reversing the generative process of the conditional DDIM.
%Here DiffAE is used to encode an image to a two-part latent code ($\vect{z}$, $\xT$), where $\vect{z}$ is produced from the image encoder and $\xT$ from reversing the generation process of the conditional DDIM. 
The resulting latent code ($\vect{z}$, $\xT$) can also be decoded back to the input image near-perfectly using the same conditional DDIM.
%or used for attribute manipulation by modifying $\vect{z}$ with a linear operation. 
%In this framework, their DDIM thus acts as both a decoder and an encoder for stochastic variations in the image. 
By modifying the semantic latent $\vect{z}$ and decoding the new ($\vect{z}'$, $\xT$), DiffAE can manipulate semantic attributes of a real input image---a capability that inspires our work.

\subsection{DiFaReli: Method overview}
\label{sec:method_overview}
The general idea of DiFaReli is to encode the input image into a feature vector that disentangles the light information from other information about the input image. Then, the relit image is produced by modifying the light encoding in the feature vector and decoding the resulting vector with a conditional DDIM (Figure \ref{pipeline_overview}).
%with a diffusion model. 
This process is similar to how DiffAE performs attribute manipulation; however, our task requires well-disentangled and interpretable light encoding that facilitates lighting manipulation by the user. 

To solve this, we use off-the-shelf estimators to encode an input image into light, shape, and camera encodings, as well as a face embedding, a shadow scalar, and a background image (Section \ref{sec:encoding}).
%and camera parameters, as well as a face embedding, cast shadow, and a background image (Section \ref{sec:encoding}). 
%The light, shape, and camera encodings will be used to render a shading reference (Section \ref{sec:render}). 
Then, these encodings are used to condition our DDIM decoder (Section \ref{sec:decoder-modulator}) with a novel conditioning technique (Section \ref{sec:conditioning}).
%that guides the decoding with a 2D shading reference physically rendered from our encodings (Section \ref{sec:conditioning}). 
%The shading reference and all encodings will be used to condition our DDIM decoder (Section \ref{sec:decoder-modulator}) with our novel conditioning technique (Section \ref{sec:conditioning}).
% with a novel conditioning technique that guides the decoding with a 2D shading reference physically rendered from our encodings (Section \ref{sec:conditioning}). 
For training, we use a standard diffusion objective to reconstruct training images (Section \ref{sec:training}). To relight, we reverse the generative process of the DDIM conditioned on the input's encodings to obtain $\xT$, modify the light encoding, and decode $\xT$ using the modified encodings (Section \ref{sec:relighting}).

\subsection{DiFaReli: Encoding}
\label{sec:encoding}
The goal of this step is to encode the input face image $I \in \mathbb{R}^{H \times W \times 3}$ into a feature vector:
\begin{equation}
    \vect{f} = (\vect{l}, \vect{s}, \mathbf{cam}, \boldsymbol{\xi}, c, \vect{bg}%, \xT
    ),\label{eq:feature_vec}
\end{equation}
where $\vect{l} \in \mathbb{R}^{9 \times 3}$ represents $2^\text{nd}$-order spherical harmonic lighting coefficients, $\vect{s} \in \mathbb{R}^{|\vect{s}|}$ represents parameterized face shape, $\mathbf{cam} \in \mathbb{R} ^ {1 + 2}$ represents orthographic camera parameters, $\boldsymbol{\xi} \in \mathbb{R}^{512}$ is a deep feature embedding based on ArcFace \cite{deng2019arcface}, $c$ is a scalar that indicates the intensity of visible cast shadows, and $\vect{bg} \in \mathbb{R}^{H \times W \times 3}$ contains the background pixels with the face, hair, neck masked out.
%and $\xT \in \mathbb{R}^{H \times W \times 3}$ is the latent variable from reversing our DDIM's generative process. 
These variables will be inferred using off-the-shelf or pretrained estimators.

\textbf{Light, shape, \& camera encodings $(\vect{l}, \vect{s}, \mathbf{cam})$.}\label{sec:cond-light}
% We use an off-the-shelf single-view 3D face reconstruction method, DECA \cite{DECA:Siggraph2021}. Given a face image, DECA predicts the 3D face shape, camera pose, albedo map, and spherical harmonic lighting (SH) coefficients.
We use DECA \cite{DECA:Siggraph2021}, an off-the-shelf single-view 3D face reconstruction method that predicts the 3D face shape, camera pose, albedo map, and SH lighting coefficients given a face image.

For our light encoding $\vect{l}$, we directly use the SH coefficients from DECA, consisting of 9 coefficients for each channel of the RGB.
DECA's 3D face shape is parameterized based on FLAME model \cite{FLAME:SiggraphAsia2017} as blendshapes with three linear bases for identity shape, pose, and expression. Their respective coefficients are denoted by $\boldsymbol{\beta}$, $\boldsymbol{\theta}$, $\boldsymbol{\psi}$. Our face shape encoding $\vect{s}$ is the combined $(\boldsymbol{\beta}, \boldsymbol{\theta}, \boldsymbol{\psi}) \in \mathbb{R} ^{|\boldsymbol{\beta}| +| \boldsymbol{\theta}| + |\boldsymbol{\psi}|}$.  DECA assumes orthographic projection and models the camera pose with isotropic scaling and 2D translation. We combine the scaling and translation parameters into $\mathbf{cam} \in \mathbb{R} ^ {1 + 2}$.Note that we do not use the predicted albedo map, as its estimation by DECA can be unreliable and was found empirically unnecessary.

\textbf{Identity encoding $(\boldsymbol{\xi})$.}
% To compute our deep feature embedding that helps preserve the input's identity, we use ArcFace\cite{deng2019arcface}, a pretrained face recognition model based on ResNet \cite{resnet}.
To help preserve the input's identity, we use ArcFace~\cite{deng2019arcface}, a pretrained face recognition model based on ResNet \cite{resnet}.
This model has been shown to produce discriminative and identity-preserving feature embeddings.
%useful for preserving the person's identity in our DDIM.

%To help preserve the person's identity during relighting, we use ArcFace\cite{deng2019arcface}, a pre-trained face recognition model based on ResNet \cite{resnet}, to infer our deep feature embedding $\boldsymbol{\xi} \in \mathbb{R}^{512}$. This model has been shown to produce discriminative and identity-preserving embeddings.
%useful for face recognition tasks.
%we condition the DDIM with a face embedding of a given image, extracted by the ArcFace\cite{deng2019arcface}. %ArcFace is the model trained to solve the face recognition. According to this foundation, a face embedding from ArcFace contains information useful in distinguishing a person's identity. 
\textbf{Cast shadow encoding $(c)$.}\label{sec:cs_encoding}
This scalar describes the intensity of visible cast shadows, typically caused by a dominant point or directional light source, such as the sun.

We trained a model to estimate $c$ from a face image ourselves and fixed this pretrained estimator. To do this, we manually labeled around 1,000 face images with binary flags indicating whether cast shadows are visible. Following a technique proposed in DiffAE \cite{preechakul2022diffusion}, we first use DiffAE's pretrained encoder to map each face image to a semantically meaningful latent code $\vect{z}$ and train a logistic regression classifier on $\vect{z}$ to predict the flag $c$ is then computed as the logit value of the logistic regression. As shown in \cite{preechakul2022diffusion}, this technique helps reduce the number of training examples required to achieve good accuracy, but we note that $c$ can be estimated in other ways, such as with a CNN. 

\textbf{Background encoding $(\vect{bg})$.}
\label{sec:cond-bg}
% To help fix the background during relighting, we condition the DDIM with an image of the input's background. The background region is detected using a face segmentation algorithm \cite{yu2018bisenet}. The ears, hair, and neck are not part of the background and can be relit by our algorithm (see Figure \ref{fig:bg_ab}).
To help fix the background during relighting, we condition DDIM with an image of the input’s background detected by a face segmentation algorithm \cite{yu2018bisenet}. The ears, hair, and neck are excluded from the background and can still be relit by our algorithm (Figure \ref{fig:bg_ab}).

\subsection{DiFaReli: DDIM decoder \& Modulator network}
\label{sec:decoder-modulator}
Our main network is a conditional DDIM that decodes our feature vector (with modified lighting information) to a relit version of the input image. In practice, the feature vector is used to \emph{condition} the DDIM that maps $\xT \sim \mathcal{N}(\vect{0}, \vect{I})$ to the original input $\xzero$ during training or maps $\xT = \text{DDIM}^{-1}(\xzero)$ from reversing the generative process to the relit output during relighting (Section \ref{sec:relighting}). 
%This conditioning involves another network called \emph{Modulator}, which maps the rendered image and the background image into spatial modulation weights for the DDIM decoder.
This conditioning involves another network called \emph{Modulator} network, which converts the light, shape, and camera encodings into spatial modulation weights for the DDIM decoder.

The architecture of the DDIM decoder is based on Dhariwal et al. \cite{dhariwal2021diffusion}, which is a modified UNet built from a stack of residual blocks interleaved with self-attention layers
% We provide full details in Appendix \ref{app:net_arch}. 
(full details in Appendix III).
% (full details in Appendix \ref{app:net_arch}).

\subsection{DiFaReli: Conditioning DDIM decoder}
\label{sec:conditioning}

Conditioning a diffusion model on a condition vector can be done in various ways, such as through adaptive group normalization \cite{wu2018group, preechakul2022diffusion, dhariwal2021diffusion} or attention-based mechanisms \cite{nichol2021glide, rombach2021highresolution}, among others. 
%For all conditional variables except the rendered image and the background image, we employ ``non-spatial'' conditioning technique based on an adaptive group normalization (AdaGN). We propose the ``spatial'' conditioning technique for the rendered image and the background image. Each conditioning method is explained in the following subsections.
In our problem, the lighting information is encoded explicitly as SH coefficients and their interaction with 3D shape, specifically the surface normals, can be precisely modeled with the SH lighting equation. 
Our idea is to ease the modeling of the known interaction by rendering a shading reference of the target relit face. The primary goal of this reference is to convey the information about the target lighting and shading in a spatially-aligned manner, not the geometry or the exact shading intensities.
%compared to treating the SH light as a 27-dim vector. 
The following sections detail this ``spatial'' conditioning technique as well as a standard non-spatial conditioning technique used for other encodings. 

%Thus, instead of flattening the SH coefficients or 3D shape parameters into condition vectors as in other conditioning techniques, our idea is to model the known interaction explicitly by rendering a shading reference of the target relit face for conditioning the DDIM.

\textbf{Spatial conditioning.} 
This technique is used for the light, shape, camera and background encodings $(\vect{l}, \vect{s}, \allowbreak \mathbf{cam}, \allowbreak \vect{bg})$. 
Given the face shape $\vect{s}$, we first convert it to a triangle mesh using the three linear bases of FLAME model \cite{FLAME:SiggraphAsia2017} and remove the ears, eyeballs, neck, and scalp from the mesh to retain only the face region (See Figure \ref{pipeline_overview}). We remove those parts because they are often inaccurate and hard to estimate correctly (e.g., occluded ears behind hair). We assume a constant gray albedo (0.7, 0.7, 0.7) and render this mesh in the camera pose described by $\mathbf{cam}$ with surface colors computed with $\vect{l}$ using the standard SH lighting equation. 
% The details are in Appendix \ref{app:render_face},
The details are in Appendix~IV,
and we discuss this albedo choice and the inherent albedo-light ambiguity in Section \ref{sec:limitations}.

Then, this shading reference $R$, which shows a shaded face in the shape and pose of the input person under the target lighting, is concatenated with the background image $\vect{bg}$ and fed to the Modulator network.
%(Please refer to Appendix \ref{app:bg_cond} for the study of background conditioning). 
%This technique is used for the rendered image and the background $(\vect{R}, \vect{bg})$. We concatenate the rendered image $\vect{R}$ with the background image and feed them to the Modulator network (Please refer to Appendix \ref{app:bg_cond} for the study of background conditioning).
Let us denote the output of each residual block $i$ in the Modulator network by $\vect{m}_i \in \mathbb{R}^{H_i \times W_i \times D_i}$, and the output of the corresponding residual block in the identical DDIM's first half by $\vect{o}_i \in \mathbb{R}^{H_i \times W_i \times D_i}$. In the DDIM, we take each residual block's output $\vect{o}_i$ and replace it with $\vect{o}'_i$, which will be used as input to the subsequent layer in the network:
\begin{align}
    \vect{o}'_{i} &= \vect{o}_{i} \odot \tanh(\vect{m}_{i}),
\end{align}
where $\odot$ is the element-wise multiplication. This conditioning technique allows the shaded image $R$ and the background to retain their spatial structure and facilitate local conditioning of the generation as they are spatially aligned with the input (e.g., their facial parts and background are in the same positions). 
% And by turning SH coefficients into shading intensity values in $R$ that correlate directly with the output RGB pixels, the DDIM can learn this linear relationship more effectively, compared to using the 27-dim SH coefficients as a non-spatial condition vector.

% non-spatial
\textbf{Non-spatial conditioning.}\label{sec:non_spatial} This technique is used for $(\vect{s}, \mathbf{cam}, \boldsymbol{\xi}, c)$.
%except $\vect{bg}$, which is the only inherently spatial encoding. 
The direct use of $\vect{s}, \mathbf{cam}$ again in this technique is empirically found to be helpful, in addition to their indirect use through the shading reference.
We use a similar conditioning technique as used in \cite{dhariwal2021diffusion, preechakul2022diffusion} based on adaptive group normalization (AdaGN) \cite{wu2018group} for these encodings and also for the time embedding in the standard diffusion model training $\gamma(t)$, where $\gamma$ is a sinusoidal encoding function \cite{dhariwal2021diffusion}.
%for all encodings except $\vect{bg}$, which is the only encoding that is not a flattened 1D vector.
%This technique is used for the light, shape, camera, identity and cast shadow encodings $(\vect{l},\vect{s},\mathbf{cam},\boldsymbol{\xi}, c)$, as well as the time embedding in the standard diffusion model training $\gamma(t)$, where $\gamma$ is a sinusoidal encoding function and $t$ is the diffusion timestep \cite{dhariwal2021diffusion}.
% These encodings are non-spatial by design or nature: $\boldsymbol{\xi}$ comes from a latent vector space of a face recognition model, and $c$ is a scalar that globally affects cast shadows in the entire image. 
%For these encodings, we use a similar conditioning technique as used in \cite{dhariwal2021diffusion, preechakul2022diffusion} based on adaptive group %normalization (AdaGN) \cite{wu2018group}. 
Given an input feature map $\vect{h}_j \in \mathbb{R}^{H_j \times W_j \times D_j}$, we compute
%, our AdaGN is computed by:
\begin{equation}
    \mathrm{AdaGN}_j(\vect{h}_j,\vect{s},\mathbf{cam},\boldsymbol{\xi}, c, t) = \vect{k}_j(\textbf{t}_{j}^{s}\mathrm{GN}(\vect{h}_j)+\textbf{t}_{j}^{b}),
\end{equation}
%where $\vect{h}$ is a modulated feature map in the previous section, 
where $\vect{k}_j=\text{MLP}^3_j(\text{Concat}(\vect{s},\mathbf{cam},\boldsymbol{\xi}, c)) \in \mathbb{R}^{D_j}$ is the output of a 3-layer MLP with the SiLU activation \cite{elfwing2018sigmoid}, and $(\textbf{t}_j^{s},\textbf{t}_j^{b}) \in \mathbb{R}^{2\times D_j} = \text{MLP}^1_j(\gamma(t))$ is the output from a single-layer MLP also with the SiLU activation. $\mathrm{GN}$ is the standard group normalization.
%and $\gamma$ is a sinusoidal encoding function. 
We apply our AdaGN in place of all the AdaGNs in the original architecture of \cite{dhariwal2021diffusion}, which occur throughout the UNet 
(Details in Appendix III).

\subsection{DiFaReli: Training}
\label{sec:training}
% Our main DDIM takes $(\text{Enc}_{c}(\vect{l'}, \vect{s}, \mathbf{cam}), \boldsymbol{\xi}, c, \vect{bg}, \xT)$ where $\text{Enc}_{c}$ is the \emph{conditioner network}, as an input and output an image $x_{0}$. 
% and define the network as a learnable generative process that try to match the DDIM generative process $q(\vect{x}_{t-1} \mid \xT, x_{0}).$ To train the network
We jointly train the DDIM decoder, parameterized as a noise prediction network $\boldsymbol{\epsilon}_\theta$, and the Modulator network $M_\phi(\vect{l}, \vect{s}, \mathbf{cam}, \vect{bg})$ using standard diffusion training \cite{ho2020denoising, song2021denoising, preechakul2022diffusion}. Here we consider the MLPs in Figure \ref{pipeline_overview} as part of the DDIM. We adopt the simplified, re-weighted version of the variational lower bound with $\boldsymbol\epsilon$ parameterization: 
\begin{equation}
L_\text{simple} = \\ \mathbb{E}_{t, \xzero, \boldsymbol{\epsilon}}\| \boldsymbol{\epsilon}_{\theta}(\vect{x}_t, t, 
M_\phi,
\vect{s}, \mathbf{cam},
\boldsymbol{\xi}, c) - \boldsymbol{\epsilon}\|^{2}_{2},
\end{equation}
where $\boldsymbol{\epsilon}_\theta$ is trained to predict the added noise $\boldsymbol{\epsilon} \sim \mathcal{N}(\vect{0}, \vect{I})$ in $\vect{x}_{t} = \sqrt{\alpha_t}\xzero + \sqrt{1-\alpha_t} \vect{\epsilon}$, given a training image $\xzero$. We define $\alpha_t$ as $\prod_{s=1}^t (1-\beta_s)$, where $\beta_t$ is the noise level at timestep $t$ in the Gaussian diffusion process $q(\xt|\xtone) = \mathcal{N}(\sqrt{1 - \beta_t}\xtone, \beta_t \mathbf{I})$. We use a linear noise schedule and a total step $T=1000$. Note that we do not reverse $\xT=\text{DDIM}^{-1}(\xzero)$ during training.
%{Modulator network}, $\textbf{Enc}^{cond}$.
%We follow  for our training protocol. 
%Given an input image $\xzero$ and $\boldsymbol{\epsilon}_{t} \sim \mathcal{N}(0,\vect{I})$, we can compute $\vect{x}_{t}$ by marginal distribution  $q(\vect{x}_{t}\mid \xzero) = \mathcal{N}(\sqrt{\alpha_{t}}\xzero, (1-\alpha_{t})\vect{I})$. The main DDIM network acts like a learnable noise predictor $\boldsymbol{\epsilon}_{\theta}(\textbf{Enc}^{cond}(\vect{l'}, \vect{s}, \mathbf{cam}), \boldsymbol{\xi}, c, \vect{bg}, \vect{x}_{t})$ that is trained to predicts the noise added $\xzero$ to get $\vect{x}_{t}$. The training objective is to optimize: 
%\begin{equation}
%L_{simple} = \sum_{t=1}^{T} \mathbb{E}_{\xzero, \boldsymbol{\epsilon}_{t}}\| \boldsymbol{\epsilon}_{\theta} - \boldsymbol{\epsilon}_{t}\|^{2}_{2}
%\end{equation}
% which is the L2-loss between the predicted noise and the ground truth noise $\boldsymbol{\epsilon}_{t}$. Here, we use $T=1000$. 

\subsection{DiFaReli: Relighting}
\label{sec:relighting}
To relight an input image, we first encode the input image into our feature vector $\vect{f}$ (Equation \ref{eq:feature_vec}), then reverse the deterministic generative process of our DDIM conditioned on $\vect{f}$, starting from the input image $\xzero$ to $\vect{x}_{T=1000}$.
\begin{align}
\vect{x}_{t+1} = \sqrt{\alpha_{t+1}}\vect{g}_\theta(\xt, t, \vect{f}) +\sqrt{1-\alpha_{t+1}}\boldsymbol{\epsilon}_\theta(\vect{x}_t, t, \vect{f}),\label{eq:ddim_reverse}
\end{align}
where $\vect{g}_\theta$ represents the predicted $\xzero$, which is reparameterized from $\boldsymbol\epsilon_\theta$ and is computed by:
\begin{equation}
\vect{g}_\theta(\xt, t, \vect{f}) = \frac{1}{\sqrt{\alpha_t}}\left( 
\xt - \sqrt{1 - \alpha_t} \boldsymbol{\epsilon}_\theta(\vect{x}_t, t, \vect{f})
\right).
\end{equation}
After obtaining $\xT$, we modify the SH light encoding $\vect{l}$ and the cast shadow flag $c$ to the target $\vect{l}'$ and $c'$, which can be set manually or inferred from a reference lighting image using DECA and our cast shadow estimator. Then, we decode the modified $\vect{f}' = (\vect{l}', \vect{s}, \mathbf{cam}, \boldsymbol{\xi}, c', \vect{bg})$ using the reverse of Equation \ref{eq:ddim_reverse}, starting from $\xT$ to produce the final output.

The reverse process to obtain $\xT$ is key to reproducing high-frequency details from the input image. As demonstrated in DiffAE \cite{preechakul2022diffusion}, DDIM will encode any information not captured in the conditioning feature vector $\vect{f}$ in the noise map $\xT$. This information includes high-frequency details, such as the hair pattern or skin texture.

\textbf{Improved DDIM sampling with mean-matching.} We observe that when the input image contains a background with extreme intensities (e.g., too dark or too bright), DDIM can produce results with a slight change in the overall brightness. We alleviate this issue by computing the mean pixel difference between each $\xt$ during DDIM's generative reversal ($\xzero \rightarrow \xT$) and $\xt$ from self-decoding of the reversed noise $\xT$. This sequence of mean differences is then applied to the decoding for relighting 
(Appendix II).
% (Appendix \ref{app:impl}).

\vspace{1em}
\section{DiFaReli++: Single-Shot Face Relighting with consistent cast shadows}\label{sec:difareli++}

We present three improvements to DiFaReli \cite{ponglertnapakorn2023difareli}: 1) the generation of consistent cast shadows under dynamic lighting, 2)
extension of relighting to non-facial parts (e.g., clothes, earrings, hats), and 3) single-shot face relighting.
% We present two improvements to DiFaReli \cite{ponglertnapakorn2023difareli}: single-shot face relighting (DiFaReli$_\text{ss}$) and  extending relighting to non-facial parts (DiFaReli$_\text{nf}$) like clothes, earrings, and hats.
%In this section, we demonstrate two improvements to DiFaReli \cite{ponglertnapakorn2023difareli}. First, we show how to enhance its inference speed, enabling relighting to be performed in a single-shot inference. Second, we extend the relightable parts to include non-facial parts such as clothes, earrings, or hats.

% \begin{figure}[ht!]
% \centering
%   \includegraphics[scale=0.4]{figures/shadow_mask_cmp.pdf}
%   \caption{Shadow masks}
%   \label{fig:shadow_mask}
% \end{figure}

\subsection{Face relighting with consistent cast shadows.}
\label{sec:cs}
% To enable relighting with cast shadow effects, we build on a similar idea from DiFaReli \cite{ponglertnapakorn2023difareli}, where lighting information is conveyed through a shading reference. In the DiFaReli's pipeline, a scalar shadow flag (see Section \ref{sec:cs_encoding}) allows users to control cast shadow intensity but does not offer control over cast shadow position. This is due to the lack of cast shadow position in the conditioning signals. 
% To enable relighting with cast shadow effects, we followed the similar idea from DiFaReli \cite{ponglertnapakorn2023difareli}, where by lighting information is conveyed through a shading reference. In the DiFaReli's pipeline, a scalar shadow flag (see Section \ref{sec:cs_encoding}) allows users to control cast shadow intensity but does not offer control over cast shadow position. This is due to the lack of cast shadow position in the conditioning signals. 
%To enable relighting with cast shadow effects, we follow the idea of DiFaReli \cite{ponglertnapakorn2023difareli} by conditioning the DDIM with shadow information in addition to the existing light information.
To enable relighting with consistent cast shadows, we build on the idea of DiFaReli, which achieves relighting by conditioning a DDIM with lighting information. Here, the idea is to extend the lighting condition to include more comprehensive shadow information beyond the single scalar shadow flag used in DiFaReli (Section \ref{sec:cs_encoding}). Specifically, we use a binary shadow map to specify the face areas under cast shadows.
Once the network learns to utilize this shadow conditioning, the relit output will feature realistic cast shadows that align with the provided shadow map under the target lighting.

The key challenges are how to estimate the shadow map of the input image for training and the target shadow map under the target lighting for inference. For this idea to work, the shadow maps used for training need to be highly accurate and spatially aligned with the input's shadows; otherwise, the network will disregard them as noisy and unhelpful for the reconstruction training objective. This requirement precludes the use of ray tracing, as employed in prior work \cite{hou2022face}, which relies on accurate geometry to estimate accurate shadow maps.

\begin{figure}[ht]
\centering
  \includegraphics[scale=0.58]{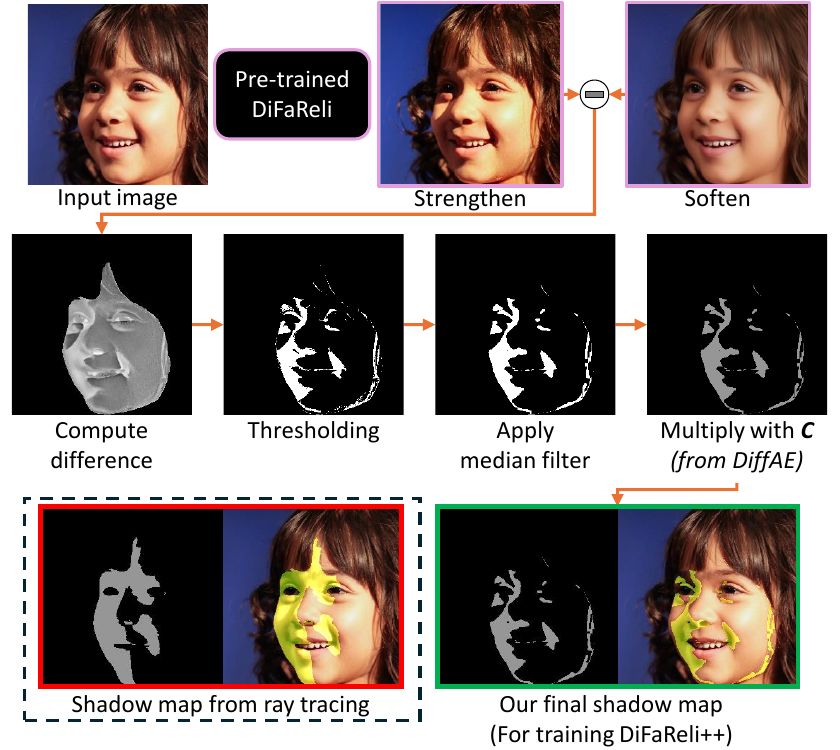}
  \caption{
  \textbf{Computing the shadow map for training.} We used a pretrained DiFaReli model to generate stronger and reduced versions of the input image, then identify shadow areas through pixel differences.
  Our process produces more accurate and spatially aligned shadow maps compared to ray-traced maps shown in red, which suffer from inaccurate lighting and geometry estimation.
  }
  \label{fig:compute_shadow_map}
\end{figure}

To estimate shadow maps, we instead propose leveraging DiFaReli to transform the input image into versions with stronger and reduced cast shadows, then identify cast shadow areas through pixel differences  (Figure \ref{fig:compute_shadow_map}). However, this technique can only be used for training but not inference, as the two versions have incorrect \emph{target} lighting shadows. Fortunately, we can leverage a ray-traced shadow map at test time and produce consistent and plausible outputs. We next explain the process in detail.

%This paper proposes an alternative way to estimate the shadow map from a single input image that does not require accurate light or face geometry estimation and can easily scale to any in-the-wild image dataset. Our shadow map effectively captures cast shadow regions in the input image, as shown in Figure \ref{fig:compute_shadow_map}.
%Details on shadow map computation, training, and inference of DiFaReli++ are in the next section.
% Following DiFaReli \cite{ponglertnapakorn2023difareli}, we train the network by concatenating the shadow map with the shading reference and passing them into the Modulator network. At test time, since we do not have access to the new shadow map of the same person, we can generate plausible cast shadow effects by computing this map using ray tracing. Notably, using ray tracing at this stage does not introduce misalignment issues or affected from the inaccurate light or face estimation. Since during inference, we only need the shadow map to guide where shadows should be generated in the relit image. In contrast, misalignment during the training stage is much more severe, as the diffusion model could completely ignore the condition due to its noisiness.

% Specifically, we use DiFaReli to compute the strengthened and diffused versions of the input image (see Section \ref{sec:shadow_mani}). We then calculate the difference between these two versions to produce the shadow map, as illustrated in Figure \ref{fig:compute_shadow_diff}. 

\subsubsection{Computing a shadow map}\label{sec:shadowmap}
We use our pretrained DiFaReli model to relight an input image into two versions, one with stronger and one with reduced cast shadows, by setting the shadow value ($c$) to the maximum and minimum estimated shadow values across all training images.
% These min-max shadow values ($c$) are derived using argmax and argmin operations on the training set.
These values are estimated by DiffAE \cite{preechakul2022diffusion} on the FFHQ dataset as explained in Section \ref{sec:cs_encoding}.
%Each shadow value in the training set is a single scalar per image, estimated by DiffAE \cite{preechakul2022diffusion} (see Section \ref{sec:cs_encoding}) on the FFHQ dataset. 
%We next compute the pixel difference between the strengthened and its diffuse version. Next, we apply thresholding to the computed pixel difference map, resulting in a binary mask where 1 indicates shadow areas and 0 indicates non-shadow areas.
We then compute the pixel difference between the two versions and apply thresholding to the difference map to create a binary mask, where 1 indicates cast shadow areas and 0 indicates non-shadow areas. By generating the two versions using consistent global maximum and minimum levels, we can determine a single threshold that works effectively across input images, despite variations in their original shadow intensities.

%The reason we use the strengthened shadow at the maximum $c$ value instead of the original input image for the shadow difference map computation is due to the varying shadow intensities in the input images. This variation makes it difficult to find a single optimal threshold value across all images in the dataset. Using the maximum $c$ version ensures that all images are consistently strengthened, attenuated, and normalized to have consistent maximum and minimum shadow levels.

% In practice, finding a single optimal threshold value across all images in the dataset is challenging. We propose using the strengthened shadow at the maximum $c$ version instead of the original input image for the shadow difference map computation, as the input image have varying shadow intensities. This ensures that all images are re-shadowed and normalized to have consistent maximum and minimum shadow levels.

Next, we apply a median filter to remove small, isolated shadow regions from the thresholded shadow map. 
Finally, we multiply this map with a normalized shadow value ($c$) of the input image, resulting in a $c$-scaled shadow map that also encodes the shadow intensity.
\subsubsection{Training DiFaReli++ with cast shadows} We concatenate the precomputed shadow map and the shading reference, feed them into the Modulator network, and follow DiFaReli's training procedure.

\subsubsection{Inferencing DiFaReli++ with cast shadows}
% At test time, since we do not have access to the new shadow difference map for the same person, we can generate plausible cast shadow effects by computing this map using ray tracing. Notably, using ray tracing at this stage does not introduce misalignment issues, as during inference, we only need the shadow map to guide where shadows should be generated in the relit image. In contrast, misalignment during the training stage is much more severe, as the diffusion model could completely ignore the condition due to its noisiness.
Once our network learns to associate the estimated shadow map with the input image, we can guide it to generate new cast shadow effects using ray tracing. 
% Note that, using ray tracing at this stage does not introduce misalignment issues, as during inference, we only need the shadow map to guide where shadows should be generated in the relit image.
Specifically, we perform ray tracing on the estimated depth map from DECA, using the dominant light direction from SH, as described in \cite{hou2022face}. 
% This shadow map is then applied during the relighting process, similar to DiFaReli (see Section \ref{sec:relighting}). 
We follow a similar relighting process to DiFaReli (see Section \ref{sec:relighting}), using the precomputed shadow map for inversion and the ray-traced shadow map for generation. 
Note that if we had used the ray-traced shadow map for inversion, this would result in over-brightening artifacts and the failure to remove certain shadowed areas 
(Figure 46 in the Appendix). 
% (Figure \ref{fig_app:overbright_results} in the Appendix).
%We hypothesize that the network incorrectly identifies non-existent shadow positions in the input image, leading to the unintended brightening of those regions during relighting.
This suggests that the network may misinterpret shadow areas in the input image, resulting in unintended brightening during relighting.

% We found that using the ray-traced shadow map for inversion often resulted in over-brigthening artifacts and some shadows area could not be removed properly (as shown in Figure \ref{fig_app:overbright_results} in Appendix). Our assumption is the network takes wrong shadow position that not exists in the input image and trying to brithen that region during relighting resulting in over brightening those areas.
% This is because inaccurate shadow areas caused the network to brighten regions without shadows, resulting in over-brightening those areas (Figure \ref{fig_app:overbright_results} in Appendix).
% We use the using the precomputed shadow map for inversion produced better qualitative results and eliminated some artifacts. Since
% \reviseded{
% \subsection{DiFaReli$_\text{nf}$: Relighting of non-facial parts.}
\subsection{Relighting of non-facial parts.} \label{sec:relit_bg}
%This improvement aims to relight other parts such as clothes, earrings, and hats
%Another goal is to expand the relightable area of DiFaReli (the face, neck, and hair) to include non-facial parts belonging to the person such as clothes, earrings, and hats, but not the rest of the background. 
Another goal is to expand the relightable area of DiFaReli beyond facial parts (e.g., face, neck, and hair) to other areas of the person, such as clothes, earrings, and hats, but not the rest of the background. 
In the original DiFaReli, these non-facial parts remain unchanged because the network is conditioned on the background encoding $(\vect{bg})$, which contains their raw pixels.
However, simply removing this condition can lead to structural changes in those parts, as shown in Figure \ref{fig:bg_ab}, while still not allowing us to relight them.

\begin{figure}[ht!]
\centering
  \includegraphics[scale=1]{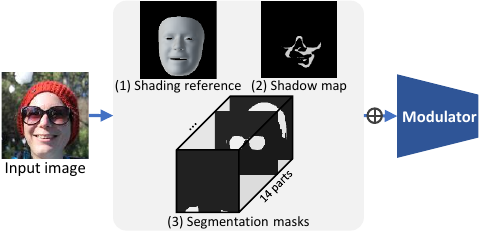}
  \caption{\textbf{Modifications of the Modulator’s input in DiFaReli++}. The input is a concatenation of the shadow map, the shading reference, and segmentation masks 
  (see all masks in Figure 47). 
  % (see all masks in Figure \ref{fig_app:seg_mask}).
  This modification allows generation of consistent cast shadows and enables relighting of non-facial parts.
  % This modification allows control over cast shadows and enables relighting of non-facial parts.
  }
  \vspace{-1.5em}
  \label{fig:new_modulator_input}
\end{figure}

To address this, we condition the network with a set of segmentation masks instead of the background encoding. This approach makes the conditioning less restrictive and still enables relighting while preserving the structure of each part. These masks are computed using Bisenet face parsing \cite{yu2018bisenet} and correspond to facial parts such as the eyes, nose, mouth, as well as non-facial parts such as the hair and neck, covering 14 distinct areas 
(Figure 47).
% (Figure \ref{fig_app:seg_mask}).
%The segmentation mask was computed from face parsing \cite{yu2018bisenet}. Specifically, the segmentation masks capture parts (e.g., eyeglasses and eyeballs) to prevent them from moving or disappearing during relighting. 
Finally, we feed a concatenation of these masks, a shadow map, and a shading reference as input to the Modulator network (See Figure \ref{fig:new_modulator_input}).

% \reviseded{
\subsection{DiFaReli++$_\text{ss}$: Single-shot face relighting network} 
\label{single_shot_inf}

\begin{figure*}[]
\centering
  \includegraphics[width=\textwidth]{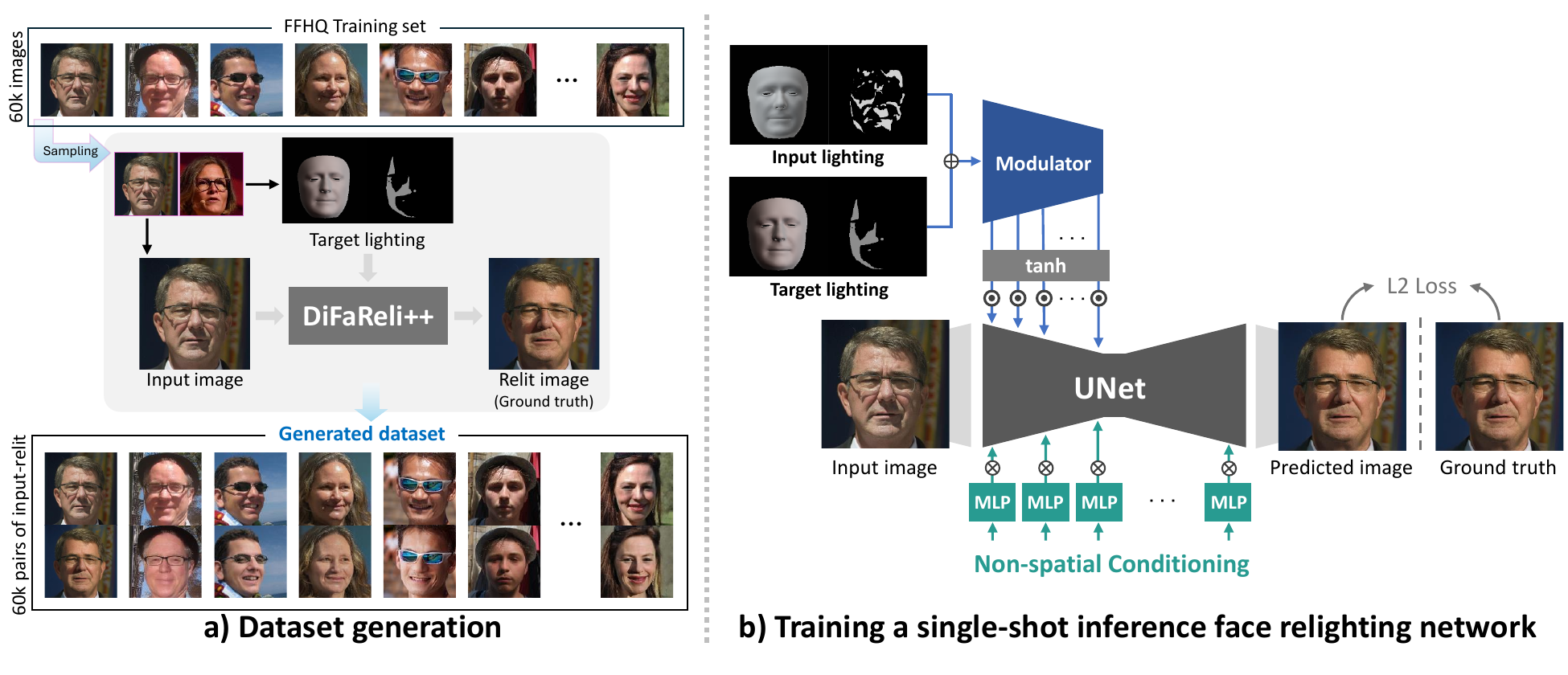}
  
  \caption{\textbf{Single-shot face relighting framework} involves a) using DiFaReli++ to generate supervised relit pairs and b) training a single-shot relighting network with the same architecture as DiFaReli++ using the training pairs with a simple L2 loss.
  %\textbf{(a)} dataset generation and \textbf{(b)} training process of a single-shot inference relighting network.
  }
  \label{fig:mod_difareli}
  \vspace{-1em}
\end{figure*}

%As the state-of-the-art in face relighting, DiFaReli \cite{ponglertnapakorn2023difareli} currently requires approximately 3 minutes to relight a single person's face. 
Currently, both DiFaReli \cite{ponglertnapakorn2023difareli} and DiFaReli++ with controllable cast shadows take approximately 3 minutes to relight a single input image.
To improve speed, the idea is to use our pretrained DiFaReli++ to generate synthetic relit pairs and use them to train a relighting network in a supervised manner. 
% This network adopts the same architecture as DiFaReli++ and requires only a single network pass to relight an input image given required pre-processed data (e.g., DECA and shadow map), without relying on diffusion sampling or DDIM inversion.
This network adopts the same architecture as DiFaReli++ and requires only a single network pass to relight an input image, given the required pre-processed data (e.g., DECA and shadow map), without relying on diffusion sampling or DDIM inversion. As in DiFaReli and DiFaReli++, this pre-processed data needs to be computed only once per input face image and can be reused for each relighting process.
Surprisingly, the resulting single-shot network, referred to as DiFaReli++$_\text{ss}$, is not only three orders of magnitude faster but also achieves better scores across all metrics than DiFaReli++, which was used to generate the synthetic training data
%than while requiring only a single network pass, as opposed to 10-1,000 passes 
(Figure \ref{fig:runtime}).
We next describe the dataset generation and training process (see Figure \ref{fig:mod_difareli} for an overview).
\subsubsection{Dataset generation}\label{sec:gen_ss}
The training set for DiFaReli++$_\text{ss}$ was created by uniformly sampling pairs of face images from all 60,000 images of the FFHQ training set. In each pair, we use one face image as input and use the lighting condition from the other image as the target lighting. The lighting condition is estimated with DECA \cite{DECA:Siggraph2021}. 
%We render the shading reference and compute the target shadow map through ray tracing under the same target lighting. 
Then, we relight the input under the target lighting using DiFaReli++, resulting in 60,000 pairs of (input, relit input) along with their corresponding shading references and shadow maps.
\subsubsection{Training}\label{sec:training_ss}
DiFaReli++$_\text{ss}$ uses a similar architecture to DiFaReli, including the Modulator network, except that the Modulator takes two sets of inputs (8 channels in total), each comprising a shading reference (3 channels) and an estimated shadow map scaled by $c$ (1 channel) of the input image. The first set uses the input's own lighting condition, while the second set corresponds to the target lighting condition. %The additional set that 
%The Modulator network takes two input sets (8 channels total), each comprising a shading reference (3 channels) and a shadow map (1 channel): one set rendered under input lighting and the other under target lighting.
%resulting in a total of 8 channels.
%: one set rendered under input lighting and the other under target lighting.
%These shading references and shadow maps are used to condition the network, enabling the transfer of lighting and cast shadow positions from the input to the target. 
We train the network using a simple L2 loss between the predicted image and the synthetic ``ground-truth'' image. 
\subsubsection{Relighting}\label{sec:inference_ss} We feed an input image and its encodings to the network along with the two sets of shading references and shadow maps for the Modulator, following the same input preprocessing as in training.

  \begin{figure*}[]
  \centering
    \includegraphics[scale=0.64]{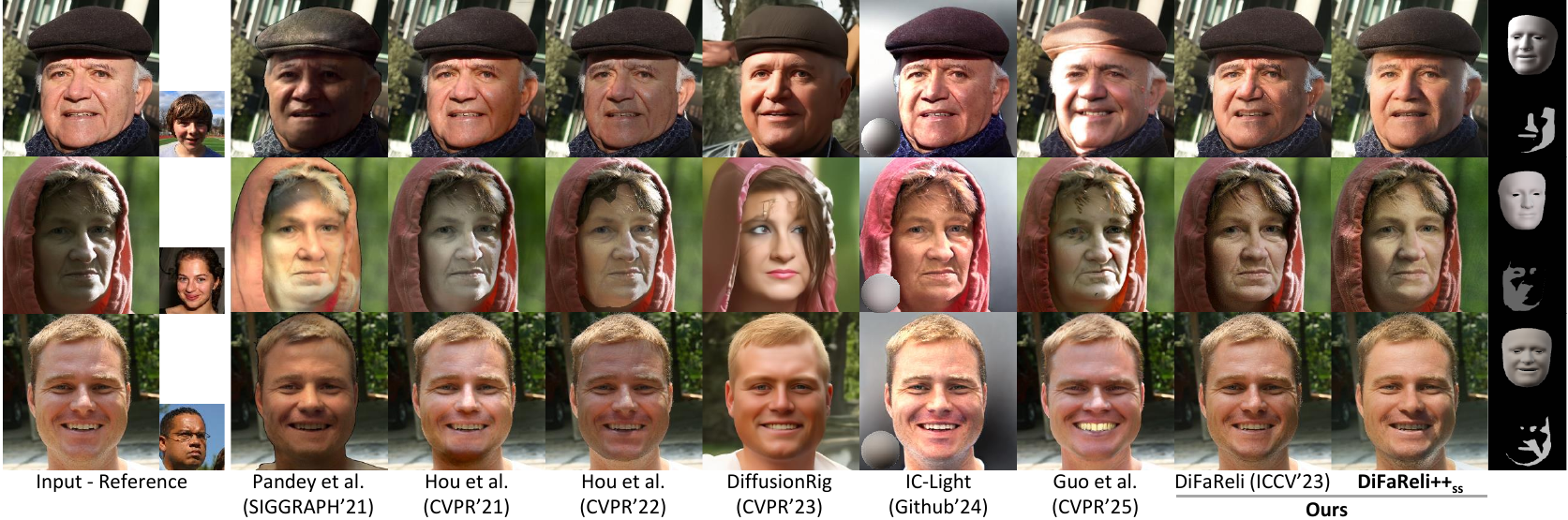}

  \caption{\textbf{Relit results on FFHQ \cite{karras2019style}}, a dataset of diverse face images captured in real-world conditions. Our method produces more realistic shading and controllable cast shadows (via the shadow map in the rightmost column), effectively removing existing shadows and adding new ones. It also relights non-facial parts (e.g., hats, hoodies, or shirts) to match the target lighting, enhancing overall realism.
  Unlike DiFaReli++, DiffusionRig\cite{ding2023diffusionrig} fails to preserve identity, as it relies on subject-specific fine-tuning on a personal photo collection, making it unsuitable for single-image relighting. For \cite{Guo_2025_CVPR, ding2023diffusionrig}, we also performed a grid search over their relighting hyperparameters to verify that the oversaturation artifacts were not caused by suboptimal tuning 
  % (see Figures~\ref{fig_app:tuninggrid_relipa_targetSH_sj1}-\ref{fig_app:tuninggrid_diffusionrig_targetSH}).
  (see Figures~40-43).
   Additional results are in Appendix (Figures~30, 31, 32, and 33).}
   % Additional results are in Appendix (Figures \ref{fig_app:ffhq_cs_app1}, \ref{fig_app:ffhq_cs_app2}, \ref{fig_app:ffhq_cs_app3}, and \ref{fig_app:ffhq_cs_app4}).}
  \label{fig:ffhq_cs}
  \vspace{-1em}
\end{figure*}

\begin{figure*}
  \centering
  % \resizebox{\textwidth}{!}{%
        % \includegraphics{figures/cs_mp_combined.pdf}}
    % \includegraphics[scale=0.57]{figures/cs_mp_combined_cut_baseline.pdf}
    % \includegraphics[scale=0.57]{figures/cs_mp_combined_cut_baseline_single.pdf}
    \includegraphics[scale=0.51]{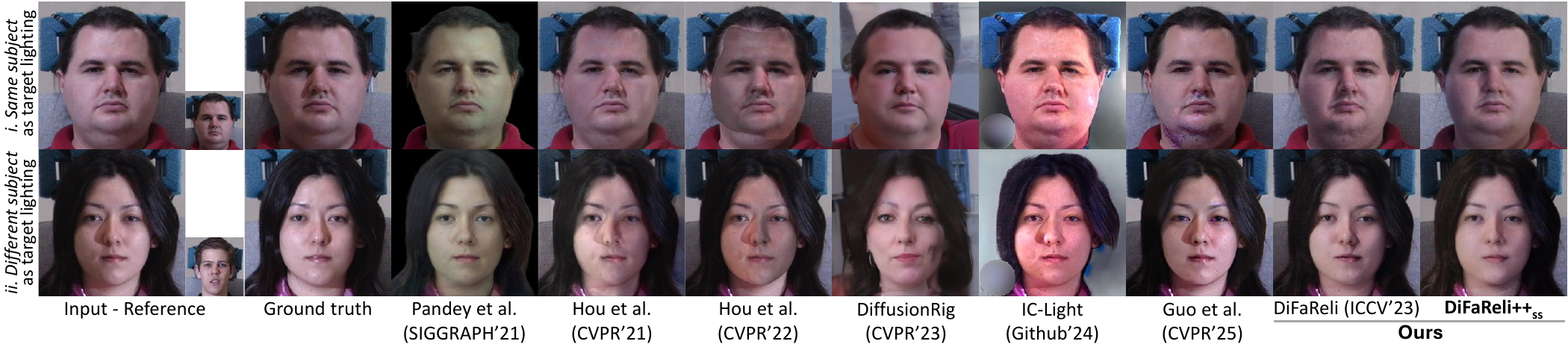}

  % \vspace{-1em}
  \caption{\textbf{Relighting results on Multi-PIE~ \cite{gross2010multi}} when the target lighting is taken from the same person (first row) and from a different person (second row).}
  \vspace{-1.5em}
  \label{fig:mutlipie_cs}
\end{figure*}

\begin{figure}
  \centering
  \includegraphics[scale=0.44]{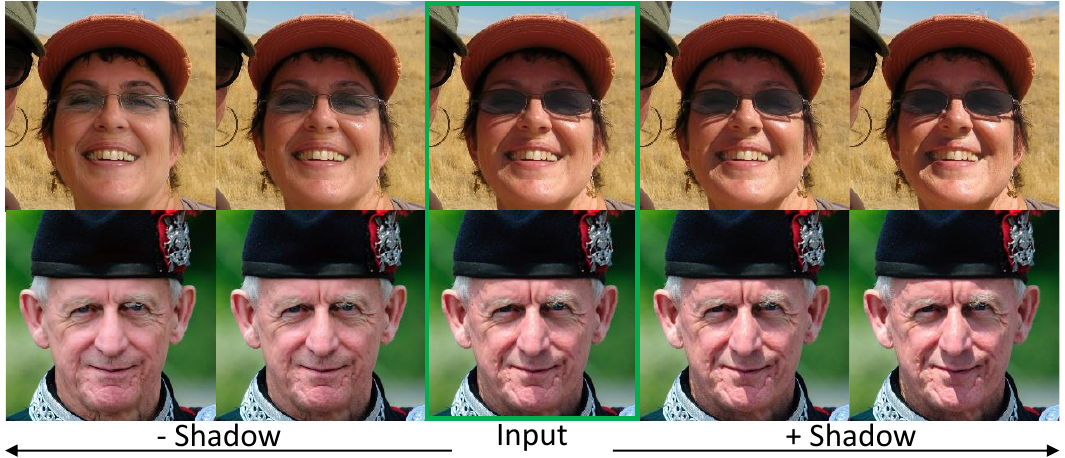}
  \caption{\textbf{Varying intensities of cast shadows.} DiFaReli's ability to change the intensity of cast shadows by adjusting the scalar $c$ and decode the modified feature vector 
  (more in Figure 45).
  % (more in Figure \ref{fig:shad}).
  }
  \vspace{-1em}
  \label{fig:cast_shadow}
\end{figure}

\begin{figure*}
  \centering
  \adjustbox{center}{
    \includegraphics[scale=0.43]{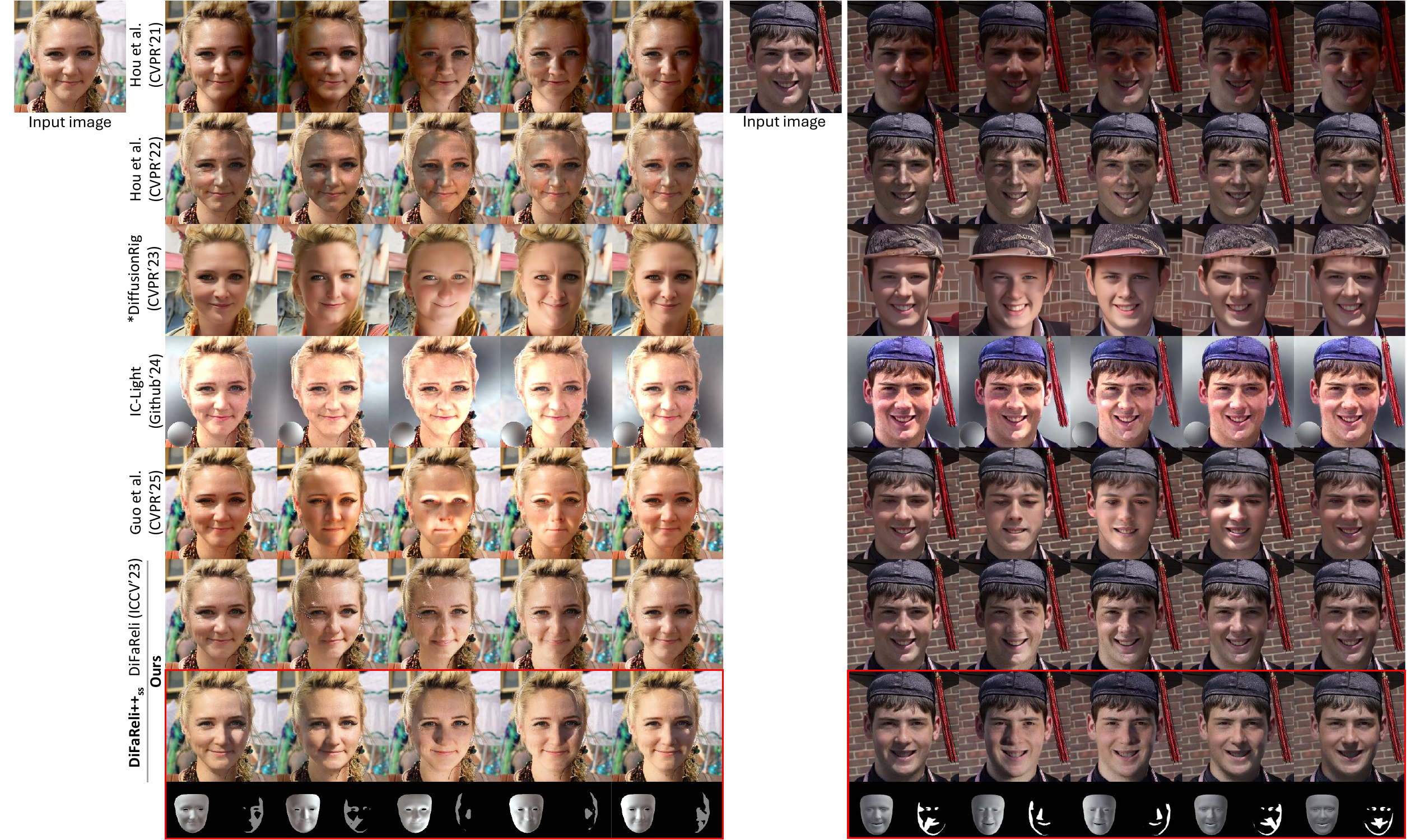}
    }

    \caption{\textbf{Relighting with consistent cast shadows under moving lights.} 
    Compared to six recent state-of-the-art methods \cite{iclight, hou2021towards, hou2022face, ding2023diffusionrig, ponglertnapakorn2023difareli, Guo_2025_CVPR}, DiFaReli++ effectively removes input cast shadows and synthesizes new ones in a realistic and consistent manner. $^*$DiffusionRig \cite{ding2023diffusionrig} was designed for fine-tuning on a personal photo collection, but since our single-image relighting setup lacks such data, its results are generated without fine-tuning, leading to identity shift issues.
    %'s results here are produced without fine-tuning the model on a personal photo collection, as such data is unavailable in our single-image relighting setup. 
    %addresses a different problem setup, requiring a photo album of the same subject ($\approx$20 images) to finetune a personalized model. 
    %As a result, it fails to preserve the person’s identity in single-image relighting. 
    We observed oversaturation in \cite{Guo_2025_CVPR, ding2023diffusionrig}, and performed a hyperparameter grid search to ensure it was not due to suboptimal tuning 
    (see Figures~36-39). 
    % (see Figures~\ref{fig_app:tuninggrid_relipa_rot_sj1}-\ref{fig_app:tuninggrid_diffusionrig_rot_sj2}). 
    The bottom row shows our shading references and shadow maps. Additional results are provided in the Appendix 
    (Figures~25, 26, 27, 28, and 29).}
    % (Figures~\ref{fig:rotate_cs_app_1}, \ref{fig:rotate_cs_app_2}, \ref{fig:rotate_cs_app_3}, \ref{fig:rotate_cs_app_4}, and \ref{fig:rotate_cs_app_5}).}

  \label{fig:rotate_cs}
  \vspace{-1em}
\end{figure*}

\begin{figure*}
  \centering
  \adjustbox{center}{

    \includegraphics[scale=0.63]{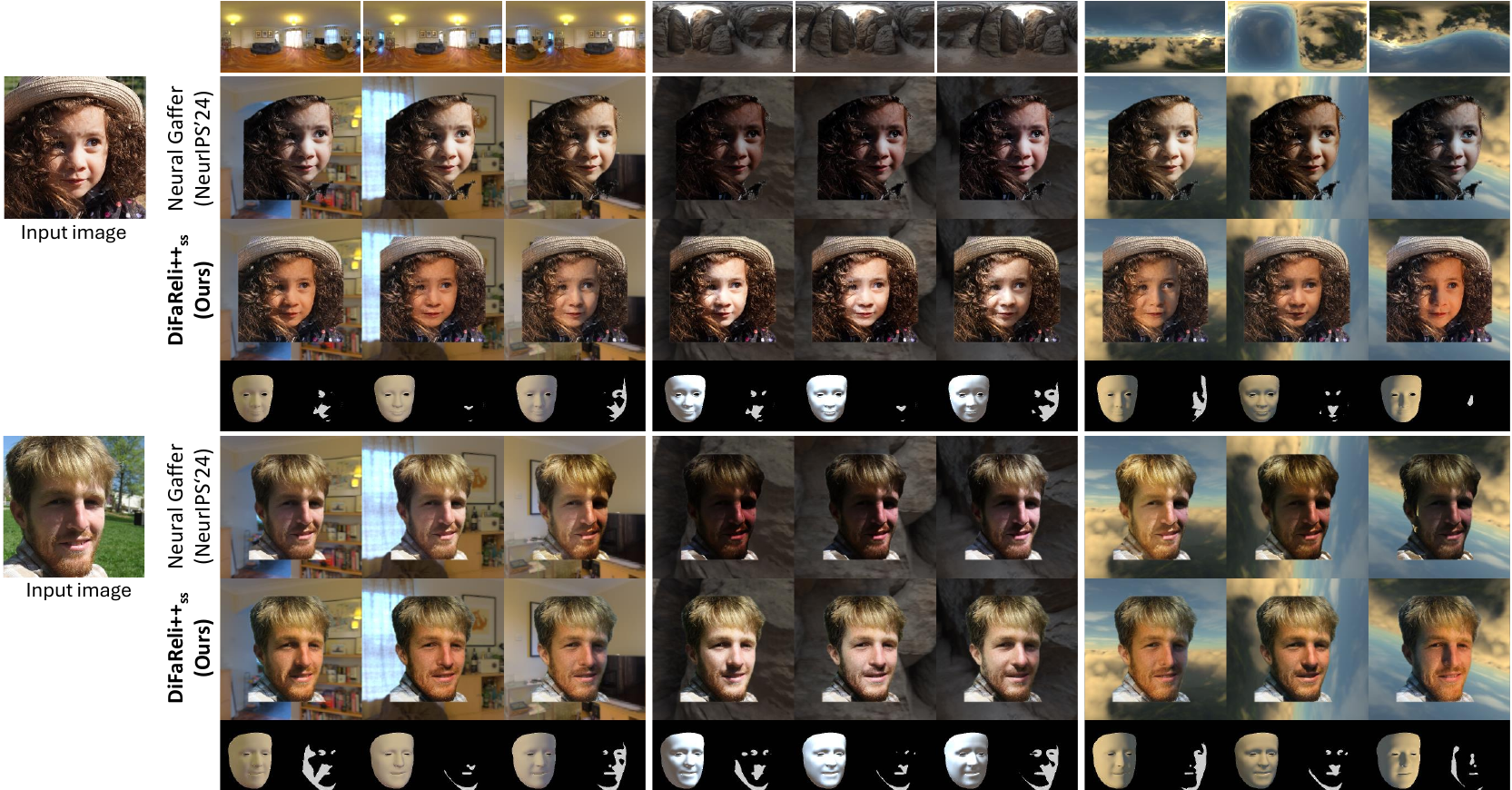}
    
    }
%that improves upon prior approaches, such as DiLightNet~\cite{zeng2024dilightnet}, 
    \caption{\textbf{Relighting comparison with a state-of-the art object-relighting method, Neural Gaffer \cite{jin2024neural_gaffer}.}
%    We compare Neural Gaffer, a state-of-the-art object-relighting method, with our DiFaReli++ on the face-relighting task. 
    The top row shows input HDR maps, and below the results are our shading references and shadow maps. Following Neural Gaffer, the background is composited only for visualization.
    %Neither method performs background harmonization. The background was rendered only for visualization. As a result, Neural Gaffer can brighten some regions but often fails to remove cast shadows, whereas DiFaReli++ effectively eliminates them and synthesizes consistent new shadows. 
   Neural Gaffer can brighten some regions but often fails to remove cast shadows, while DiFaReli++ effectively eliminates them and synthesizes consistent new ones. Fine-tuning Neural Gaffer for faces requires multi-view light stage data, which is hard to obtain and scale.
    % More results are in Figures~\ref{fig_app:ffhq_cs_hdr_app1} and \ref{fig_app:ffhq_cs_hdr_app2} in the Appendix.
    More results are in Figures~34 and 35 in the Appendix.
    }
  \label{fig:rotate_hdr}
  \vspace{-1em}
\end{figure*}

\begin{figure*}
  \centering
  \adjustbox{center}{
    \includegraphics[scale=0.5]{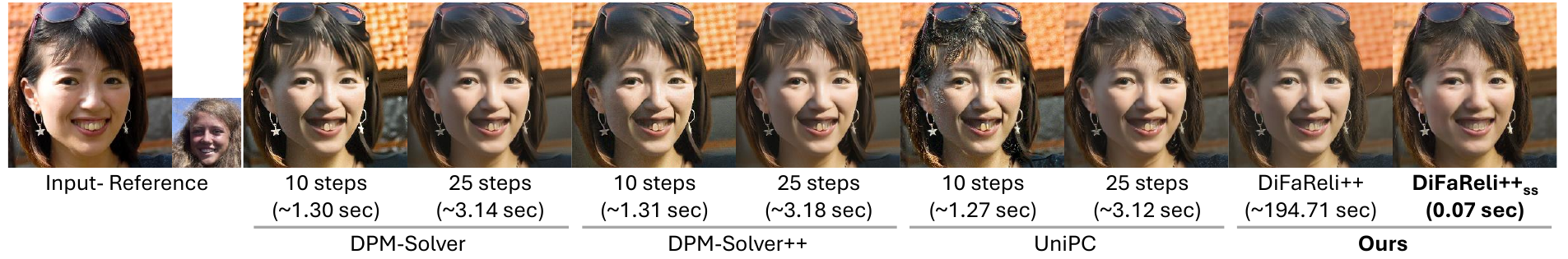}
  }

  \caption{
  \textbf{Results using various acceleration techniques.} with different sampling steps on an input image from FFHQ.
  While these techniques can reduce the sampling steps, they introduce artifacts and blurriness. 
  In contrast, our distilled version of DiFaReli++ (DiFaReli++$_\text{ss}$) delivers the highest quality, the least noisy output, and runs in just 0.07 seconds.
  }
  \label{fig:speedup}
  \vspace{-1em}
\end{figure*}

\section{Experiments}
In this section, we present quantitative and qualitative results of our proposed method. We provide a comparison of our relighting performance with state-of-the-art methods on the Multi-PIE dataset (Section \ref{sec:LP}), a runtime comparison (Section \ref{sec:runtime_impr}), a user study on the FFHQ dataset (Section \ref{sec:user_study}), and ablation studies on the conditioning mechanisms (Section \ref{sec:AS}). Implementation and dataset details are in Appendix 
II.
% \ref{app:impl}.

\textbf{Evaluation metrics.}
We use DSSIM \cite{nestmeyer2020learning}, LPIPS \cite{zhang2018unreasonable}, and MSE. DSSIM measures structural dissimilarity, while LPIPS measures perceptual quality. All metrics are computed between each relit image and its ground-truth image only on the face region, following \cite{hou2022face, hou2021towards} using the same face parsing algorithm \cite{yu2018bisenet}.
%This face region is computed using  parsing \cite{yu2018bisenet}, used in \cite{hou2022face}, to retain only the face region.

% \vspace{-0.5em}
\subsection{Relighting performance}
\label{sec:LP}
We evaluate our relighting performance on the Multi-PIE dataset \cite{gross2010multi} against recent state-of-the-art methods \cite{hou2021towards, hou2022face, nestmeyer2020learning, pandey2021total, ding2023diffusionrig, iclight, ponglertnapakorn2023difareli, Guo_2025_CVPR}. 
Note that several prior works \cite{pandey2021total, iclight, ding2023diffusionrig, ren2023relightful, mei2024holo, Guo_2025_CVPR} tackle different problem setups: \cite{ding2023diffusionrig} requires a photo album of the same subject ($\approx${20} images) to fine-tune a personalized model, which our setting lacks. \cite{Guo_2025_CVPR} aims to animate and relight a single portrait image given target keypoints and lighting.
Other methods \cite{pandey2021total, mei2024holo, kim2024switchlight} require an HDR environment map as input, which must first be estimated from a target image, while \cite{iclight, ren2023relightful} require a background image as input. These make comparisons with \cite{pandey2021total, mei2024holo, ding2023diffusionrig, iclight, ren2023relightful, Guo_2025_CVPR} not entirely apples-to-apples.

In our experiments, we used the pretrained Stage 1 model from DiffusionRig~\cite{ding2023diffusionrig} without any finetuning, as the fine-tuned model (Stage 2) requires a subject-specific photo collection, which was not available in our single-image setting. Additionally, the authors of~\cite{ding2023diffusionrig} show that fine-tuning on a single image struggles to adapt to new lighting conditions (see Figure 7 in~\cite{ding2023diffusionrig}). For the relightable keypoint-based portrait animation technique~\cite{Guo_2025_CVPR}, we modified the input by using the keypoints of the input image as the target, which disables portrait animation and thus limits control to relighting only 
(see Figure~49 in the Appendix for the modified input).
% (see Figure~\ref{fig_app:relipa_input} in the Appendix for the modified input).

We also provided each method with its required target lighting. Pandey et al. \cite{pandey2021total} used their own light estimator to generate a target HDR map from a target image.
Zhang et al. \cite{iclight} do not include a light estimator and instead require a target background. Thus, we provided a proxy background as the target lighting, created by unwrapping a rendered sphere under the target SH lighting. Examples of the proxy backgrounds are shown in 
Figure~48 (Appendix). 
% Figure \ref{fig_app:proxy_lighting} (Appendix). 
All results for \cite{pandey2021total} were generated by the authors themselves, and the results for \cite{iclight} were generated using their official code. Test sets and code of \cite{mei2024holo, ren2023relightful, kim2024switchlight} were not released
% \footnote{We reached out to the authors of \cite{kim2024switchlight, ren2023relightful, mei2024holo}, but were unable to obtain their results even after more than two months. Note that these methods also tackle different problem setups.}. 
Instead, we provide a qualitative comparison with \cite{mei2024holo, kim2024switchlight} using their input images, target lighting, and results cropped directly from their papers 
% (see Section \ref{sec_app:compare_more_hdr} for details and results).
(see Section~VI-A for details and results).
% The results of \cite{pandey2021total, mei2024holo, ren2023relightful} in our experiment were generated by the authors themselves. For Pandey et al. \cite{pandey2021total} they used their own light estimator estimate target HDR maps. In case of Mei et al. \cite{mei2024holo, ren2023relightful, iclight} has no light estimators, we unwrapped the sphere of rendered SH under target lighting and give them to use as the background or environment map. We show examples of this enviroment amaps from SH in Figure \ref{fig_app:ball_to_env} (Appendix).
% Other test sets and code of \cite{pandey2021total, mei2024holo, ren2023relightful} were not released.

% Note that Pandey et al. \cite{pandey2021total} and Mei et al. \cite{mei2024holo} solve a different problem setup (also \cite{wang2020single, yeh2022learning}) and require an HDR environment map as input, which has to be first estimated from a target image.

% These making a comparison with \cite{pandey2021total, mei2024holo, iclight, ren2023relightful} not entirely apples-to-apples. 
% The results of \cite{pandey2021total, mei2024holo, ren2023relightful} in our experiment were generated by the authors themselves, including the HDR maps. Other test sets and code of \cite{pandey2021total} were not released.
 
 Our experiment has two setups for the target lighting: 
 
 \textbf{i). From the same person.} This setup uses the same test set as \cite{hou2021towards}, which contains 862 testing samples from 329 subjects.
 
 \textbf{ii). From a different person.} This setup contains 200 random triplets of input, target, and ground-truth images, where the target image is of a different person.

The results are shown in Table \ref{tab:TL} and Figure \ref{fig:mutlipie_cs}.
For both setups, our method achieves the best performance across all metrics with minimal artifacts and convincingly relights areas like the neck and ears or removes cast shadows, such as the shadow from the nose of the lady in Figure \ref{fig:mutlipie_cs}. 
We include a comparison with \cite{sengupta2018sfsnet, zhou2019deep, sun2019single} 
% in Appendix \ref{app:more_results}.
in Appendix VI.

%\revised{In addition, we compare our method with \cite{pandey2021total}, which represents HDR-based methods \cite{wang2020single, pandey2021total, yeh2022learning}. Note that these methods solve different problem setups that require both an HDR environment map and light stage data. As shown in Appendix \ref{app:more_results}, using these methods may result in an off-color skin tone.}

%We achieve the lowest errors across all metrics and produce minimal artifacts.
% The input and target images are from the same subject. 

% The reference image was sampled to have the same light as the target image but from a different subject. 
 % The error metrics were computed between the input image and the target image.

% The input and target image was sampled from the same subject under different light conditions

% , and the reference image was under the same light condition as target image, but from a difference subject.
% were sampled from a different subject with the same light as target image.

% \begin{table}[!htpb]
% \resizebox{1\columnwidth}{!}{%
% \begin{tabular}{l|ccc}
% \toprule
% \multicolumn{1}{l|}{Method} & DDSIM$\downarrow$ & MSE$\downarrow$ & LPIPS$\downarrow$ \\ \midrule
% Hou et al.(CVPR'21)\cite{hou2021towards} & 0.1056 & 0.0247 & 0.1989 \\
% Hou et al.(CVPR'22)\cite{hou2022face} & 0.1150 & 0.0238 & 0.2215 \\
% \textbf{Ours} & \textbf{0.0969} & \textbf{0.0215} & \textbf{0.1669} \\ \bottomrule
% \end{tabular}}
% \caption{Quantitative results on Multi-PIE (Different subject as target lighting)}
% \label{tab:DTL}
% \end{table}

\begin{table}[!htpb]
\caption{State-of-the-art comparison on Multi-PIE.}
\vspace{-0.2cm}
\resizebox{1\columnwidth}{!}{%
\setlength{\tabcolsep}{5pt}
\small
\begin{tabular}{llll}
\toprule
\multicolumn{1}{l}{Method} & \multicolumn{1}{c}{DDSIM$\downarrow$} & \multicolumn{1}{c}{MSE$\downarrow$} & \multicolumn{1}{c}{LPIPS$\downarrow$} \\ \midrule
\multicolumn{4}{l}{\cellcolor[HTML]{EFEFEF}\textbf{i). Same subject as target lighting}} \\
\multicolumn{1}{l}{\ \ \ Nestmayer et al. \cite{nestmeyer2020learning}} & \multicolumn{1}{c}{0.2226} & \multicolumn{1}{c}{0.0588} & {0.3795} \\
\multicolumn{1}{l}{\ \ \ Pandey et al. (Total Relighting) \cite{pandey2021total}} & \multicolumn{1}{c}{0.0875} & \multicolumn{1}{c}{0.0165} & {0.2010} \\
\multicolumn{1}{l}{\ \ \ Hou et al. \cite{hou2021towards}} & \multicolumn{1}{c}{0.1186} & \multicolumn{1}{c}{0.0303} & {0.2013} \\
\multicolumn{1}{l}{\ \ \ Hou et al.  \cite{hou2022face}} & \multicolumn{1}{c}{0.0990} & \multicolumn{1}{c}{0.0150} & {0.1622} \\
\multicolumn{1}{l}{\ \ \ Ding et al. (DiffusionRig) \cite{ding2023diffusionrig}} & \multicolumn{1}{c}{0.0870} & \multicolumn{1}{c}{0.0098} & {0.2098} \\
\multicolumn{1}{l}{\ \ \ Zhang et al. (IC-Light) \cite{iclight}} & \multicolumn{1}{c}{0.1978} & \multicolumn{1}{c}{0.0499} & {0.1887} \\
\multicolumn{1}{l}{\ \ \ Guo et al. \cite{Guo_2025_CVPR}} & \multicolumn{1}{c}{0.1088} & \multicolumn{1}{c}{0.0234} & {0.1733} \\
\multicolumn{1}{l}{\ \ \ {Ours (DiFaReli)}} & \multicolumn{1}{c}{{0.0711}} & \multicolumn{1}{c}{{0.0122}} & {0.1370}\\
% \multicolumn{1}{l}{\ \ \ \textbf{Ours (DiFaReli++)}} & \multicolumn{1}{c}{\textbf{0.0604}} & \multicolumn{1}{c}{\textbf{0.0090}} & \textbf{0.1043}\\
\multicolumn{1}{l}{\ \ \ Ours (DiFaReli++)} & \multicolumn{1}{c}{0.0604} & \multicolumn{1}{c}{0.0090} & {0.1043}\\
\multicolumn{1}{l}{\ \ \ \textbf{Ours (DiFaReli++$_\text{ss}$)}} & \multicolumn{1}{c}{\textbf{0.0590}} & \multicolumn{1}{c}{\textbf{0.0075}} & \textbf{0.1023}
%\bottomrule
\vspace{0.5em} \\
\multicolumn{4}{l}{\cellcolor[HTML]{EFEFEF}\textbf{ii). Different subject as target lighting}} \\
\multicolumn{1}{l}{\ \ \ Pandey et al. (Total Relighting) \cite{pandey2021total}} & \multicolumn{1}{c}{0.1000} & \multicolumn{1}{c}{0.0252} & {0.2053} \\
\multicolumn{1}{l}{\ \ \ Hou et al. \cite{hou2021towards}} & \multicolumn{1}{c}{0.1056} & \multicolumn{1}{c}{0.0247} & {0.1989} \\
\multicolumn{1}{l}{\ \ \ Hou et al. \cite{hou2022face}} & \multicolumn{1}{c}{0.1150} & \multicolumn{1}{c}{0.0238} & {0.2215} \\
\multicolumn{1}{l}{\ \ \ Ding et al. (DiffusionRig) \cite{ding2023diffusionrig}} & \multicolumn{1}{c}{0.1130} & \multicolumn{1}{c}{0.0180} & {0.2292} \\
\multicolumn{1}{l}{\ \ \ Zhang et al. (IC-Light) \cite{iclight}} & \multicolumn{1}{c}{0.2239} & \multicolumn{1}{c}{0.0654} & {0.2102} \\
% \multicolumn{1}{l}{\ \ \ Ren et al. (CVPR'24) \cite{ren2023relightful}} & \multicolumn{1}{c}{} & \multicolumn{1}{c}{} &  \\
% \multicolumn{1}{l}{\ \ \ Kim et al. (CVPR'24) \cite{kim2024switchlight}} & \multicolumn{1}{c}{} & \multicolumn{1}{c}{} &  \\
% \multicolumn{1}{l}{\ \ \ Mei et al. (CVPR'24) \cite{mei2024holo}} & \multicolumn{1}{c}{} & \multicolumn{1}{c}{} &  \\
\multicolumn{1}{l}{\ \ \ Guo et al. \cite{Guo_2025_CVPR}} & \multicolumn{1}{c}{0.1230} & \multicolumn{1}{c}{0.0255} & {0.1869} \\
\multicolumn{1}{l}{\ \ \ {Ours (DiFaReli)}} & \multicolumn{1}{c}{{0.0969}} & \multicolumn{1}{c}{{0.0215}} & {0.1669} \\
% \multicolumn{1}{l}{\ \ \ \textbf{Ours (DiFaReli++)}} & \multicolumn{1}{c}{\textbf{0.0824}} & \multicolumn{1}{c}{\textbf{0.0169}} & \textbf{0.1223} \\
\multicolumn{1}{l}{\ \ \ Ours (DiFaReli++)} & \multicolumn{1}{c}{0.0824} & \multicolumn{1}{c}{0.0169} & {0.1223} \\
\multicolumn{1}{l}{\ \ \ \textbf{Ours (DiFaReli++$_\text{ss}$)}} & \multicolumn{1}{c}{\textbf{0.0801}} & \multicolumn{1}{c}{\textbf{0.0148}} & \textbf{0.1123} \\
\bottomrule
\end{tabular}}
\label{tab:TL}
\vspace{-2em}
\end{table}

\subsection{Runtime improvement}
\label{sec:runtime_impr}
\begin{figure*}
\centering
    \adjustbox{center}{
    \includegraphics[width=2\columnwidth]{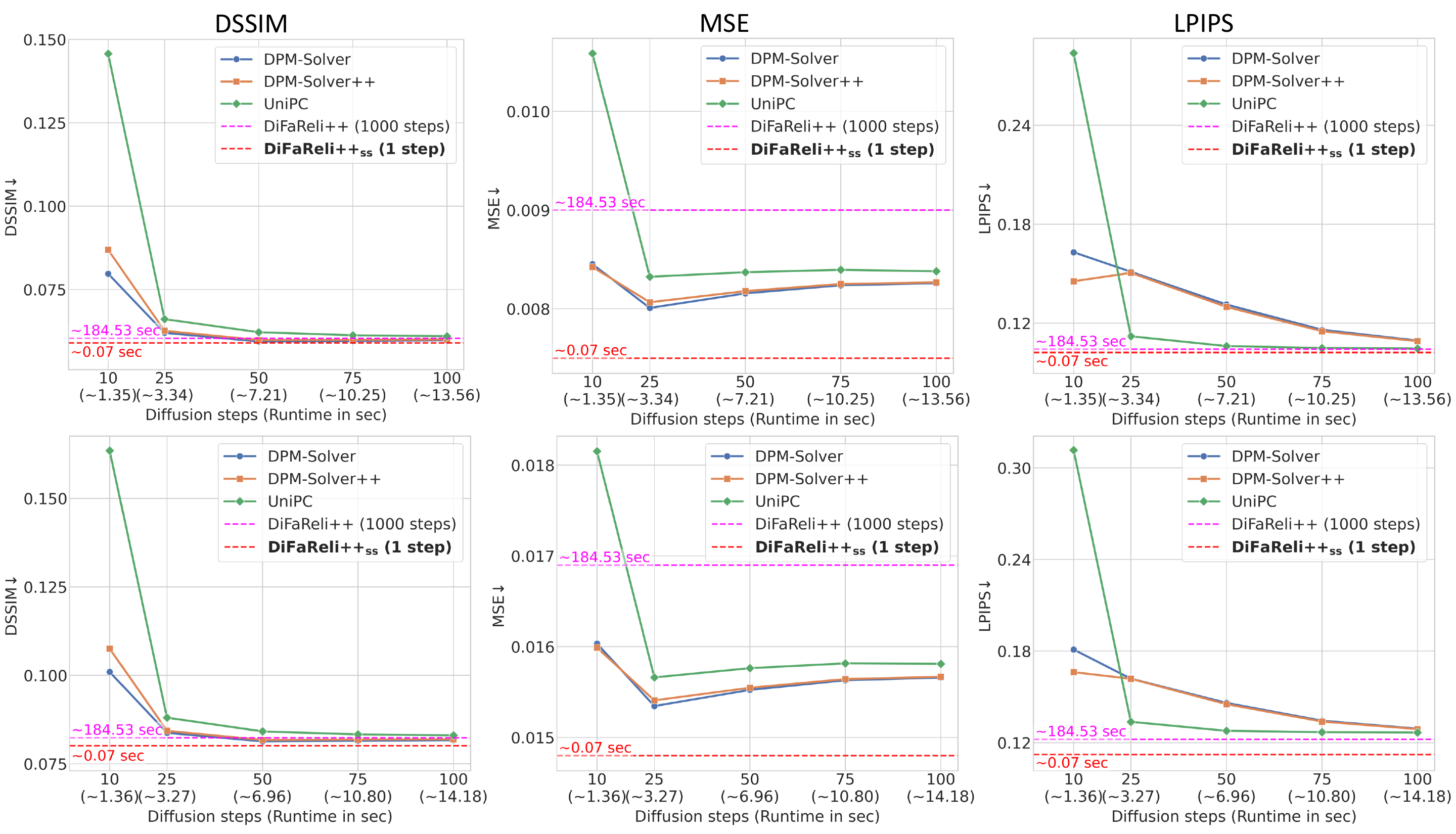}
  }
  % \vspace{-0.5cm}
  \caption{\textbf{Trade-off between runtime and relighting performance of different acceleration techniques} measured on three metrics: DSSIM, MSE, and LPIPS. The \textbf{first row} shows results on the test set where the target lighting is taken from the same subject, while the \textbf{second row} uses target lighting from a different subject. The red dashed line represents our single-shot face relighting score (DiFaReli++$_\text{ss}$), and the magenta dashed line represents our original DiFaReli++ score.}  
  \label{fig:runtime}
  \vspace{-1.0em}
\end{figure*}

We compare our single-shot network against three training-free diffusion acceleration techniques: DPM-Solver \cite{lu2022dpm}, DPM-Solver++ \cite{lu2022dpm++}, and UniPC \cite{zhao2023unipc}, each offering a variable trade-off between sampling speed and quality. We apply these techniques to our pretrained DiFaReli++ model and report DSSIM, MSE, and LPIPS scores for various numbers of sampling steps in Figure \ref{fig:runtime}. All solvers use a multistep schedule and their orders are set to 2.

We measured the time required for network inference alone, assuming all one-time pre-processed data (e.g., DECA, shadow difference map) is provided. 
Our method runs in 0.07 seconds and outperforms three acceleration techniques at 10 sampling steps across all metrics, while those techniques already take over one second to run.
%Our method runs in 0.07 seconds and outperforms three acceleration techniques at 10 sampling steps across all metrics, which already take over one second to run.
Increasing the sampling steps further (e.g., to 100) for those techniques can improve their quality but slows down runtime by two orders of magnitude (13.56s). Conversely, reducing the steps below 10 can speed up these techniques but the quality would become extremely poor.
Surprisingly, the single-shot DiFaReli++$_\text{ss}$ also outperforms the original DiFaReli++ across all metrics, even though it was trained on synthetic data generated by DiFaReli++. 
The improvement may result from the single-shot model directly learning image relighting, whereas the original model must disentangle light information and perform DDIM inversion for relighting, which are more complex and prone to errors.

\subsection{User study}
\label{sec:user_study}
%We conducted a user study to evaluating 
%As there is no established quantitative benchmark for evaluating the relighting performance for non-facial parts, 
%There is very little literature that addresses the relighting of non-facial parts, and none of these studies have specifically evaluated non-facial parts relighting. Furthermore, there is no benchmark dataset designed to measure the performance and realism of relighting in non-facial areas. Relying solely on qualitative results could lead to biased and potentially misleading assessments of the actual performance. 
In addition to the quantitative evaluation, we conducted three user studies to assess relighting quality, as follows: 
\subsubsection{Relighting quality on facial and non-facial parts}\label{sec:user_study_targetSH}
This study comprises 100 input images, another 100 images serving as target lighting, and 100 relit results from each method. The input and target lighting images are randomly selected from the FFHQ test set. For each input image, we generated relit results using six recent state-of-the-art baselines \cite{iclight, hou2021towards, hou2022face, ding2023diffusionrig, pandey2021total, Guo_2025_CVPR} and our method. We then asked 30 participants to choose the result that most convincingly relights the input image to match the target lighting, considering: (1) only the face, and (2) the entire person, including both facial and non-facial parts but excluding the background.
In total, we received 600 responses (100 inputs, 2 questions per input, 3 participants per question).
\subsubsection{Relighting quality under moving lights}\label{sec:user_study_cs}
This study examines relighting quality under dynamic lighting conditions, enabling users to assess the temporal consistency of cast shadows.
We compare our method against five recent state-of-the-art baselines \cite{iclight, hou2021towards, hou2022face, ding2023diffusionrig, Guo_2025_CVPR} using 20 randomly selected input images from the FFHQ test set. 
For each input, we generate two 60-frame videos for each method by rotating the input's SH lighting 360 degrees, one around the up axis (yaw) and the other around the forward axis (roll). We then asked 30 participants to choose the video that most convincingly relights the input image with realistic cast shadow effects and consistent shadow movement. We received 600 responses (20 inputs, 2 questions per input, 15 participants per question).

\subsubsection{Relighting quality under rotating HDR environment maps}\label{sec:user_study_cs_hdr}
To compare with \cite{jin2024neural_gaffer}
% , which uses environment maps as input
, we randomly selected three HDR maps from its official repository and rotated them around the yaw and roll axes (similar to Section~\ref{sec:user_study_cs}).
We evaluated on 20 randomly selected FFHQ test inputs, collecting a total of 1,800 responses (3 HDR maps, 20 inputs, 2 questions, 15 participants).

In all studies, we compared each method in a 1-vs-1 manner. All images and videos are in 256x256 resolution. The interfaces used on Amazon Turk are shown in Figures 
% \ref{fig_app:user_study_ui_image}, \ref{fig_app:user_study_ui_video} and \ref{fig_app:user_study_ui_hdr} (Appendix). 
50, 51, and 52 (Appendix). 
The study results are reported in Table~\ref{tab:user_study_res_combine}, with qualitative examples provided in Figures~\ref{fig:ffhq_cs}, \ref{fig:rotate_cs}, and \ref{fig:rotate_hdr}.
Our method was the most preferred across all scenarios in all studies, with average preference rates of 0.70 vs. 0.30 for relighting quality on facial and non-facial parts, 0.72 vs. 0.28 under moving lights, and 0.65 vs. 0.35 under rotating HDR environment maps.
\begin{table}[!htpb]
\centering
% \caption{\textbf{User study results.} Percentage of times users preferred other methods over ours in 1-on-1 comparisons. (1) Relighting quality of facial and non-facial parts (Section \ref{sec:user_study_targetSH}, N=600). (2) Relighting quality under moving lights (Section \ref{sec:user_study_cs}, N=600). (3) Relighting quality under rotating HDR (Section \ref{sec:user_study_cs_hdr}, N=1800), all evaluated on the FFHQ test set.}
\caption{\textbf{User study results.} Percentage of times other methods were preferred over ours in 1-vs-1 comparisons (details are provided in Sections~\ref{sec:user_study_targetSH}–\ref{sec:user_study_cs_hdr}).}
\vspace{-0.2cm}
\resizebox{\columnwidth}{!}{
\fontsize{18}{23}\selectfont 
\newcolumntype{C}{>{\centering\arraybackslash}X}
\begin{tabularx}{\textwidth}{lCCCCCC}
%\toprule
\specialrule{0.08em}{0pt}{0pt}
\multirow{2}{*}{Method} & \multicolumn{3}{c}{\rule{0pt}{18pt}Relit area (\ref{sec:user_study_targetSH})} & \multicolumn{3}{c}{Light rotation (\ref{sec:user_study_cs})} \\
\cmidrule(lr){2-4}
\cmidrule(lr){5-7}
&  Face & \makebox[0pt]{Non-face} & Avg. & Yaw & Roll & Avg. \\ \midrule

Pandey et al. \cite{pandey2021total} & 29\% & 22\% & 26\% & - & - & - \\

Hou et al. \cite{hou2021towards} & 38\% & 40\% & 39\% & 27\% & 32\% & 29\% \\

Hou et al. \cite{hou2022face} & 25\% & 22\% & 24\% & 31\% & 32\% & 32\% \\

Ding et al. \cite{ding2023diffusionrig} & 36\% & 34\% & 35\% & 27\% & 28\% & 28\% \\

Zhang et al. \cite{iclight} & 29\% & 26\% & 28\% & 34\% & 28\% & 31\% \\

Guo et al. \cite{Guo_2025_CVPR} & 34\% & 31\% & 32\% & 24\% & 21\% & 22\% \\

\specialrule{0.08em}{0pt}{0pt}
\end{tabularx}
}

\vspace{0.5em}
% Second table
\resizebox{0.52\columnwidth}{!}{
\fontsize{18}{23}\selectfont
\newcolumntype{C}{>{\centering\arraybackslash}X}
\begin{tabularx}{0.52\textwidth}{lCCC}
\specialrule{0.08em}{0pt}{3pt}
\multirow{2}{*}{Method} & \multicolumn{3}{c}{HDR rotation (\ref{sec:user_study_cs_hdr})} \\
\cmidrule(lr){2-4} & Yaw & Roll & Avg. \\ \midrule
Jin et al. \cite{jin2024neural_gaffer} & 37\% & 33\% & 35\% \\
\specialrule{0.08em}{0pt}{0pt}
\end{tabularx}
}
\label{tab:user_study_res_combine}
\vspace{-2em}
\end{table}

% }
\subsection{Qualitative evaluations}
\label{sec:qual_eval}
% \subsubsection{Shadow flag Condition}

% \revised{It is important to note that the ability to remove shadows is not our main goal like \cite{zhang2020portrait} and is merely one of the potential uses of our method (see Appendix \ref{app:more_results})}.
% %, and none of the previous works cannot do such a cast shadow manipulation.

% \textbf{Shadow flag condition.} We show a novel ability to change the strength of cast shadows on FFHQ\cite{karras2019style} in Figure \ref{fig:cast_shadow}. We generate these results by varying the cast shadow's logit value ($c$). Our method can realistically remove shadows (e.g., those cast by eyeglasses or face geometry) or intensify their effects. Figure \ref{fig-intro} demonstrates how the direction and appearance of cast shadows can change according to a new target lighting condition (the bottom guy's chin).
% %, and none of the previous works cannot do such a cast shadow manipulation.

% \subsubsection{Relighting on FFHQ dataset}
% \textbf{Relighting on FFHQ dataset.} We show qualitative results with more subject variations compared to two recent SOTA \cite{hou2021towards, hou2022face} on FFHQ\cite{karras2019style} in Figure \ref{fig:more_ffhq}. Our method can realistically remove the hard shadow and produce highly realistic relit images, while previous works could suffer from mispredicting  the albedo\cite{hou2022face} and the ratio image\cite{hou2021towards}, resulting in unrealistic relit images.

\textbf{Relighting on FFHQ dataset.} 
In Figure \ref{fig:ffhq_cs}, we present a qualitative comparison with seven recent state-of-the-art methods \cite{pandey2021total, hou2021towards, hou2022face, ding2023diffusionrig, iclight, Guo_2025_CVPR, ponglertnapakorn2023difareli} on subjects with different genders, races, accessories, and clothing.
Our approach produces highly realistic results by effectively handling facial highlights, removing hard shadows and those cast by external objects (Figure \ref{fig:imp_fail}) while simultaneously synthesizing new cast shadows guided by the provided shadow map. In contrast, competing methods often leave behind shadow or shading residuals due to inaccurate albedo predictions for these in-the-wild images.
Furthermore, unlike \cite{ponglertnapakorn2023difareli}, which focuses primarily on the face, our method can realistically relight both the face and clothing to match the target lighting, further enhancing the realism.

\textbf{Relighting with consistent cast shadows.}
In Figure \ref{fig:rotate_cs} and 
Figures 25-29, 
% Figures \ref{fig:rotate_cs_app_1}-\ref{fig:rotate_cs_app_5}, 
we present a qualitative comparison assessing the consistency of cast shadows under moving light conditions on the FFHQ dataset \cite{karras2019style} against six recent competing techniques \cite{hou2021towards, hou2022face, ding2023diffusionrig, iclight, Guo_2025_CVPR, ponglertnapakorn2023difareli}. The input images for this comparison contain varying cast shadows, ranging from hard shadows caused by direct sunlight or a point light source to softer shadows produced under overcast conditions.

Moreover, we compared our method against the recent object-relighting technique, Neural Gaffer \cite{jin2024neural_gaffer}, as shown in Figure~\ref{fig:rotate_hdr} and  
% Figures~\ref{fig_app:ffhq_cs_hdr_app1} and \ref{fig_app:ffhq_cs_hdr_app2} 
Figures~34 and 35
of the Appendix. Since this method takes an HDR environment map as input, we rotated the map and rendered the corresponding shading reference for our method. While Neural Gaffer can relight facial images using an environment map, it is unable to remove existing cast shadows. In contrast, our approach removes such shadows more effectively and produces consistent cast shadow effects.

By leveraging a shadow map, DiFaReli++ can generate realistic and consistent cast shadows that align with a guided shadow map, extending the original DiFaReli's capabilities (best observed in our supplementary videos).

\textbf{Shadow flag condition.}\label{sec:shadow_mani} We show DiFaReli's novel ability to adjust the strength of cast shadows on FFHQ \cite{karras2019style} in Figure \ref{fig:cast_shadow} by varying the cast shadow's value ($c$).
% These results are generated by varying the cast shadow's value ($c$). 
% Our method can realistically remove shadows (e.g., those cast by eyeglasses or face geometry) or intensify their effects. 
Our method can realistically remove or intensify shadows (e.g., those cast by eyeglasses or face geometry).
DiFaReli++ offers similar control by scaling the shadow value used to scale the shadow map (Section \ref{sec:shadowmap}), as shown in our supplementary video.
% DiFaReli++ can achieve the same ability by increasing or decreasing the shadow value used to scale the shadow map (Section \ref{sec:shadowmap}), as shown in our supplementary video.
%to strengthen or attenuate the intensity of cast shadows.

% We can achieve the same masnipulation with DiFaReli++ by increasing or decreasing the shadow value in the shadow map to strengthen or attenuate the intensity of cast shadows.

\subsection{Ablation studies}
\label{sec:AS}

\subsubsection{Light conditioning}
\label{ab:light_cond}
We compare our full pipeline with two alternatives for conditioning the DDIM: a) Removing our Modulator network and directly feeding the reference shading to the DDIM by concatenating it with each $\xt$ at every timestep; and b) Omitting the shading reference entirely and instead concatenating the light encoding $\vect{l}$ with $(\vect{s}, \mathbf{cam}, \boldsymbol{\xi}, c)$ using the non-spatial conditioning technique.

We report the results in Table \ref{tab:AB} and show a qualitative comparison in 
% Appendix \ref{app:more_results}. 
Appendix VI. 
Using the light encoding as part of a non-spatial vector (b) performs the worst among all three methods, while feeding the shading reference directly to the DDIM without our Modulator (a) shows improvement but still lags behind our proposed pipeline.

\begin{figure}
\centering
  \includegraphics[scale=0.33]{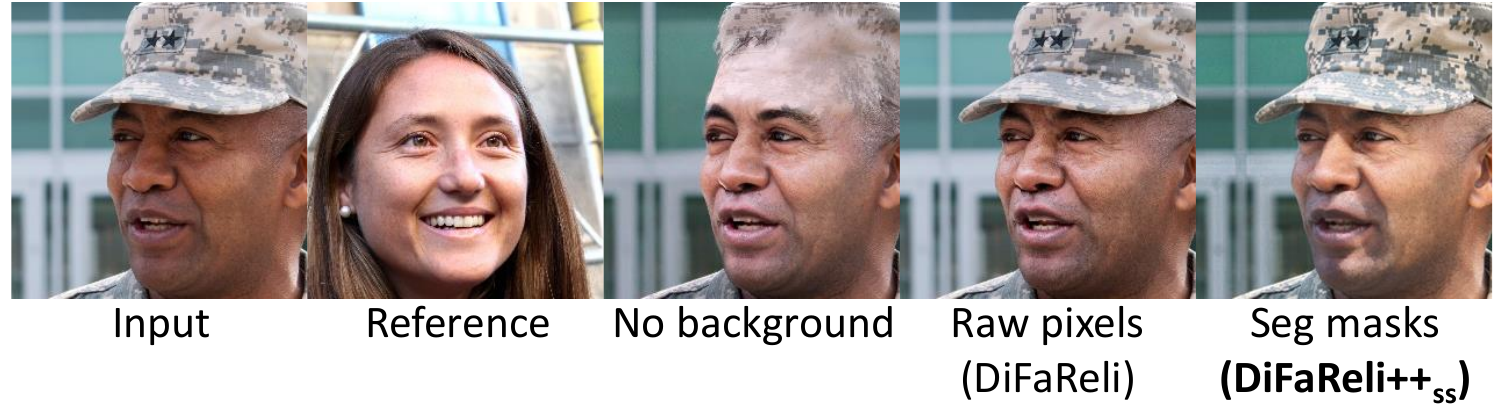}
  \caption{\textbf{Background conditioning ablation.} Without background conditioning, non-facial regions like hats may disappear. Conditioning on raw pixels in DiFaReli preserves the hat, while conditioning on segmentation masks in DiFaReli++$_\text{ss}$ not only preserves it but also enables its relighting.
  }
  \label{fig:bg_ab}
  \vspace{-1em}
\end{figure}

\subsubsection{Non-spatial conditioning}
\label{ab:nonspa}
In this section, we study the benefits of non-spatial, face-related conditions extracted from ArcFace ($\boldsymbol{\xi}$) and DECA ($\vect{s}, \mathbf{cam})$ by evaluating the relight performance on: c) Our method with no $\vect{s}, \mathbf{cam}, \boldsymbol{\xi}$. d) Our method with no $\vect{s}, \mathbf{cam}$. e) Our method with no $\boldsymbol{\xi}$.

We report the results in Table \ref{tab:AB} and a qualitative comparison 
% in Appendix \ref{app:more_results}. 
in Appendix~VI. 
Removing all of $\vect{s}, \mathbf{cam}, \boldsymbol{\xi}$ performs the worst, while removing $\vect{s}, \mathbf{cam}$ but retaining $\boldsymbol{\xi}$ obtains a better MSE score. In contrast, our full pipeline outperforms these alternatives on both DDSIM and LPIPS metrics, aligning with human perception.

\begin{table}[!htpb]
\caption{Ablation study on conditioning inputs.}
\vspace{-0.2cm}
\resizebox{1\columnwidth}{!}{%
\setlength{\tabcolsep}{3pt}
\Large
\begin{tabular}{llll}
\toprule
\multicolumn{1}{l}{Method} & \multicolumn{1}{c}{DDSIM$\downarrow$} & \multicolumn{1}{c}{MSE$\downarrow$} & \multicolumn{1}{c}{LPIPS$\downarrow$} \\ \midrule
\multicolumn{4}{l}{\cellcolor[HTML]{EFEFEF}\textbf{Light conditioning}} \\
\multicolumn{1}{l}{\ \ \ a) No \emph{Modulator}} & \multicolumn{1}{c}{0.0749} & \multicolumn{1}{c}{0.0081} & 0.0868 \\
\multicolumn{1}{l}{\ \ \ b) Used as non-spatial} & \multicolumn{1}{c}{0.0885} & \multicolumn{1}{c}{0.0098} & 0.0947 \\
\multicolumn{1}{l}{\ \ \ \textbf{Ours (DiFaReli)}\ \ \ \ \ \ \ \ \ \ \ \ \ \ \ \ \ \ \ \ \ \ \ \ \ \ \ \ \ \ \ \ \ \ } & \multicolumn{1}{c}{\textbf{0.0670}} & \multicolumn{1}{c}{\textbf{0.0077}} & \textbf{0.0789} 
\vspace{0.5em} \\
\multicolumn{4}{l}{\cellcolor[HTML]{EFEFEF}\textbf{Non-spatial condition vector}} \\
\multicolumn{1}{l}{\ \ \ c) No $\vect{s}, \mathbf{cam}, \boldsymbol{\xi}$} & \multicolumn{1}{c}{0.0713} & \multicolumn{1}{c}{0.0082} & 0.0909 \\
\multicolumn{1}{l}{\ \ \ d) No $\vect{s}, \mathbf{cam}$} & \multicolumn{1}{c}{0.0674} & \multicolumn{1}{c}{\textbf{0.0063}} & \multicolumn{1}{c}{0.0846} \\
\multicolumn{1}{l}{\ \ \ e) No $\boldsymbol{\xi}$} & \multicolumn{1}{c}{0.0686} & \multicolumn{1}{c}{0.0074} & 0.0847 \\
\multicolumn{1}{l}{\ \ \ \textbf{Ours (DiFaReli)}} & \multicolumn{1}{c}{\textbf{0.0670}} & \multicolumn{1}{c}{0.0077} & \textbf{0.0789} \\
\bottomrule
\end{tabular}}
\label{tab:AB}
% \vspace{-1em}
\end{table}

\subsubsection{Conditioning mechanism alternatives}
\label{sec:controlnet_ab}
We compare our conditioning techniques with ControlNet~\cite{zhang2023adding}, published concurrently with our conference version \cite{ponglertnapakorn2023difareli}.
% ControlNet also spatially conditions the UNet by processing conditioning inputs through a duplicate UNet encoder and feeding the output feature maps to the decoder through a zero-conv layer. This idea is conceptually similar to our modulator, which imposes spatial conditioning (e.g., a shadow map) on the UNet, except that we apply the conditioning to the encoder. 
Unlike our modulator, which injects spatial conditioning directly into the UNet encoder, ControlNet injects it into the UNet \emph{decoder}, preceded by a zero-conv layer.
For comparison, we replace our spatial conditioning (Modulator) with ControlNet and/or our non-spatial conditioning with a cross-attention layer, as used in Stable Diffusion (SD)~\cite{rombach2021highresolution} with ControlNet 
% (see details in Section~\ref{app:imp_controlnet}).
(see details in Section~III-C).
%conditions the UNet decoder via a zero-conv layer.
%In this ablation, we replace (a) non-spatial conditioning with a cross-attention layer, used in Stable Diffusion + Controlnet; and (b) spatial conditioning (our Modulator network) with a ControlNet. 
Unlike the original SD+ControlNet setup, where the pretrained SD model is frozen while ControlNet is being trained, we train both ControlNet and our diffusion model jointly from scratch to match our training objective.
%The ControlNet is trained from scratch 
%The ControlNet is trained from scratch, without any Stable Diffusion finetuning.
We evaluate on 100 randomly selected, disjoint self-target lighting pairs from Multi-PIE~\cite{gross2010multi}.

We report results in Table~\ref{tab:controlnet_ab} and show qualitative examples in 
% Figures~\ref{fig_app:controlnet_aba} and \ref{fig_app:controlnet_aba_cs} (Appendix).
Figures~21 and 22 (Appendix).
Quantitatively, we found that the performance is broadly comparable, with our architecture performing slightly better. Using ControlNet for spatial conditioning (Rows a and b) often fails to change the skin tone to match the target lighting. 
However, we found that a simple modification that removes the initial zero convolutional block in ControlNet's original design, denoted by CN-mod in 
% Table~\ref{tab:controlnet_ab}
Table~IV
, addresses the skin-tone matching problem 
% (Figures~\ref{fig_app:controlnet_aba}) 
(Figures~21) 
and further improves quantitative scores. 
%We hypothesize that inputting $\vect{x}_t$ to ControlNet may create an undesirable learning shortcut through skip connections to the last layer of the diffusion model, causing the output to adhere to the input's color tone. 

%We hypothesize that this behavior originates from the ControlNet design, as DiFaReli++’s Modulator with cross-attention does not exhibit such effects. In ControlNet, the spatial conditioning undergoes two rounds of encoding: first through the initial convolutional block, which reshapes it to match $\mathbf{x}_t$, and then again through the copied U-Net encoder. As a result, the encoder processes a pre-compressed latent rather than raw pixels. In contrast, our method feeds raw pixels directly into the encoder. 

%We hypothesize that this behavior originates from the ControlNet design, since DiFaReli++’s Modulator with cross-attention does not exhibit such effects.
%We found that this behavior stems from the initial convolutional block in ControlNet, which reshapes the spatial conditioning to match $\mathbf{x}_t$ before adding it to $\mathbf{x}_t$ and passing the output to the main ControlNet block. 
%In contrast, our method directly encodes the raw pixels without this intermediate reshaping. To further validate this, we include a ControlNet-modified variant in which the initial convolutional block is removed while the rest of the architecture is kept intact. This variant exhibits reduced skin-tone leakage from the input and achieves a closer match to the ground truth.
Regardless of these variations, our proposed shadow map conditioning is architecture-agnostic, enabling all combinations to generate cast shadows 
% (Figure~\ref{fig_app:controlnet_aba_cs}), 
(Figure~22), 
suggesting adaptability to emerging backbones (e.g., DiTs~\cite{lin2024ctrladapterefficientversatileframework, zhang2025easycontrol}).
%We will also release the preprocessed shadow maps to facilitate development with future architectures.

\begin{table}[t]
\centering
\caption{\textbf{Comparison on conditioning mechanism alternatives.}
We replace DiFaReli++’s spatial and non-spatial components with ControlNet (CN) or Cross-attention (CA). (CN-mod) is ControlNet without its first convolutional block.}
\label{tab:controlnet_ab}
\resizebox{1\columnwidth}{!}{%
\begin{tabular}{llccc}
\toprule
\multicolumn{2}{c}{Conditioning Type} & \multirow{2}{*}{DDSIM$\downarrow$} & \multirow{2}{*}{MSE$\downarrow$} & \multirow{2}{*}{LPIPS$\downarrow$} \\
\cmidrule(lr){1-2}
\multicolumn{1}{c}{Spatial} & \multicolumn{1}{c}{Non-Spatial} & & & \\
\midrule
a) CN                 & CA            & 0.0983 & 0.0129 & 0.0868 \\
b) CN                 & DiFaReli++    & 0.1014 & 0.0157 & 0.0840 \\
c) CN-mod         & CA            & 0.0697 & 0.0076 & 0.0751 \\
d) DiFaReli++ & CA            & 0.0694 & 0.0070 & 0.0695 \\
e) DiFaReli++ & DiFaReli++ & \textbf{0.0664} & \textbf{0.0044} & \textbf{0.0689} \\
\bottomrule
\vspace{-2em}
\end{tabular}
}
%\resizebox{0.9\columnwidth}{!}{
%    \footnotesize{$^*$CN-mod: ControlNet without the initial conv block}
%}
\end{table}

\begin{figure}
\centering
    \includegraphics[scale=0.5]{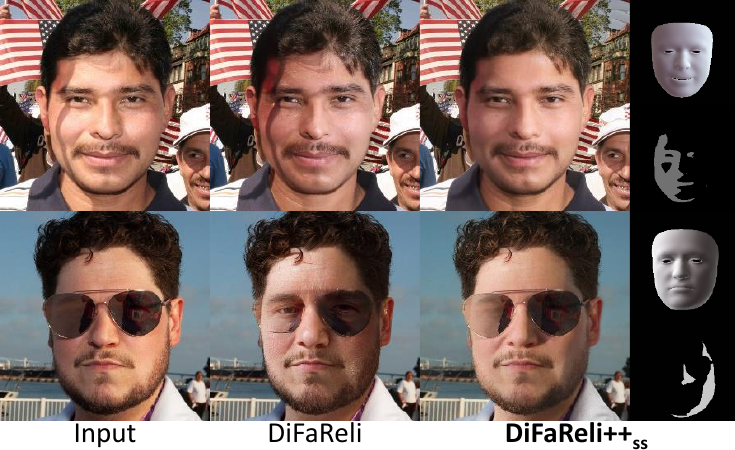}
  \caption{\textbf{Improvements over DiFaReli's failure cases.} 
  DiFaReli++$_\text{ss}$ better remove shadows cast by external objects (top) and better preserves sunglasses (bottom).}
  \label{fig:imp_fail}
\end{figure}

\begin{figure}
\centering
    \includegraphics[scale=0.40]{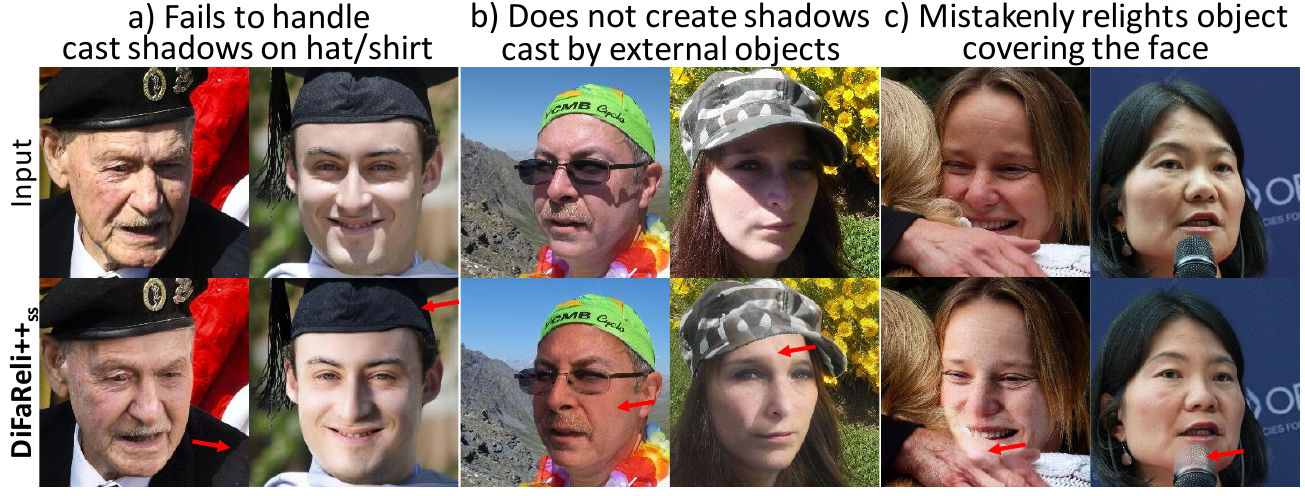}
  \caption{\textbf{Failure cases.} Our method a) may fail to add or remove cast shadows on non-facial parts (e.g., hats, clothing), b) may not produce shadows cast by external objects, or
  c) may mistakenly relight objects occluding the face (e.g., hands), leading to unrealistic relighting in some cases.}
  \label{fig:difareli++_failure}
  \vspace{-1.5em}
\end{figure}

\section{Limitations and discussion}
\label{sec:limitations}
% While our results look photorealistic and plausible, successfully controlling cast shadows by removing and synthesizing new ones according to the guided shadow map, this capability is inherently linked to the geometric estimation of the face. Although our method does not directly rely on precise geometry, achieving more physically accurate results would require a more accurate 3D geometry to obtain an accurate shadow map.
% Our results look photorealistic and plausible, with the ability to successfully control cast shadows by removing or adding new ones according to the guided shadow map. 
While our method produces photorealistic, plausible relit results with consistent cast shadows, the generated cast shadows may not always be physically accurate. Additionally, relighting to match a reference lighting image can be inaccurate as we rely on a light estimator to determine the target lighting. This process is susceptible to ambiguity where it is unclear whether a dark appearance is due to the skin tone or dim lighting.
%is susceptible to the ambiguity where it is unclear if, e.g., a dark appearance is caused by the skin tone or dim lighting.
% However, this capability inherently depends on two key factors:
% First, replicating the real subject's facial structure requires accurate geometric estimation. For example, inaccuracies in predicting the height of facial features (e.g., nose or cheeks) can produce a shadow map that misaligns with the subject's actual geometry, causing the generated cast shadows in the relit image to align incorrectly with the facial structure.
% Second, relighting to match a reference lighting image can be inaccurate as we rely on a light estimator, which is susceptible to ambiguities such as determining whether a dark appearance is caused by the subject's skin tone or dim lighting.
Other limitations (Figure \ref{fig:difareli++_failure}) include mistakenly relit objects overlapping the face. Shadows cast by external objects (e.g., hats, glasses) are not generated, as our current shadow map is specific to the face area.
%this would require estimating the geometry of these parts, which is more challenging than estimating face geometry. 
%Our method may inaccurately relight cast shadows on non-facial regions, causing areas under the shadows to brighten but not to the same level as areas outside them.
% Our method may inaccurately relight cast shadows on non-facial regions, resulting in inconsistent intensity matching with areas outside the shadows.
%Our method may inaccurately relight shadows on non-facial areas incorrectly, resulting in inconsistent blending with the brighter surrounding regions.
While our method can relight non-facial areas by adjusting their shading, it often fails to synthesize or remove cast shadows from them.
%Our method may not synthesize or remove cast shadows on non-facial areas like clothing, resulting in 

Although our method, including the single-shot model, requires multiple conditionings (e.g., DECA, shadow map) to be pre-processed before relighting, this preprocessing is required only once per input image and can be reused indefinitely.
Since our method does not rely on precise intrinsic decomposition, it opens up the possibility of extending relighting to full human bodies, objects, or scenes, where such decomposition is more challenging and prone to errors. Additionally, exploring alternative lighting models, such as HDRI, could also further enhance the framework's capabilities.

\section{Conclusion}
\label{sec:conclusion}
We present a diffusion-based face relighting method that eliminates the need for accurate intrinsic decomposition and can be trained on 2D images without requiring multiview images, relit pairs, light stage data, or 3D ground truth.
Our key component is a conditional diffusion implicit model and a novel conditioning technique that maps a disentangled light representation to a relit image. 
This extended paper enhances the realism and consistency of cast shadows while expanding relightable regions to non-facial areas using segmentation masks and a shadow map estimated by our novel method.
Moreover, we distill our method into a single-shot relighting network that runs in a single network pass and even surpasses the teacher model.
Our method achieves state-of-the-art results and produces highly photorealistic outputs with temporally consistent cast shadows in real-world scenarios.

  \FloatBarrier
  {
    \small
    \bibliographystyle{ieee_fullname}
    \bibliography{content/egbib.bib}
  }
\begin{IEEEbiography}[{\vspace{-4em}\includegraphics[width=1in,clip,keepaspectratio]{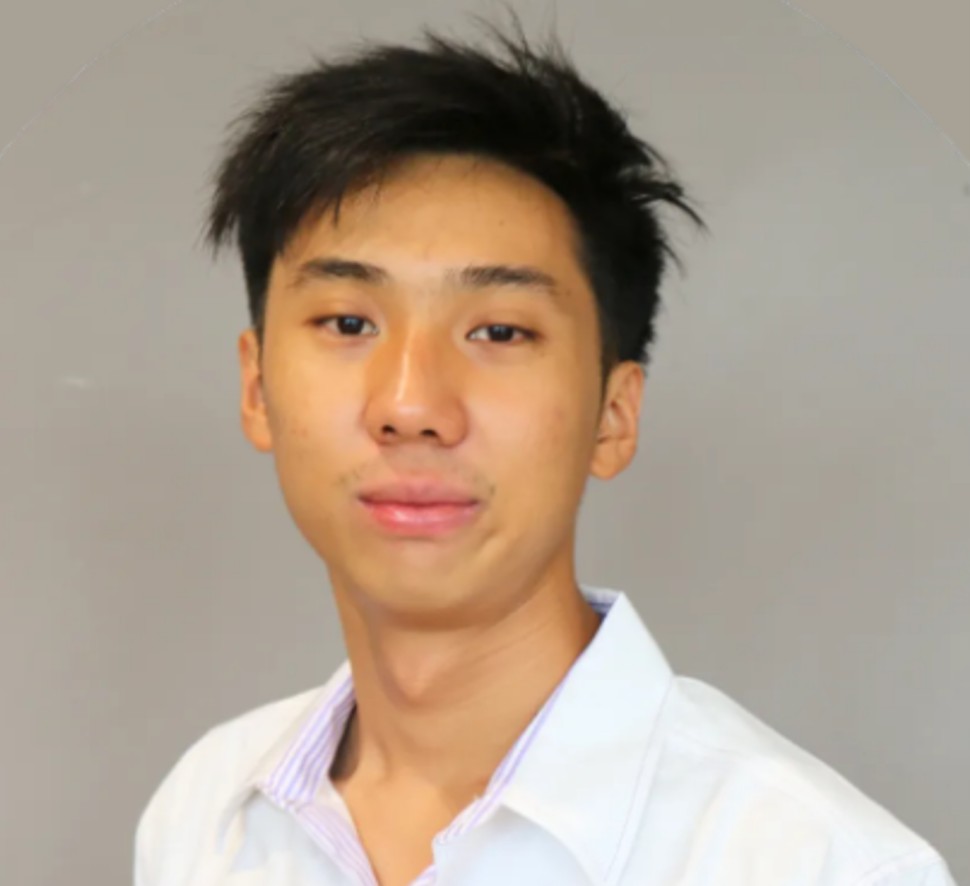}}]{Puntawat Ponglertnapakorn} is currently a Ph.D. student at the School of Information Science and Technology, Vidyasirimedhi Institute of Science and Technology (VISTEC), Thailand. He received the B.Eng. degree from Prince of Songkla University (PSU), Thailand, in 2018. His research interest is computer vision.
\end{IEEEbiography}
\vspace{-4em}
\begin{IEEEbiography}[{\vspace{-4em}\includegraphics[width=1in,clip,keepaspectratio]{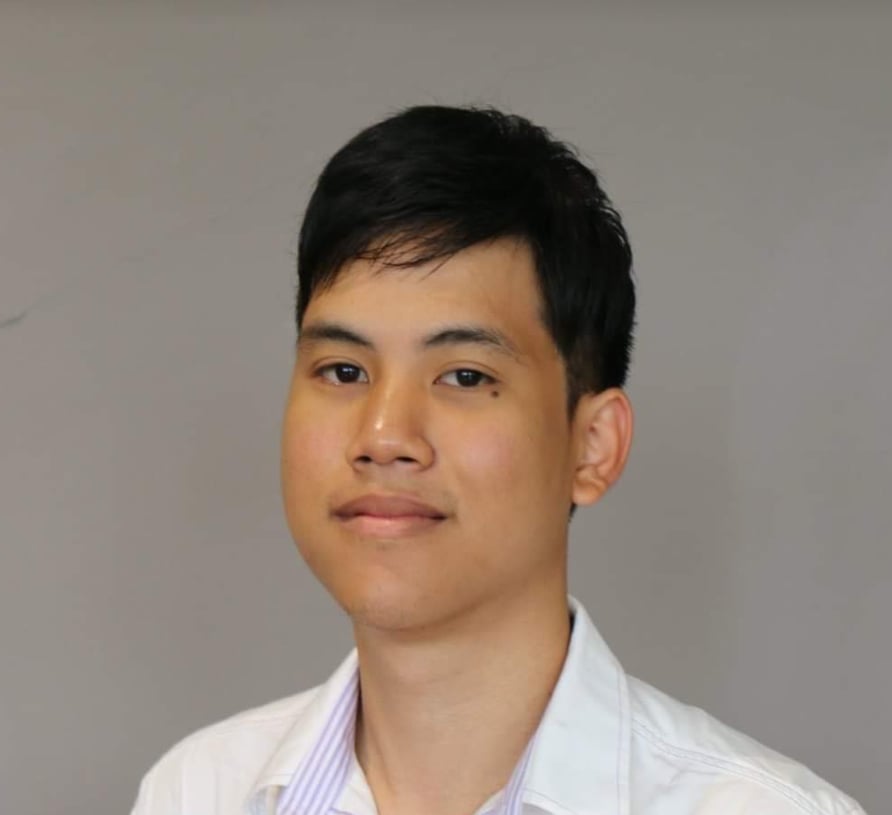}}]
{Nontawat Tritrong} is currently a Ph.D. student at the School of Information Science and Technology, Vidyasirimedhi Institute of Science and Technology (VISTEC), Thailand. He received the B.Sc. degree from Suranaree University of Technology (SUT), Thailand, in 2019. His research interest is representation learning.
\end{IEEEbiography}
\vspace{-4em}
\begin{IEEEbiography}[{\vspace{-1em}\includegraphics[width=1in,clip,keepaspectratio]{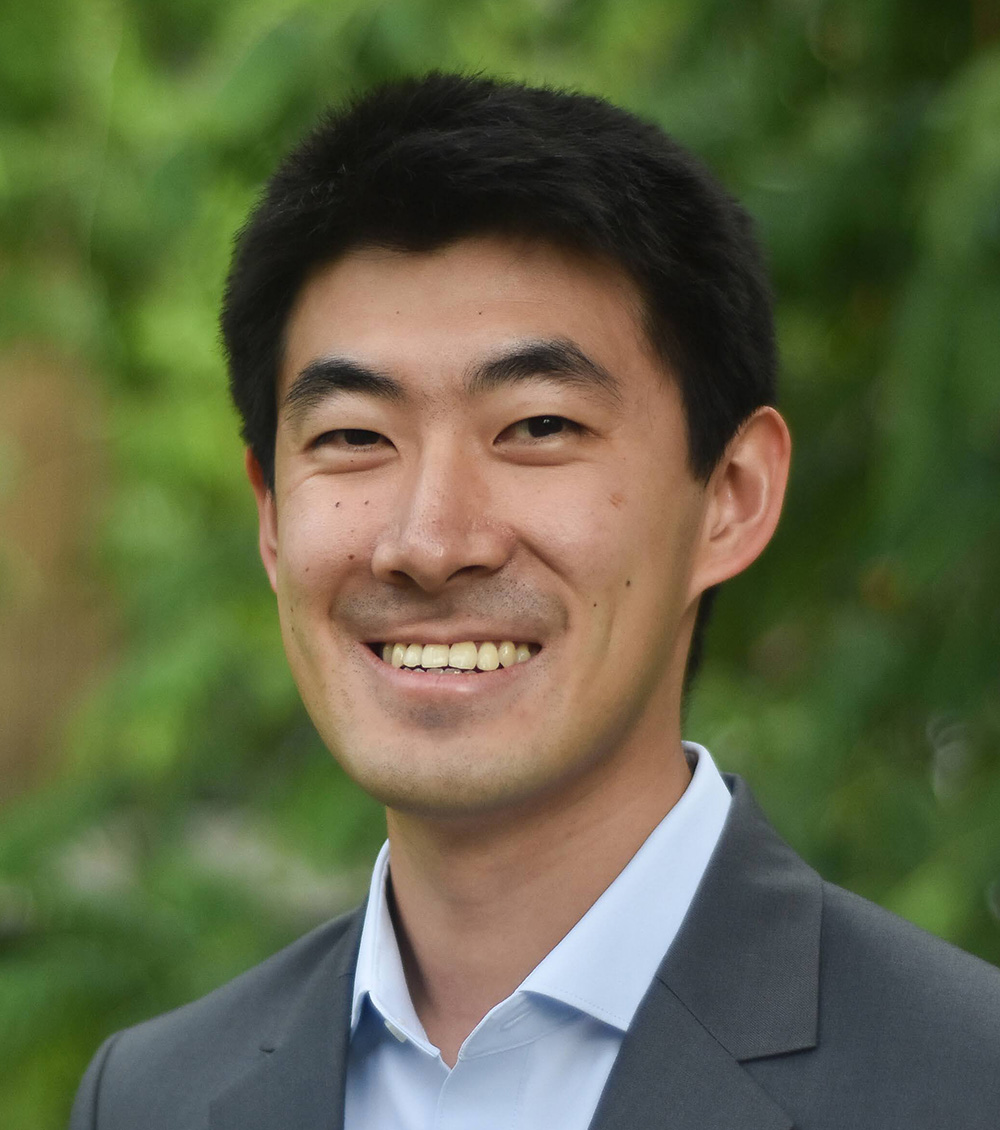}}]{Supasorn Suwajanakorn} received the B.Eng. degree from Cornell University, in 2011 and the Ph.D. degree from the University of Washington, in 2017. He is currently a lecturer with the school of Information Science and Technology (IST), Vidyasirimedhi Institute of Science and Technology (VISTEC), Thailand. His research interests lie in the intersection of computer vision, deep learning, and computer graphics.
\end{IEEEbiography}

% \appendix
\clearpage
\newpage
% \title{Appednix: Diffusion Face Relighting}
% \maketitle
% \Large\textbf{Appednix: Diffusion Face Relighting}

{\fontsize{13}{0}\selectfont\textbf{Appendix for DiFaReli++: Diffusion Face Relighting with Consistent Cast Shadows}}
\setcounter{section}{0} % Reset section counter

\section{Overview}
\label{app:sup_mat}

In this Appendix, we present:
\begin{itemize}[noitemsep,topsep=0pt]
    \item Section \ref{app:impl}: Implementation details.
    \item Section \ref{app:net_arch}: Network architectures.
    \item Section \ref{app:render_face}: 3D Face rendering.
    % \item Section \ref{app:ablation_studies}: Additional ablation studeis.
    \item Section \ref{app:related_work}: Additional related works.
    \item Section \ref{app:more_results}: Additional results.
    \item Section \ref{app:negative}: Potential negative societal impacts.
    \item Section \ref{app:user_study_ui}: User interface for relighting user study.
\end{itemize}

\section{Implementation details}
\label{app:impl}

\subsection{Datasets}
For all experiments in Section \ref{sec:LP}, we trained our network on the FFHQ dataset \cite{karras2019style}, which consists of 70,000 aligned face images (60k for training and 10k for testing). We evaluated the relighting performance on Multi-PIE dataset \cite{gross2010multi}, which contains 337 subjects captured under 19 flashes. In ``self target lighting,'' we use the same test set as \cite{hou2022face}, which contains pairs of images from the same person but in different lighting. For ``target lighting from others,'' we randomly pick 200 triplets of the input, target, and ground truth, where the target image is of a different person. For all ablation studies (Section \ref{sec:AS}), to cap the computational resources, each ablated variation is trained on the FFHQ dataset at 128$\times$128 resolution and evaluated on Multi-PIE dataset. For evaluation, we randomly pick 200 pairs, using the same policy as the ``self target lighting,'' from the disjoint set of other experiments in Section \ref{sec:LP}.

\subsection{Training and Inference}
We normalize the training images to [-1,1], and precompute their encodings from DECA, ArcFace, and our shadow estimator. We train our DDIM and Modulator using training hyperparameters in Table \ref{tab:hyperparams}. 

\textbf{128$\times$128 resolution.} We used four Nvidia RTX2080Tis for training and one Nvidia RTX2080Ti for testing. The training took around 1 day using batch size 32, and the inference took $101.38\pm{0.64}$s per image.

\textbf{256$\times$256 resolution.} We used four Nvidia V100s for training and one Nvidia RTX2080Ti for testing. The training took around 8 days using batch size 20, and the inference took $194.29\pm{9.17}$s per image.

\subsection{Improved DDIM sampling with mean-matching}
\label{sec:meanmatch}
We observe that when the input image contains background pixels with extreme intensities (e.g., too dark or too bright), the output tends to have a slight change in the overall brightness, most noticeable in the background (see Figure \ref{fig:mm}). This behavior also occurs with DDIM inversion that involves no relighting, i.e., when we reverse $\xT = \text{DDIM}^{-1}(\xzero)$ and decode $\xzero' = \text{DDIM}(\xT)$ without modifying the light encoding, $\xzero'$ can look slightly different from $\xzero$ in terms of the overall brightness. 

We found that we can correct the overall brightness with a simple, global brightness adjustment within DDIM's generative process as follows. We first perform self-reconstruction by running DDIM's reverse generative process starting from the input $\xzero$ to produce $\xzero, \vect{x}_1, ..., \xT$, then decoding back $\xT', \vect{x}_{T-1}', ..., \xzero'$, where $\xT' = \xT$ using Equation 4 in the main paper and its reverse. Then, our correction factor sequence, $\mu_0, \mu_1, ..., \mu_T$, is computed by taking the difference between the mean pixel values of $\vect{x}$ and $\vect{x}'$:
\begin{equation}
    \mu_t = \mathrm{mean}(\xt') - \mathrm{mean}(\xt) \label{eq:mm_mu}.
\end{equation}
We compute the mean separately for each RGB channel and compute this correction sequence \emph{once} for each input image. Then, during relighting, we add $\mu_t$ to the generative process conditioned on the modified feature vector, starting from $\xT$. That is, we use the reverse of Equation 4 in the main paper to first produce $\vect{x}_{T-1}$ from $\xT$, and add $\mu_{T-1}$ to it: $\vect{x}_{T-1} \leftarrow \vect{x}_{T-1} + \mu_{T-1}$. Then, we continue the process until we obtain the relit output at $t=0$.

\section{Network architectures}
\label{app:net_arch}

\subsection{Conditional DDIM \& Modulator}
\label{app:cond_ddim_mod}
\begin{figure}[!htpb]
  \centering
  \includegraphics[scale=0.52]{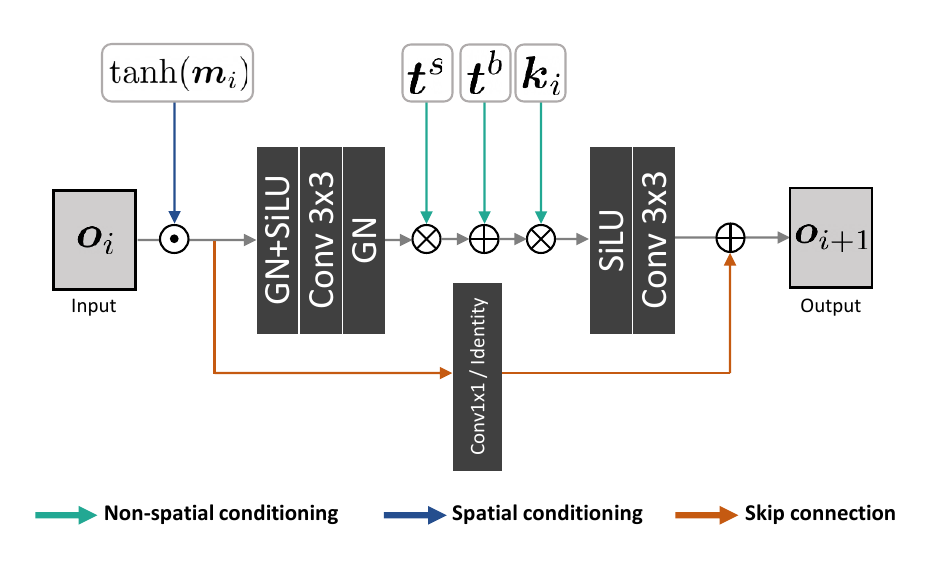}
  \caption{Diagram of one of the residual blocks inside the first half of our conditional DDIM.}
  \label{fig:arch}
  \vspace{-1.0em}
\end{figure}

Our conditional DDIM architecture is based on Dhariwal et al. \cite{dhariwal2021diffusion}, with two main differences: (1) the output of each residual block in the first half of the UNet is modulated by the signal from the Modulator network, and (2) we use our own version of adaptive group normalization. The Modulator shares the same architecture and hyperparameters as the first half of DDIM’s UNet but does not share weights. Each residual block in the first half of the network uses both spatial and non-spatial conditioning (Figure \ref{fig:arch}), while those in the second half use only non-spatial conditioning.

\begin{table}[!htpb]
\centering
\caption{Our conditional DDIM's configuration is based on the architecture of \cite{dhariwal2021diffusion}.}
\begin{tabular}{l|cc}
\toprule
\multicolumn{1}{l|}{\textbf{Parameter}} & \multicolumn{1}{c}{\textbf{FFHQ 128}} & \multicolumn{1}{c}{\textbf{FFHQ 256}} \\
\midrule
Base channels & 128 & 128 \\
Channel multipliers & [1,1,2,3,4] & [1,1,2,2,4,4] \\
Attention resolution & [16, 8] & [16, 8] \\
Batch size & 32 & 20 \\
Image trained & 1.6M & 1.7M \\
Diffusion step & \multicolumn{2}{c}{1000} \\
Learning rate & \multicolumn{2}{c}{1e-4} \\
Weight decay & \multicolumn{2}{c}{-} \\
Noise scheduler & \multicolumn{2}{c}{Linear} \\
Optimizer & \multicolumn{2}{c}{AdamW} \\
 \bottomrule
\end{tabular}
\label{tab:hyperparams}
\end{table}

\subsection{Non-spatial encoding}
\label{app:non_spa_sec}
\begin{figure}[!htpb]
  \centering
  \includegraphics[scale=0.55]{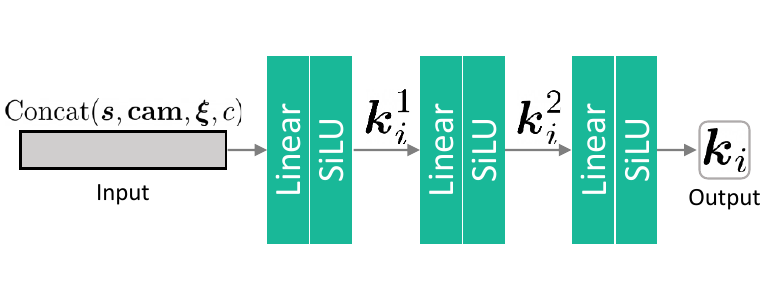}
  \caption{Diagram of one of the 3-layer MLPs in the non-spatial conditioning branch.}
  \label{fig:lin}
  \vspace{-1.0em}
\end{figure}

The concatenation of $(\vect{s}, \mathbf{cam}, \boldsymbol{\xi}, c)$ is passed through 3-layer MLPs (Figure \ref{fig:lin}). For each $\text{MLP}_{i}$, we use fixed-dimension hidden layers $\vect{k}^{1}_{i}, \vect{k}^{2}_{i} \in \mathbb{R}^{512}$, while the dimension of each $\vect{k}_{i}$ depends on the channel dimension of each residual block.

\begin{figure*}[!htpb]
  \centering
  \includegraphics[scale=0.47]{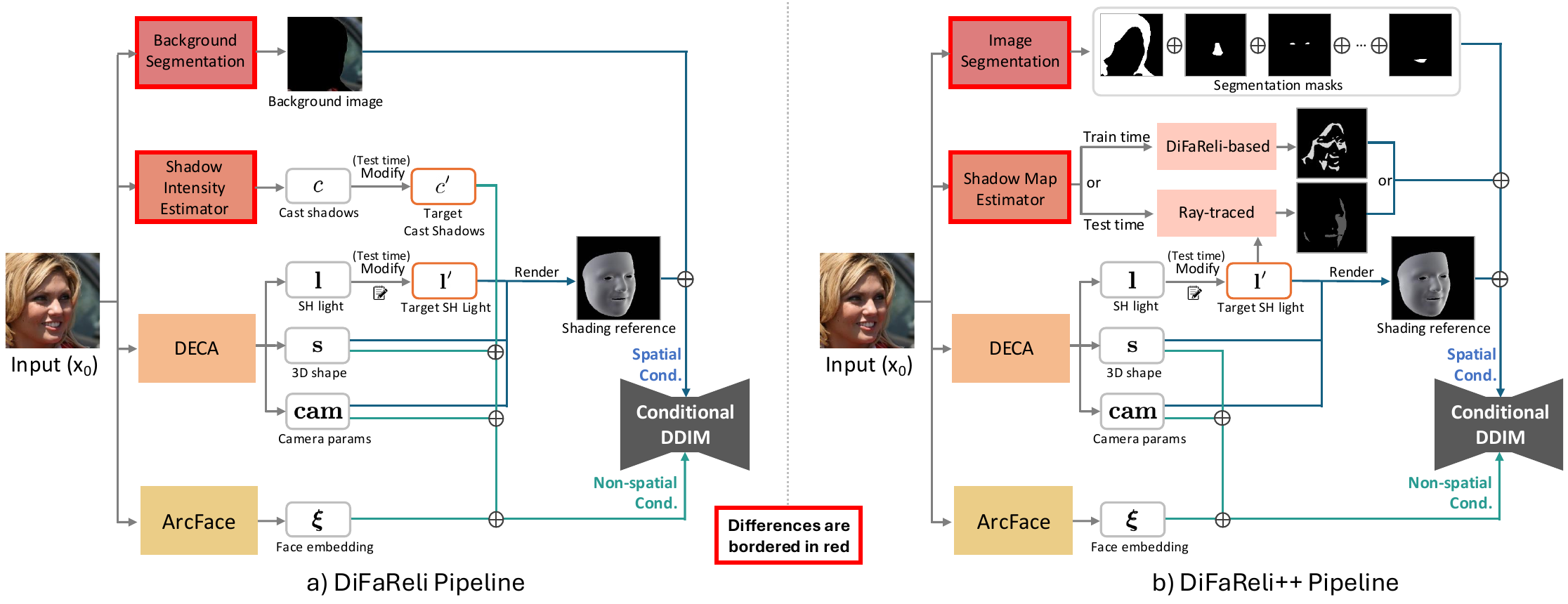}
  \caption{\textbf{Comparison of DiFaReli and DiFaReli++ pipelines.} Differences are highlighted with red borders. Key changes are: 1) Background conditioning: replacing the background image with a concatenation of segmentation masks to enable relighting of non-facial parts. 2) Shadow estimator: using a shadow map with an encoded shadow scalar for improved consistency in cast shadows generation. 3) The cast shadow scalar $c$ is not longer part of the non-spatial conditioning.}
  \label{fig:pipelinecompare}
\end{figure*}

\subsection{ControlNet implementation for conditioning mechanism alternatives ablation (Section \ref{sec:controlnet_ab})}
\label{app:imp_controlnet}

We follow the official implementations of ControlNet and cross-attention provided in the \href{https://github.com/lllyasviel/ControlNet.git}{GitHub repository}~\cite{zhang2023adding}, with slight modifications. For all settings in Table \ref{tab:controlnet_ab}, we use the same hyperparameters (e.g., base channels, channel multipliers, and attention resolutions) as our conditional DDIM (see Table \ref{tab:hyperparams}). For setting b) CN + DiFaReli++, we apply our non-spatial conditioning to ControlNet by injecting the non-spatial vector through AdaGN within each ResNet block (similar to DiffAE~\cite{preechakul2022diffusion}), instead of through the cross-attention layer. For setting c) CN-Mod, we remove only the initial zero-convolution block. And for setting d), we feed our non-spatial vector into the cross-attention layer.

\section{3D face rendering}
\label{app:render_face}
We compute the shading reference $R$ used in the spatial conditioning by:
\begin{equation}
    R_{i,j} = A \odot \sum_{k=1}^{9}\vect{l}_{k}H_{k}(N_{i,j}),
\end{equation}

where $i,j$ denote pixel $(i,j)$ in image space, $A=[0.7,0.7,0.7]$ is a constant gray albedo, $l_{k}\in \mathbb{R}^{3}$ is the $k$-th second-order spherical harmonic RGB coefficient predicted from DECA, $H_{k}:\mathbb{R}^{3}\rightarrow \mathbb{R}$ is the $k$-th spherical harmonic basis function, $N_{i,j} \in \mathbb{R}^3$ is the normalized surface normal at pixel $(i, j)$.

\section{Additional related works}
\label{app:related_work}
\textbf{Conditional DDPMs.} Diffusion models (DDPMs) \cite{ho2020denoising} and scored-based models \cite{pmlr-v37-sohl-dickstein15, NEURIPS2019_3001ef25} have been used to solve multiple conditional generation tasks \cite{croitoru2022diffusion}, such as conditional image synthesis \cite{dhariwal2021diffusion, ho2022classifier, sinha2021d2c, chao2022denoising}, image-to-image translation \cite{saharia2022palette}, image super-resolution \cite{ho2022cascaded, pandey2021vaes}, image segmentation \cite{amit2021segdiff, baranchuk2021label} and image manipulation \cite{preechakul2022diffusion, meng2021sdedit}. Many recent approaches use cross-modal embeddings from popular language models \cite{radford2021learning, vaswani2017attention, raffel2020exploring} as conditions for diffusion models \cite{ramesh2022hierarchical, saharia2022photorealistic, rombach2022high, ruiz2022dreambooth, nichol2021glide, balaji2022ediffi, poole2022dreamfusion, zhang2023adding}, which enables general text-to-image generation and image manipulation. However, they lack the ability to precisely manipulate lighting attributes or directions. DiffAE \cite{preechakul2022diffusion} conditions a DDIM with a 1D latent vector that is learned to capture semantically meaningful information. Manipulating this novel latent vector allows manipulation of various semantic face attributions. Unlike DiffAE, which implicitly models semantic attributes via a learnable latent code, our method requires an explicit and interpretable light encoding, which can be controlled by the user.
% Recently, ControlNet \cite{zhang2023adding} extends clip-based diffusion model to be conditioned on various conditions (e.g., segmentation map, normal map) by introducing a control network which conditions pre-trained diffusion model via 

\textbf{Single-view 3D face modeling.} Our work uses DECA \cite{feng2021learning} to estimate the 3D shape and spherical harmonic lighting information.
Based on the pioneer work of Blanz and Vetter \cite{blanz1999morphable}, DECA regresses the parameters of a FLAME model \cite{FLAME:SiggraphAsia2017}, which represents the face shape with three linear bases corresponding to the identity shape, pose, and expression, and further recovers person-specific details that can change with expression. Our work only uses the FLAME estimate from DECA without the additional facial details. Note that other 3D face modeling techniques, such as \cite{deng2019accurate, genova2018unsupervised, jiang20183d, li2018feature}, can also be used in our framework.

\textbf{Face recognition model for deep face embedding.} Our work leverages a face recognition model, ArcFace \cite{deng2019arcface}, to preserve the identity of the relit face. 
Most previous face recognition models are trained using softmax loss \cite{taigman2014deepface, parkhi2015deep, masi2019face} and triplet loss \cite{schroff2015facenet, liu2015targeting} 
(See \cite{wang2021deep} for a review.)
However, they do not generalize well with open-set recognition and large scale recognition.
%because softmax loss transformation matrix grows linearly with the number of identities and triplet loss sampling is difficult. 
ArcFace adopts Additive Angular Margin loss, which retains discriminativeness while avoiding the sampling problem of the triplet loss. Arcface also proposed a sub-center procedure, which helps improve the robustness of the embedding. Note again that other face embedding models, such as \cite{taigman2014deepface, parkhi2015deep, masi2019face, liu2015targeting}, can also be used in our framework.
%can also be used in our method.

\section{Additional results}
\label{app:more_results}
In this section, we provide additional results:
\begin{itemize}[noitemsep]
    \item Table \ref{tab:full} shows full statistics of Table 1 (Top, main paper) with standard errors, as well as more results from additional baselines \cite{sengupta2018sfsnet, zhang2020portrait, shu2017portrait}. 
    \item Figures \ref{fig_app:controlnet_aba} and \ref{fig_app:controlnet_aba_cs} show qualitative results from the ablation study on alternative conditioning mechanisms (Section \ref{sec:controlnet_ab}).
    \item Figure \ref{fig:aba_light} shows qualitative results of the ablation study on light conditioning (Section \ref{ab:light_cond}).
    \item Figure \ref{fig:aba_nonspa} shows qualitative results of the ablation study on non-spatial conditioning (Section \ref{ab:nonspa}).
    \item Figure \ref{fig:bg_ab} shows qualitative results of the ablation study on background conditioning.
    % \item Figure \ref{fig_app:runtime_difareli} and \ref{fig_app:speedup_difareli} present additional qualitative results, demonstrating that the same single-shot inference framework (Section \ref{single_shot_inf}) can also be applied to our original DiFaReli \cite{ponglertnapakorn2023difareli}, resulting in runtime improvements and outperforming the original DiFaReli across all metrics.
    % \item Figure \ref{fig:aba_arch_ss} shows qualitative results of the ablation study on architecture design for single-shot inference face relighting (Section \ref{ab:arch_ss}).
    \item Figure \ref{fig:rotate_cs_app_1}, \ref{fig:rotate_cs_app_2}, \ref{fig:rotate_cs_app_3}, \ref{fig:rotate_cs_app_4}, \ref{fig:rotate_cs_app_5}, \ref{fig_app:ffhq_cs_app1}, \ref{fig_app:ffhq_cs_app2}, \ref{fig_app:ffhq_cs_app3} and \ref{fig_app:ffhq_cs_app4} show additional qualitative results on the FFHQ dataset.
    \item Figures \ref{fig_app:holo_res} and \ref{fig_app:switch_res} show qualitative comparisons with two recent state-of-the-art methods, HoloRelighting \cite{mei2024holo} and SwitchLight \cite{kim2024switchlight}.
    \item Figures \ref{fig_app:ffhq_cs_hdr_app1} and \ref{fig_app:ffhq_cs_hdr_app2} show qualitative comparisons with recent object relighting methods, Neural Gaffer \cite{jin2024neural_gaffer}.
    \item Figures \ref{fig_app:tuninggrid_relipa_rot_sj1}, \ref{fig_app:tuninggrid_relipa_rot_sj2}, \ref{fig_app:tuninggrid_diffusionrig_rot_sj1}, \ref{fig_app:tuninggrid_diffusionrig_rot_sj2}, \ref{fig_app:tuninggrid_relipa_targetSH_sj1}, \ref{fig_app:tuninggrid_relipa_targetSH_sj2}, \ref{fig_app:tuninggrid_relipa_targetSH_sj3} and \ref{fig_app:tuninggrid_diffusionrig_targetSH} show qualitative comparisons on tuning the relighting hyperparameters for DiffusionRig \cite{ding2023diffusionrig} and Guo et al.\cite{Guo_2025_CVPR}.
    \item Figure \ref{fig:mm} shows a qualitative comparison for the ablation study of the mean-matching algorithm (Section \ref{sec:meanmatch}). 
    \item Figure \ref{fig:shad} shows more qualitative results on cast shadow manipulation.
\end{itemize}

\begin{table*}[!htbp]
\centering
\caption{\textbf{State-of-the-art comparison on Multi-PIE.} We report the means and standard errors. Our method outperforms all previous methods on all metrics with p-values $< 0.001$.}
\resizebox{1.8\columnwidth}{!}{%
\setlength{\tabcolsep}{8pt}
\begin{tabular}{l|cc|cc|cc}
\toprule
\multirow{2}{*}{Method} & \multicolumn{2}{c|}{DDSIM$\downarrow$} & \multicolumn{2}{c|}{MSE$\downarrow$} & \multicolumn{2}{c}{LPIPS$\downarrow$} \\
 & \multicolumn{1}{c}{Mean} & \multicolumn{1}{c|}{SE}  & \multicolumn{1}{c}{Mean} & \multicolumn{1}{c|}{SE}  & \multicolumn{1}{c}{Mean} & \multicolumn{1}{c}{SE}  \\ \midrule
SfSNet \cite{sengupta2018sfsnet} & 0.2918 & 0.0013 &  0.0961 & 0.0017 &  0.5222& 0.0025  \\
DPR \cite{zhou2019deep} & 0.1599 & 0.0019 & 0.0852 & 0.0018 & 0.2644 & 0.0028  \\
SIPR \cite{sun2019single} & 0.1539 & 0.0015 & 0.0166 & 0.0004 & 0.2764 & 0.0025 \\
Nestmayer et al. \cite{nestmeyer2020learning} & 0.2226 & 0.0046 & 0.0588 & 0.0018 & 0.3795 & 0.0078 \\
Pandey et al. \cite{pandey2021total} & 0.0875 & 0.0007 & 0.0165 & 0.0003 & 0.2010 & 0.0022 \\
Hou et al.(CVPR'21) \cite{hou2021towards} & 0.1186 & 0.0013 & 0.0303 & 0.0006 & 0.2013 & 0.0023 \\
Hou et al.(CVPR'22) \cite{hou2022face} & 0.0990 & 0.0013 & 0.0150 & 0.0004 & 0.1622 & 0.0017\\
Ding et al. (CVPR'23) \cite{ding2023diffusionrig} & 0.0870 & 0.0009 & 0.0098 & 0.0002 & 0.2098 & 0.0014\\
Zhang et al. (IC-Light's github) \cite{iclight} & 0.1978 & 0.0015 & 0.0499 & 0.0017 & 0.1887 & 0.0063\\
Guo et al. (CVPR'25) \cite{Guo_2025_CVPR} & 0.1088 & 0.0015 & 0.0234 & 0.0006 & 0.1733 & 0.0018\\
Ours (DiFaReli \cite{ponglertnapakorn2023difareli}) & 0.0711 & 0.0011 & 0.0122 & 0.0005 & 0.1370 & 0.0020 \\
% \textbf{Ours (DiFaReli++)} & \textbf{0.0604} & \textbf{0.0010} & \textbf{0.0090} & \textbf{0.0003} & \textbf{0.1043} & \textbf{0.0013} \\ 
Ours (DiFaReli++) & 0.0604 & 0.0010 & 0.0090 & 0.0003 & 0.1043 & 0.0013 \\ 
\textbf{Ours (DiFaReli++$_\text{ss}$)} & \textbf{0.0590} & \textbf{0.0009} & \textbf{0.0075} & \textbf{0.0002} & \textbf{0.1023} & \textbf{0.0013} \\ \bottomrule

\end{tabular}}
\label{tab:full}
\end{table*}

\subsection{Additional results and comparison with concurrent relighting methods that use HDR environment map}
\label{sec_app:compare_more_hdr}

In this section, we show additional qualitative results of Pandey et al. \cite{pandey2021total}, as a supplement to Figure \ref{fig:ffhq_cs} from the main paper. We also compare our method with two concurrent works, HoloRelighting \cite{mei2024holo} and SwitchLight \cite{kim2024switchlight}. 
As none of these methods \cite{pandey2021total, mei2024holo, kim2024switchlight} released their source code, and their datasets are proprietary, we requested the authors to test their algorithms on the standard Multi-PIE and FFHQ datasets. Only Pandey et al. \cite{pandey2021total} provided results generated by the authors themselves, including the estimated environment maps. For HoloRelighting \cite{mei2024holo} and SwitchLight \cite{kim2024switchlight}, we cropped and ran our algorithm on the samples provided in their papers.

Figures \ref{fig_app:ffhq_cs_app1}, \ref{fig_app:ffhq_cs_app2}, \ref{fig_app:ffhq_cs_app3}, and \ref{fig_app:ffhq_cs_app4} show additional subjects compared with Pandey et al. \cite{pandey2021total}. Our method produces more photorealistic results with synthesized cast shadows, whereas \cite{pandey2021total} tends to produce blurrier outputs with white area artifacts. For example, such artifacts are observable on the forehead of the first subject (Figure \ref{fig_app:ffhq_cs_app4}) and the third subject (Figure \ref{fig_app:ffhq_cs_app1}).

\begin{figure*}[]
  \centering
  \includegraphics[scale=0.31]{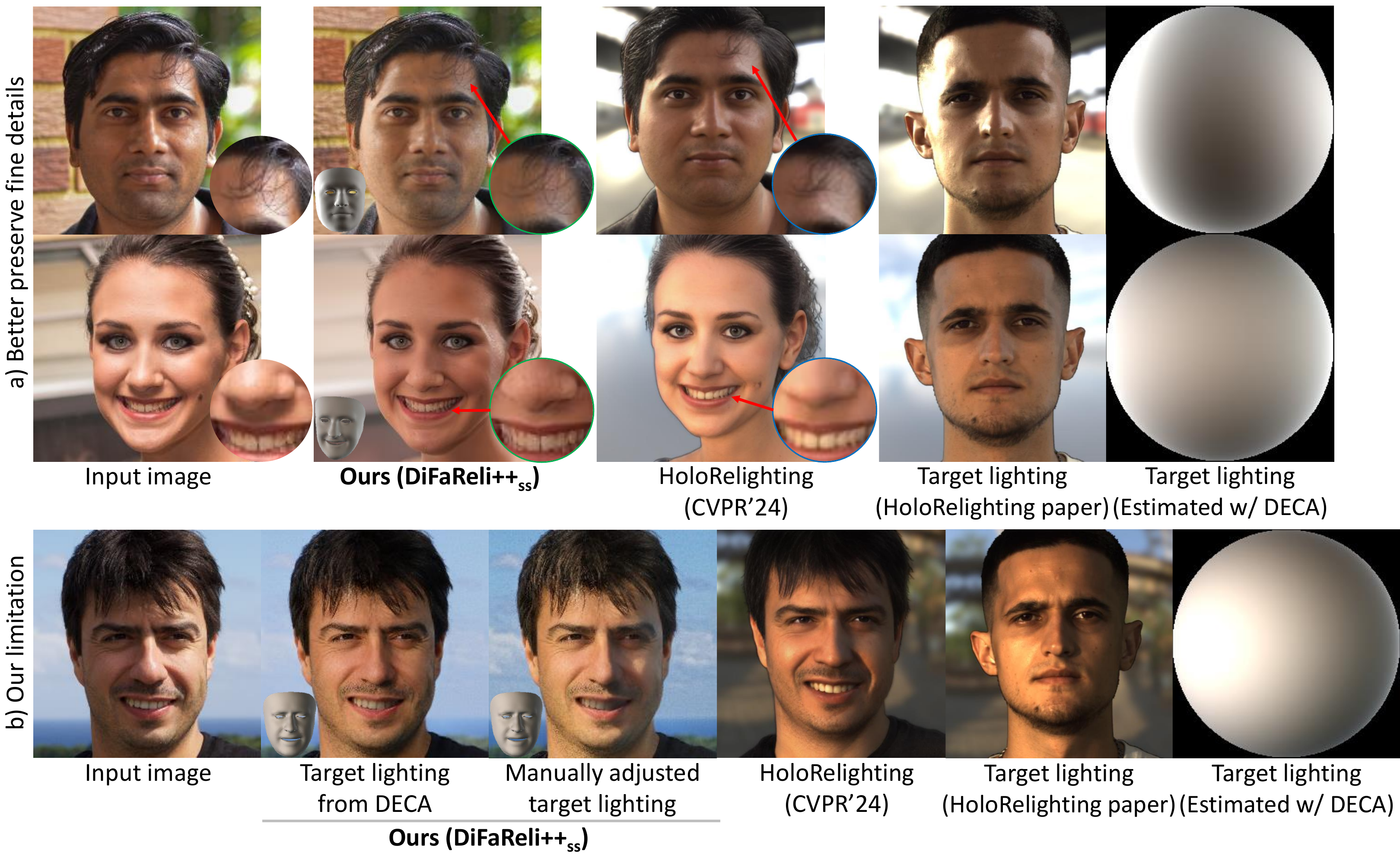}
  \includegraphics[scale=0.49]{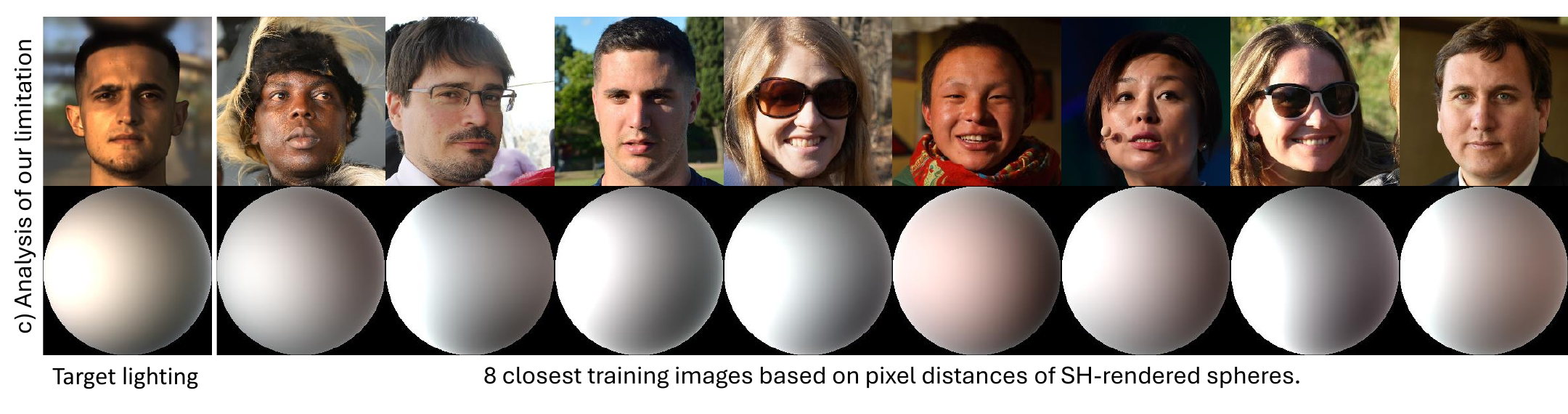}

  \caption{\textbf{Comparison against HoloRelighting \cite{mei2024holo} and visual analysis of our limitations.} a) Our method better preserves fine details, such as hair and teeth, compared to HoloRelighting results, taken directly from their paper due to the lack of source code. Note that our target lighting was estimated using DECA \cite{DECA:Siggraph2021} from the target image. b) The overall lighting in our result lacks the strong orange shading present in the target shading. c) To analyze this issue, we retrieve the 8 closest training images based on SH coefficient distances, revealing that none of them has the orange-tinted shading. This suggests that the current limitation may stem from the lack of such lighting conditions in our FFHQ training data. Nonetheless, our pipeline can readily be trained on additional 2D training images, including extreme lighting conditions, without requiring ground truth.
  %any 2D images for training.
  %Their results and target lighting images were taken directly from their paper. We used DECA \cite{DECA:Siggraph2021} to estimate the target lighting from the target image.
  %In the zoomed-in view, our method preserves finer details, such as hair and teeth. 
  %We present our failure case, where relighting with the target lighting from DECA failed to produce shading similar to the target lighting image. However, we achieved better shading by manually adjusting the target lighting, which emphasizes the inaccuracies of DECA's light estimator. Moreover, we suspect that this failure case could arise from a discrepancy between the target and training lighting. In the visual analysis below, we show the input face, the estimated lighting, and its 8 closest matches in the training set. These results indicate that none of the top 8 closest faces contain similar yellow-tinted shading.}
  % In the second row, it removes cast shadows more effectively (a brightened version is also provided below to highlight the remaining cast shadows). 
  % Despite these improvements, our method does not perfectly match the target lighting or face color shading. We hypothesize that this is due to the limitation of the light estimator, which struggles to disambiguate between skin tone and actual lighting, as shown in Figure \ref{fig_app:light_dist}.}
}
  \label{fig_app:holo_res}
\end{figure*}

\begin{figure*}[]
  \centering
  \includegraphics[scale=0.31]{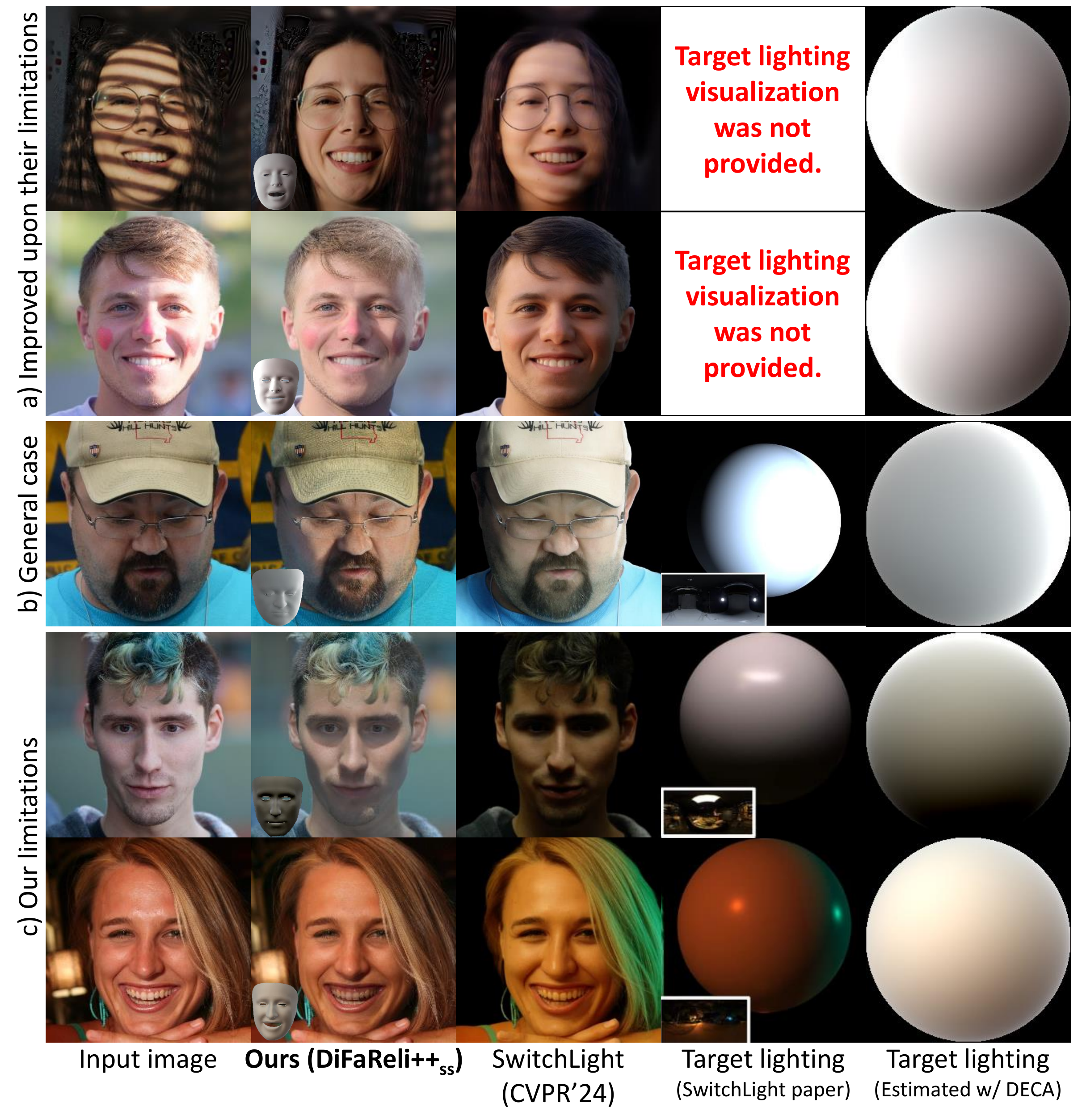}
  \includegraphics[scale=0.45]{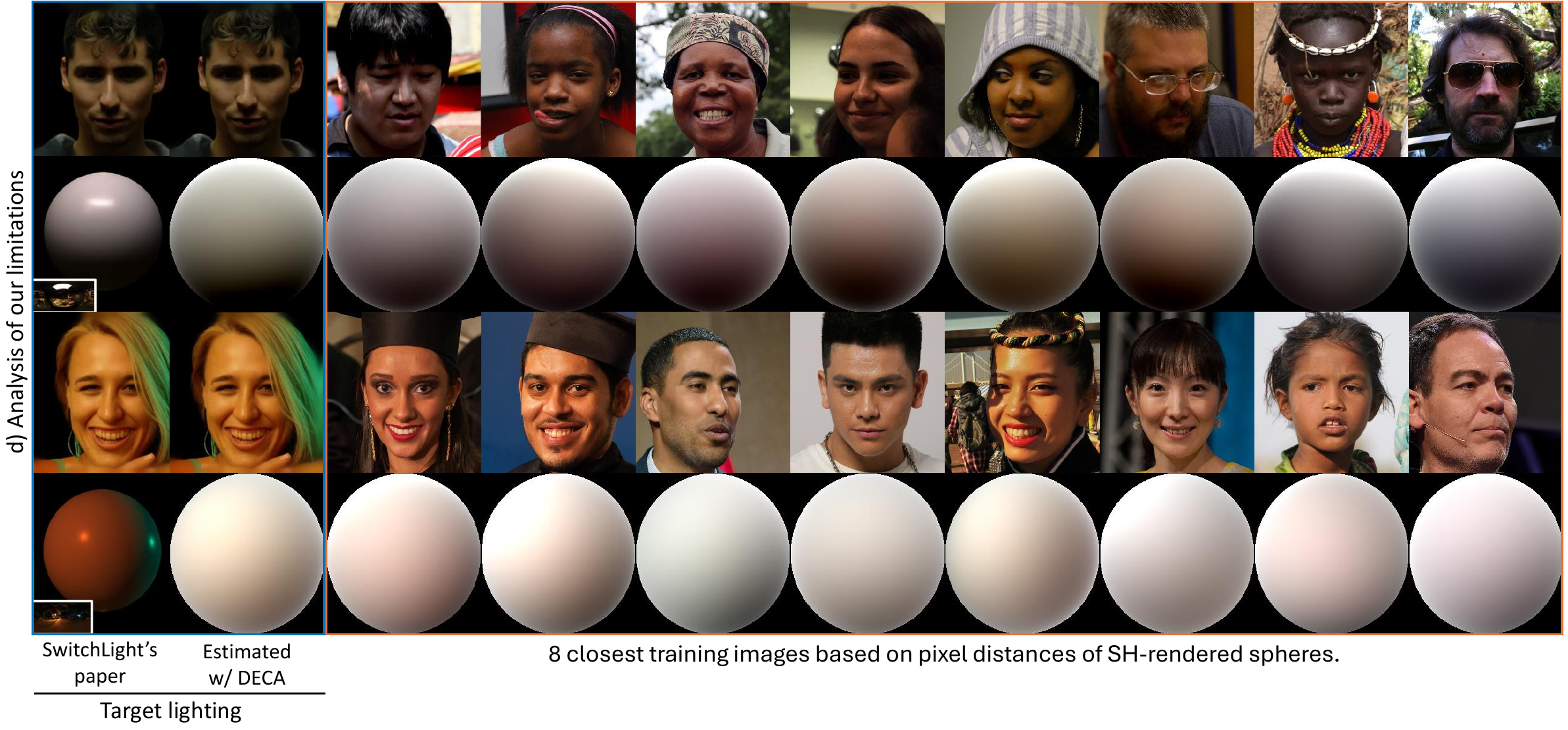}

  \caption{\textbf{Comparison against SwitchLight \cite{kim2024switchlight} and visual analysis of our limitations.} SwitchLight's results were taken directly from their paper due to the lack of source code. Our method addresses SwitchLight's limitations: a) our method effectively removes hard cast shadows and better preserves makeup details, and b) produces sharper details.  c) Our results appear less consistent with the target lighting, lacking the sufficiently dark lighting or the green shading. As in Figure \ref{fig_app:holo_res}, we retrieve the closest training images for these cases. The results show that our light estimator, DECA, struggles with skin-tone and light ambiguity, retrieving dark-skinned individuals instead of images with dark lighting, resulting in a biased training set. Additionally, the retrieved set lacks green lighting, as seen in the test image of the woman, which may explain our difficulties. This can be addressed by scaling up the training set, which only requires 2D images.
  %to do with our framework.
  %, resulting in our inability to reproduce the green shading.
  %and can lead to training samples that do not have dark lighting. 
  %SwitchLight's results were taken directly from their paper. Since no target lighting was provided, we used their results as the target lighting image. Our method improves upon their limitation cases (noted in their paper), as it effectively removes hard cast shadows (first row) and preserves makeup details (second row). 
  %We present our failure cases along with a visual analysis where the target lighting is far from the training lighting distribution (e.g., extremely dark faces or orange-greenish hues). In each row, we show the input face, the estimated lighting, and its 8 closest matches in the training set. These results reveal that none of the top 8 closest matches exhibit similarly extreme dark lighting or orange-green hues. Additionally, these issues may stem from inaccuracies in DECA's light estimator, which fails to replicate the orange-tinted or greenish hues seen in the target lighting comparison in the first two columns of the analysis.
}
  \label{fig_app:switch_res}
\end{figure*}

\begin{figure*}[]
  \centering
  \includegraphics[scale=0.6]{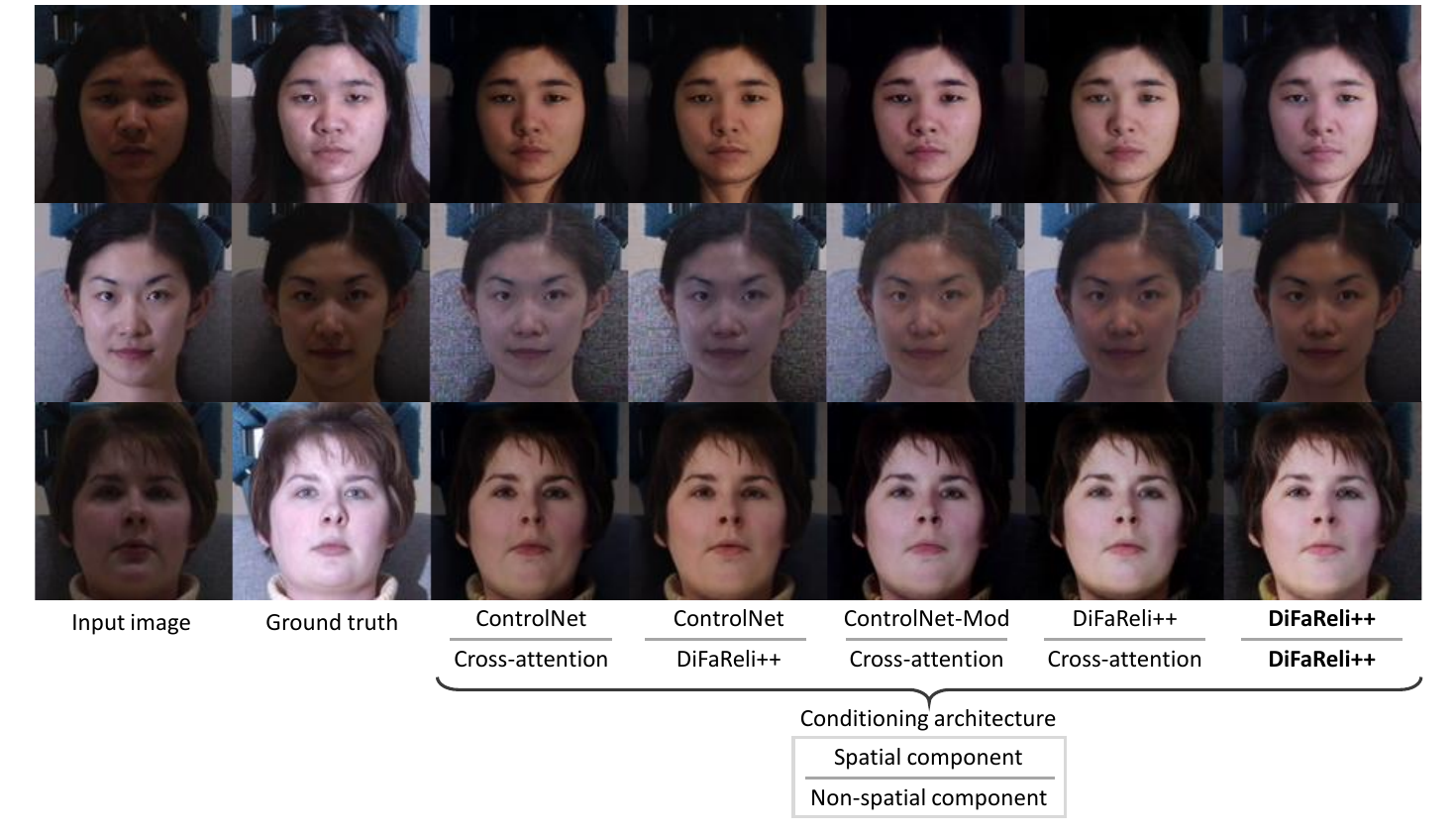}
  \vspace{-0.5em}
  \caption{\textbf{Ablation study of the conditioning mechanisms on MultiPIE}~\cite{gross2010multi} (Section \ref{sec:controlnet_ab} in the main paper).}
  \label{fig_app:controlnet_aba}
  % \vspace{-1.0em}
\end{figure*}

\begin{figure*}[]
  \centering
  \includegraphics[scale=0.54]{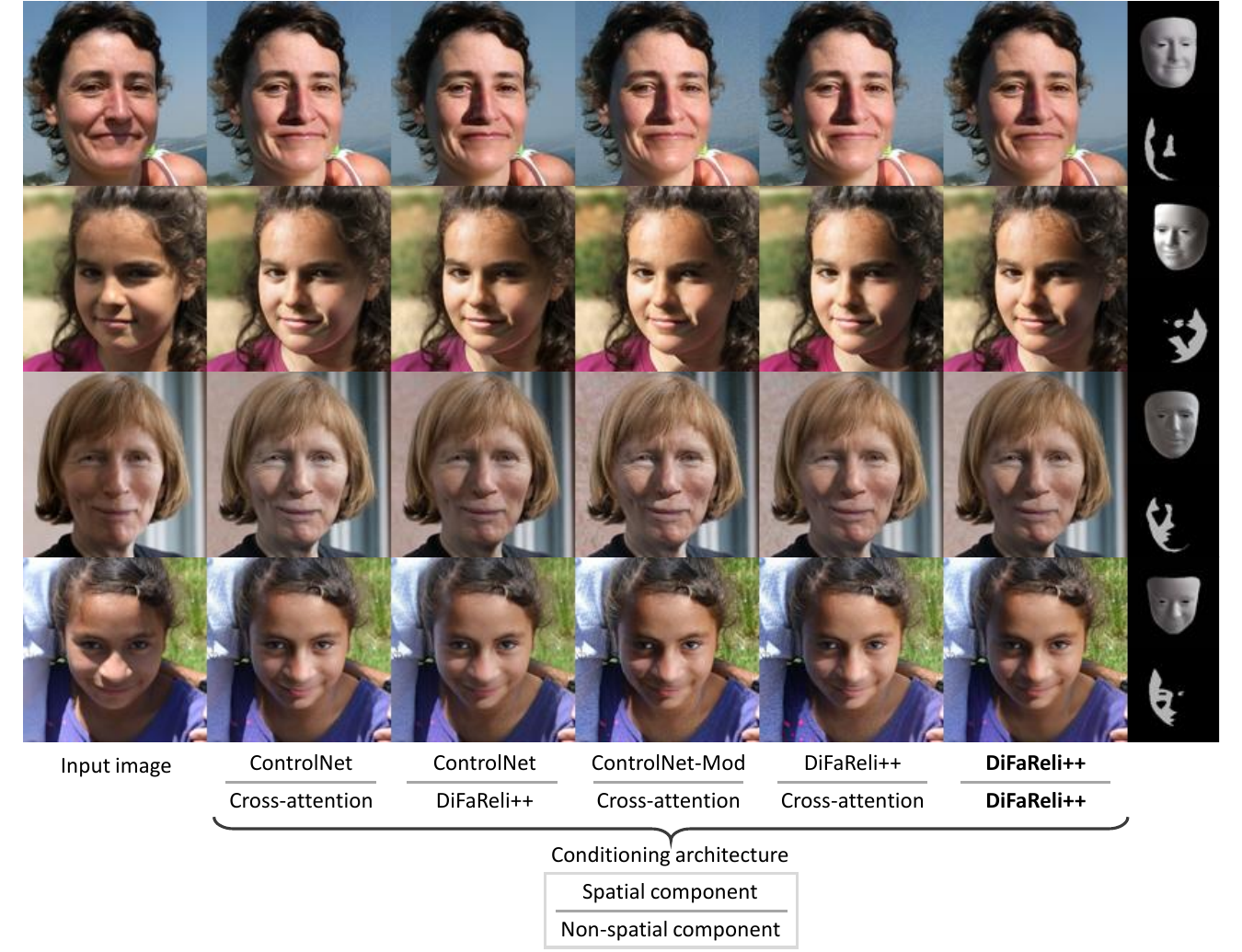}
  \vspace{-0.5em}
  \caption{\textbf{Ablation study of the conditioning mechanisms on FFHQ}~\cite{karras2019style} (Section \ref{sec:controlnet_ab} in the main paper).}
  \label{fig_app:controlnet_aba_cs}
  % \vspace{-1.0em}
\end{figure*}

\begin{figure*}[]
  \centering
  \includegraphics[scale=0.53]{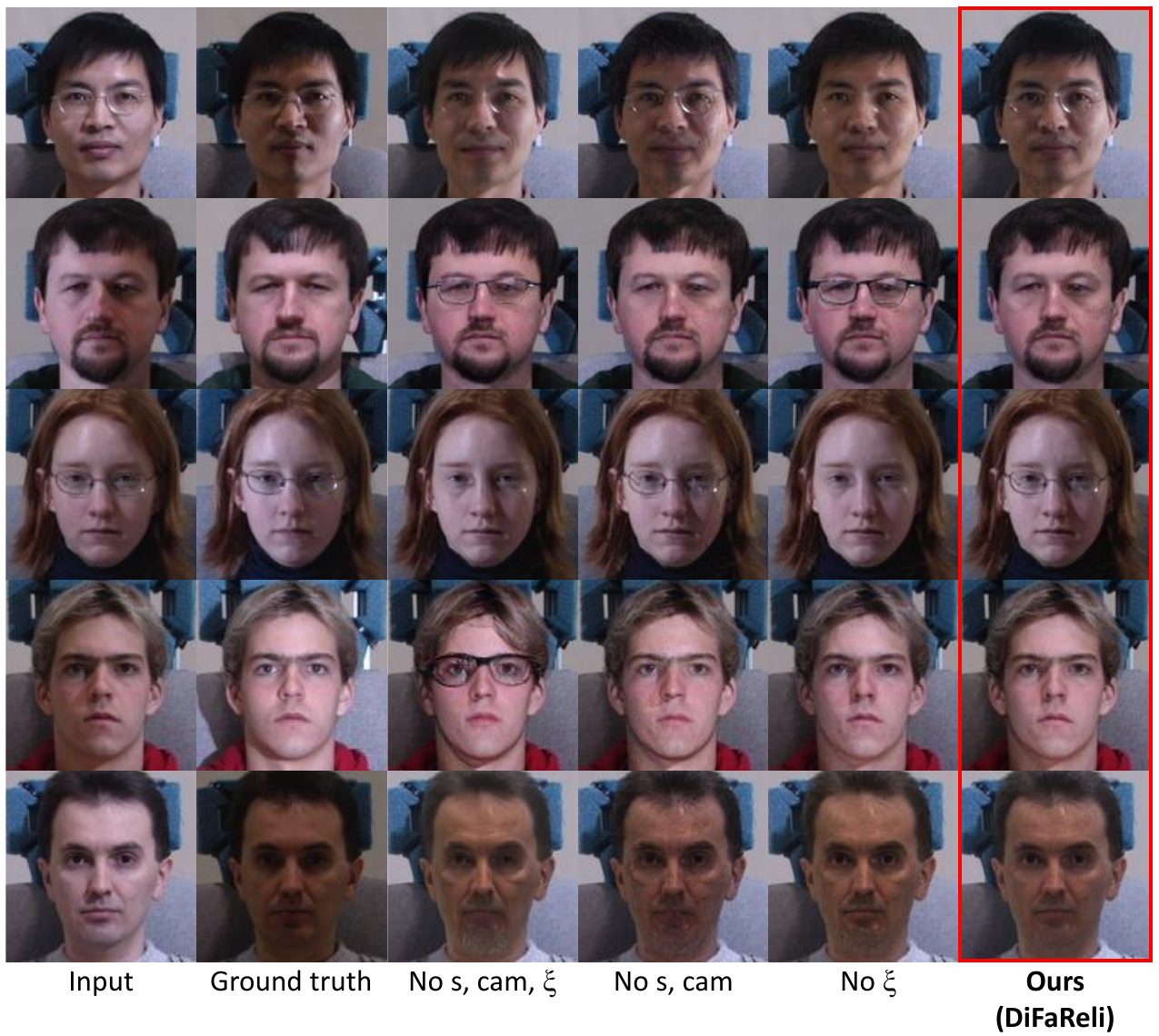}
  \vspace{-0.5em}
  \caption{\textbf{Ablation study of the light conditioning} (Section \ref{ab:light_cond} in Appendix).}
  \label{fig:aba_light}
  % \vspace{-1.0em}
\end{figure*}

\begin{figure*}[]
  \centering
  \includegraphics[scale=0.53]{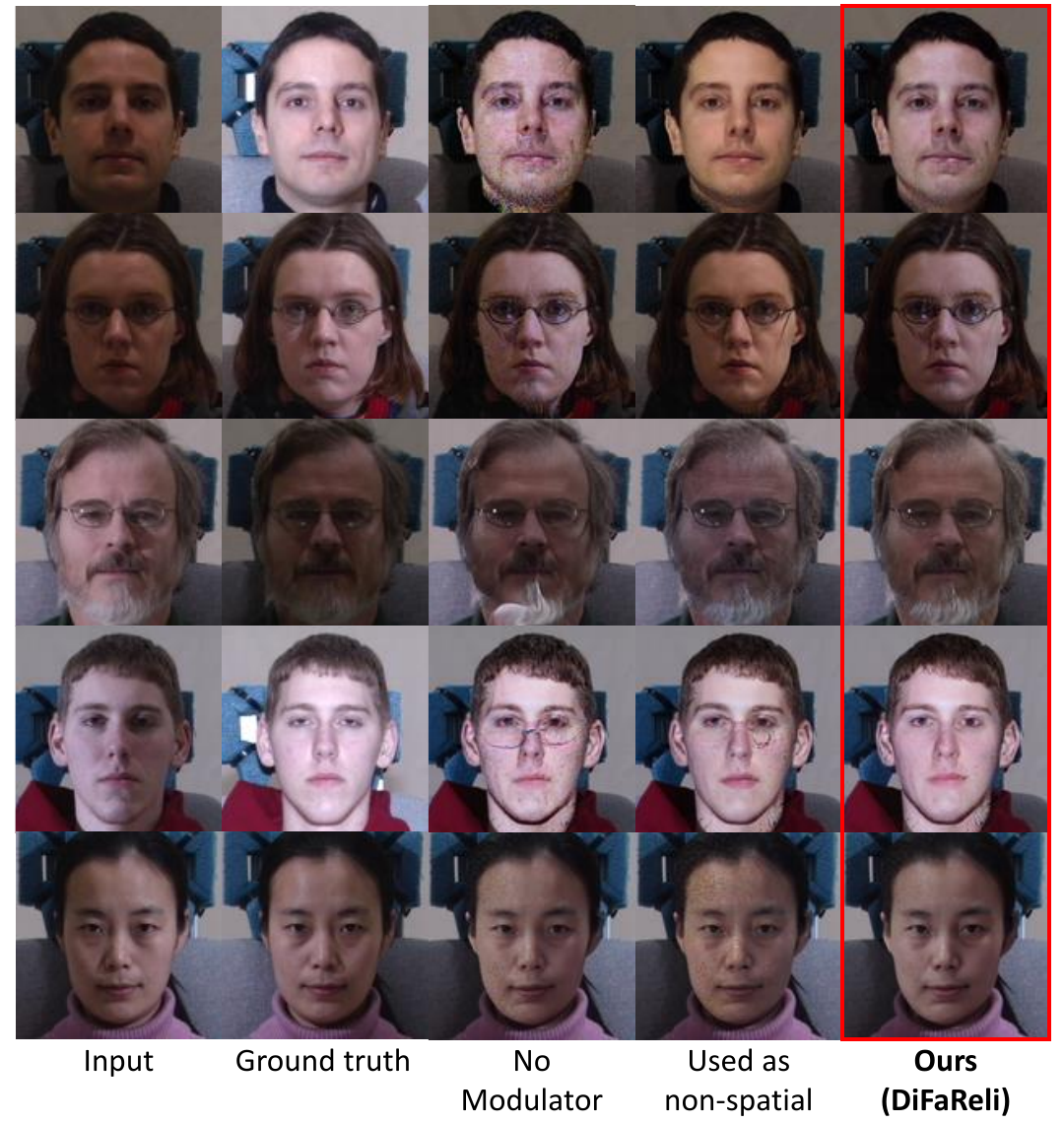}
  \vspace{-0.5em}
  \caption{\textbf{Ablation study of the non-spatial conditioning variable} (Section \ref{ab:nonspa} in Appendix).}
  \label{fig:aba_nonspa}
  \vspace{-1.0em}
\end{figure*}

We provide a qualitative comparison with HoloRelighting in Figure \ref{fig_app:holo_res} and SwitchLight in Figure \ref{fig_app:switch_res}. Compared to HoloRelighting \cite{mei2024holo}, our method preserves finer details like hair and teeth (first and third rows) and more effectively removes cast shadows from the input images (second rows).
We also address the limitations discussed in SwitchLight~\cite{kim2024switchlight}, showing improved removal of hard cast shadows (first row) and preserving makeup from the input image (second row). 

Despite these improvements, our relighting results might not perfectly match the target lighting, as our method relies on the DECA light estimator~\cite{DECA:Siggraph2021}. 
As shown in Figure \ref{fig_app:holo_res} and \ref{fig_app:switch_res},
%This light estimator can be inaccurate as shown in the bottom of Figure \ref{fig_app:holo_res} and \ref{fig_app:switch_res}, 
the estimated lighting from DECA fails to replicate the shading colors in the input images (e.g., yellow-tinted or greenish tones).
We also suspect a discrepancy between the target lighting and the lighting conditions in our training set.

To test this, we performed a visual analysis of our failure cases, shown below each qualitative comparison in Figure \ref{fig_app:holo_res} and \ref{fig_app:switch_res}.
Each row shows the top eight closest lighting examples, retrieved by computing pixel-wise L1 distances between rendered sphere images from the FFHQ training set and the target lighting image. The spheres were rendered with estimated spherical harmonics (SH) from DECA \cite{DECA:Siggraph2021}, using the FFHQ training set as the retrieval set and the target lighting image (from \cite{mei2024holo, kim2024switchlight}) as the query. Areas outside the sphere were masked during computation. 
This analysis shows that the training set used to train DiFaReli++ does not contain similar yellow-tinted or greenish effects, even in the top closest lighting examples. 

Although our method relies on this light estimator, it is designed to be flexible and can easily integrate with any state-of-the-art light estimator to address such cases. However, there are currently few publicly available light estimators, especially for portrait images, and this area of research remains under active development.

\iffalse
\begin{figure*}[]
  \centering
  % \includegraphics[scale=0.53]{./figures/ph.pdf}
  \includegraphics[scale=0.53]{./figures/aba_arch_ss.pdf}
  \caption{\textbf{Ablation study on architecture choices for our single-shot relighting network} (Section \ref{ab:arch_ss} in the main text).}
  \label{fig:aba_arch_ss}
\end{figure*}
\fi

\begin{figure*}
  \centering
  \adjustbox{center}{
    \includegraphics[scale=0.4]{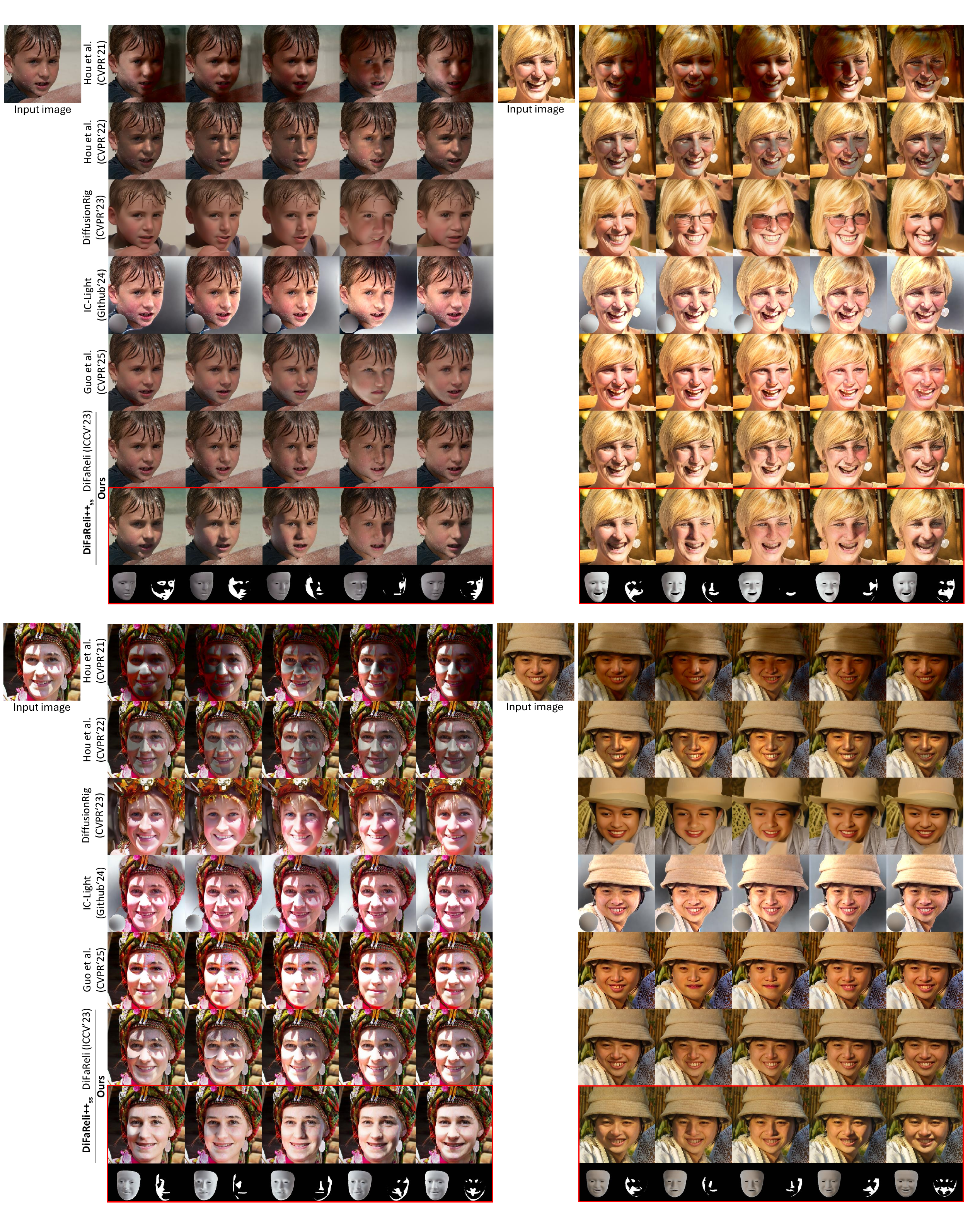}
    }

  \caption{\textbf{Relit results under rotating light around the forward axis (roll) on the FFHQ test set \cite{karras2019style}.}}
  \label{fig:rotate_cs_app_1}
\end{figure*}

\begin{figure*}
  \centering
  \adjustbox{center}{
    \includegraphics[scale=0.4]{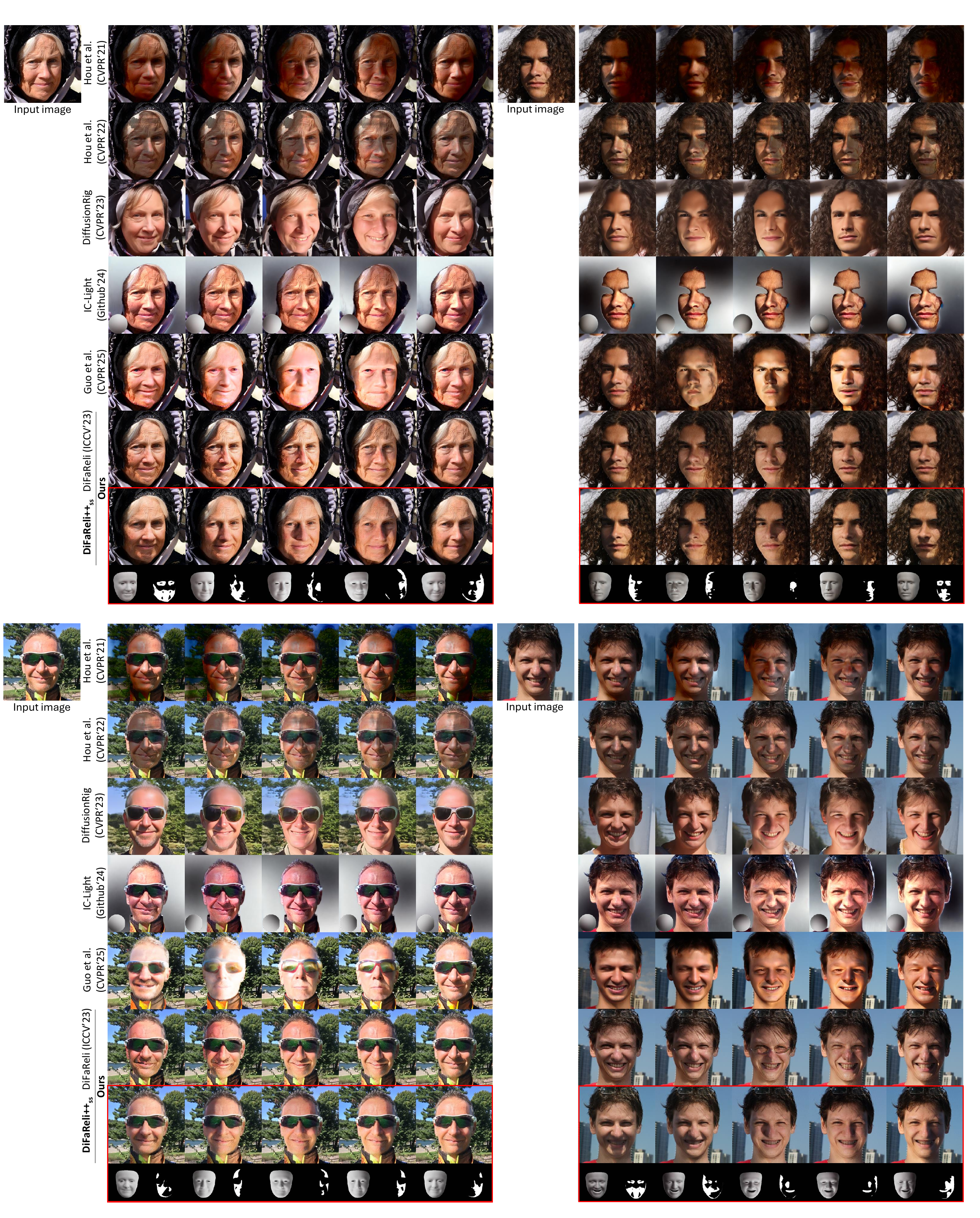}

    }

  \caption{\textbf{Relit results under rotating light around the forward axis (roll) on the FFHQ test set \cite{karras2019style}.}}
  \label{fig:rotate_cs_app_2}
\end{figure*}

\begin{figure*}
  \centering
  \adjustbox{center}{
    \includegraphics[scale=0.4]{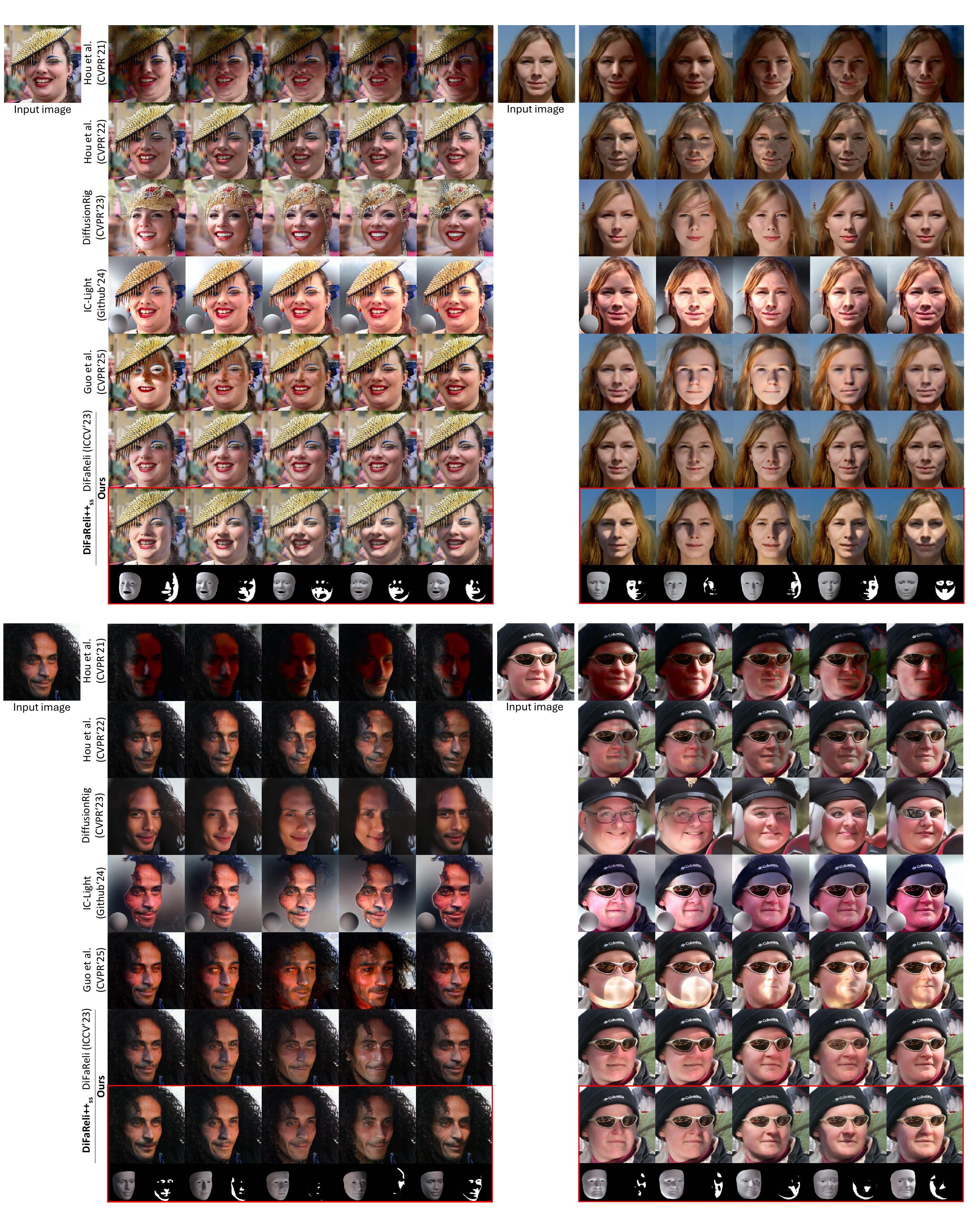}
    }

  \caption{\textbf{Relit results under rotating light around the forward axis (roll) on the FFHQ test set \cite{karras2019style}.}}
  \label{fig:rotate_cs_app_3}
\end{figure*}

\begin{figure*}
  \centering
  \adjustbox{center}{
    \includegraphics[scale=0.4]{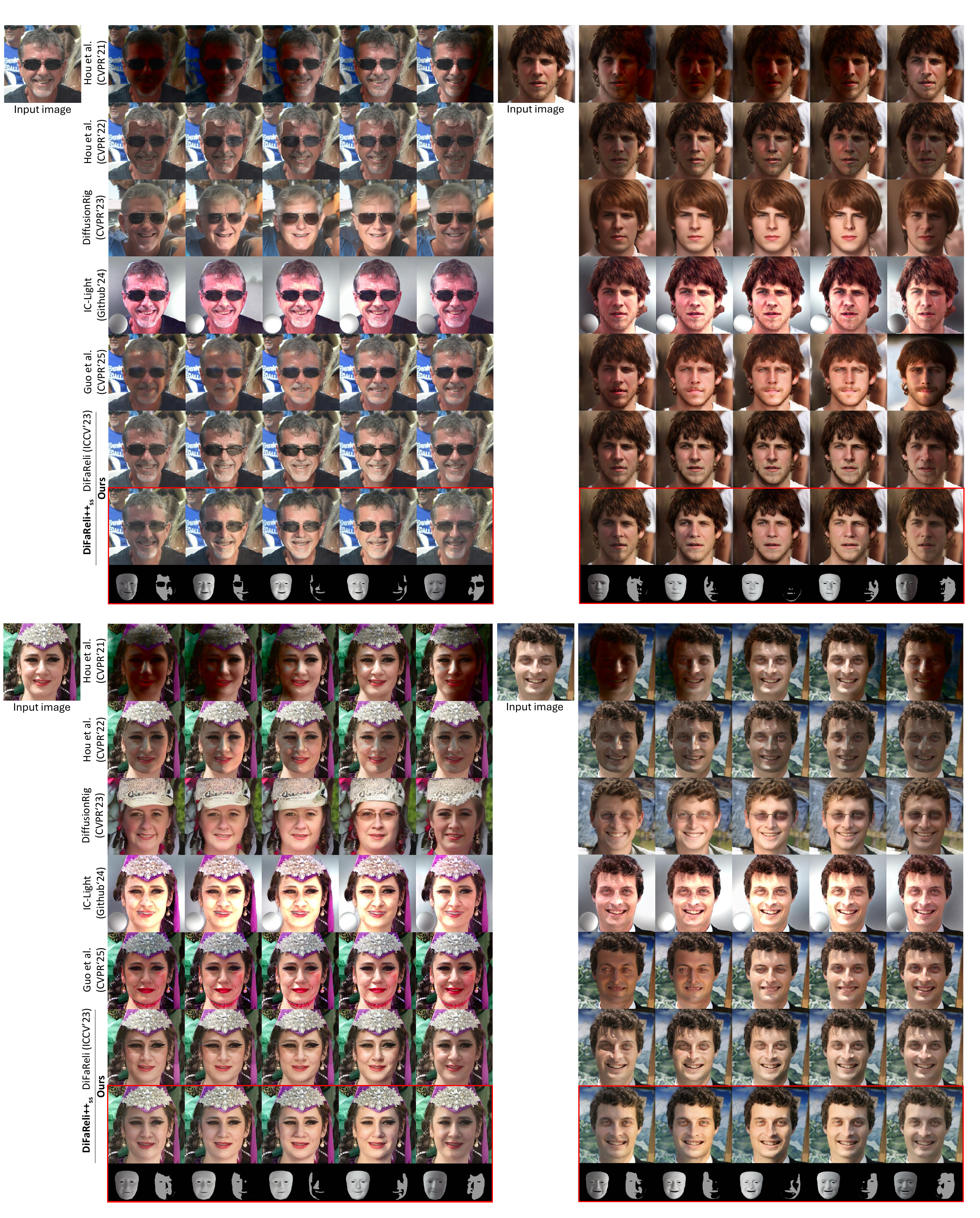}
    }

  \caption{\textbf{Relit results under rotating light around the up axis (yaw) on the FFHQ test set \cite{karras2019style}.} }
  \label{fig:rotate_cs_app_4}
\end{figure*}

\begin{figure*}
  \centering
  \adjustbox{center}{
    \includegraphics[scale=0.4]{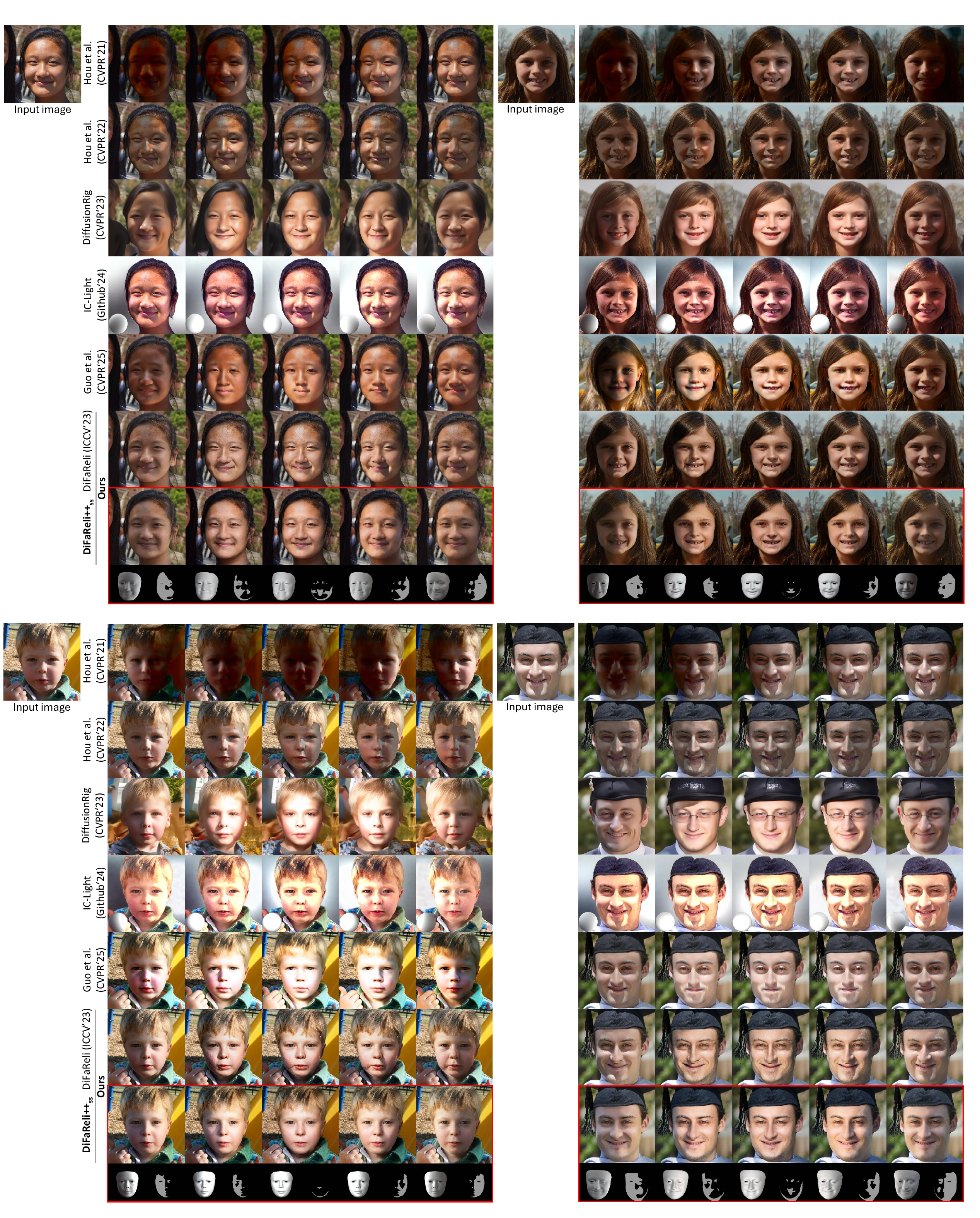}
    }

  \caption{\textbf{Relit results under rotating light around the up axis (yaw) on the FFHQ test set \cite{karras2019style}.}}
  \label{fig:rotate_cs_app_5}
\end{figure*}

\begin{figure*}[]
  \centering
  \vspace{-1.0cm}
  \includegraphics[width=\textwidth]{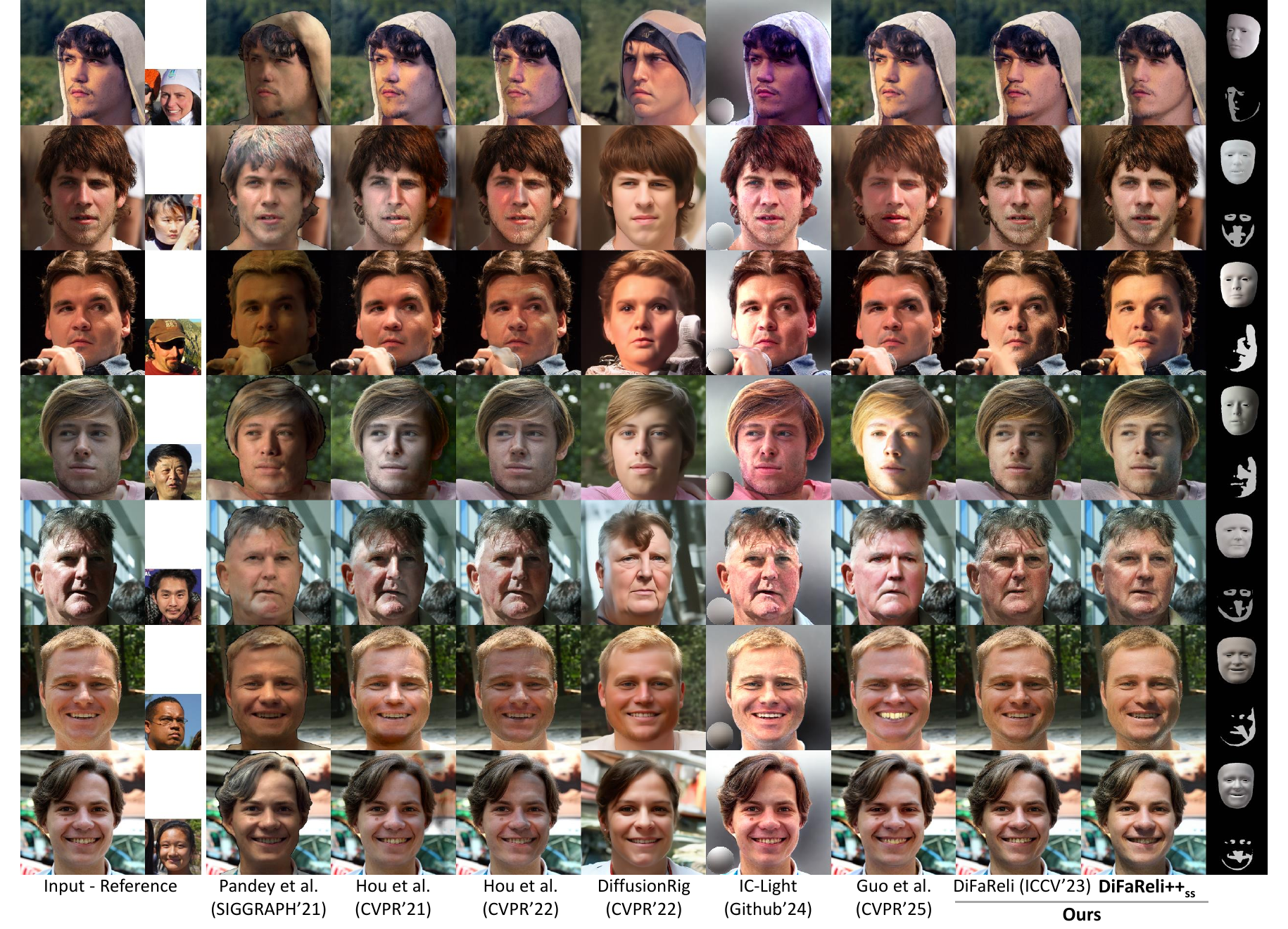}
  \caption{\textbf{Relit results under a given target lighting image on the FFHQ test set \cite{karras2019style}.}}
  \label{fig_app:ffhq_cs_app1}
\end{figure*}

\begin{figure*}[]
  \centering
  \vspace{-1.0cm}
  \includegraphics[width=\textwidth]{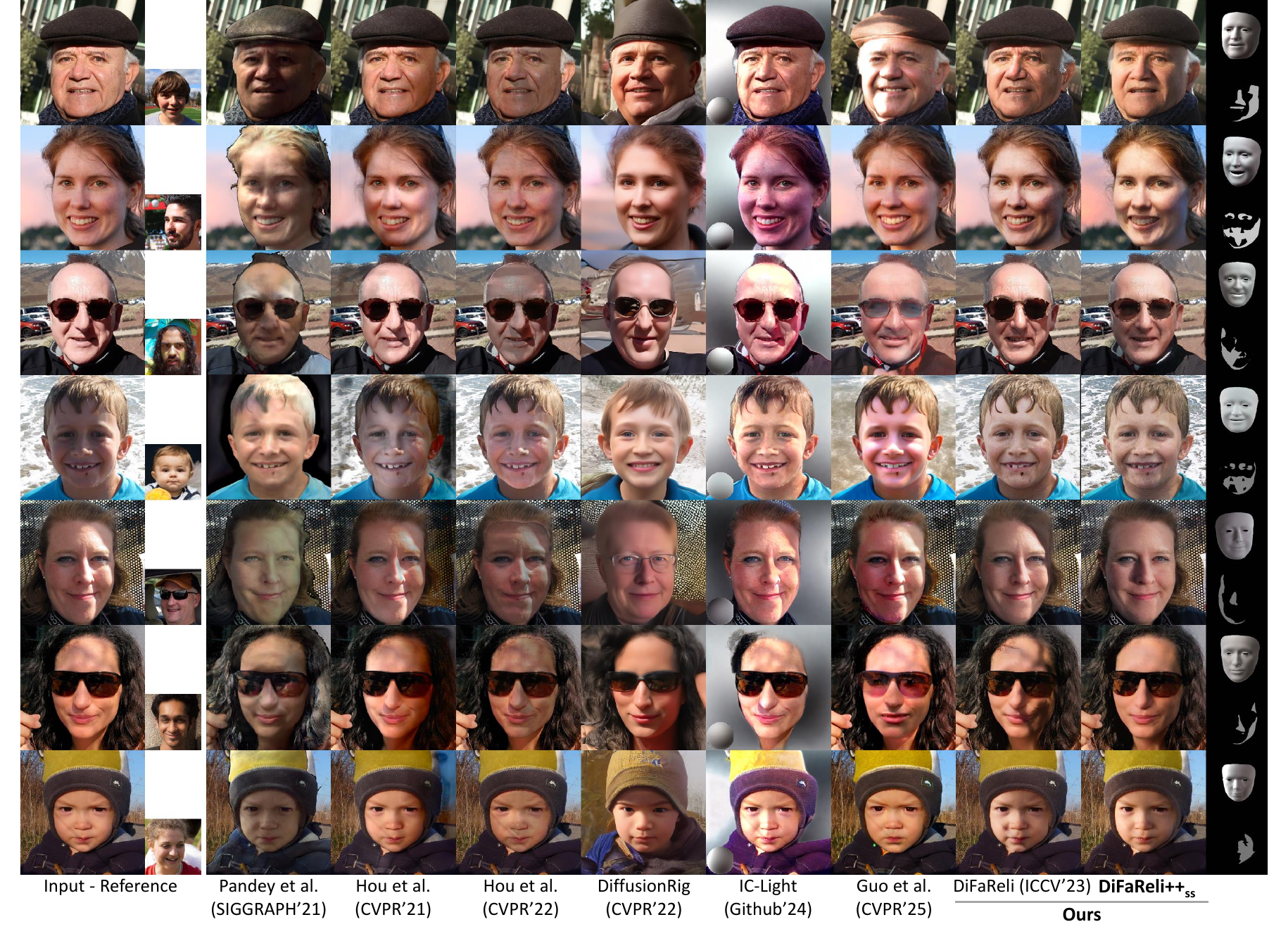}
  \caption{\textbf{Relit results under a given target lighting image on the FFHQ test set \cite{karras2019style}.}}
  \label{fig_app:ffhq_cs_app2}
\end{figure*}

\begin{figure*}[]
  \centering
  \vspace{-1.0cm}
  \includegraphics[width=\textwidth]{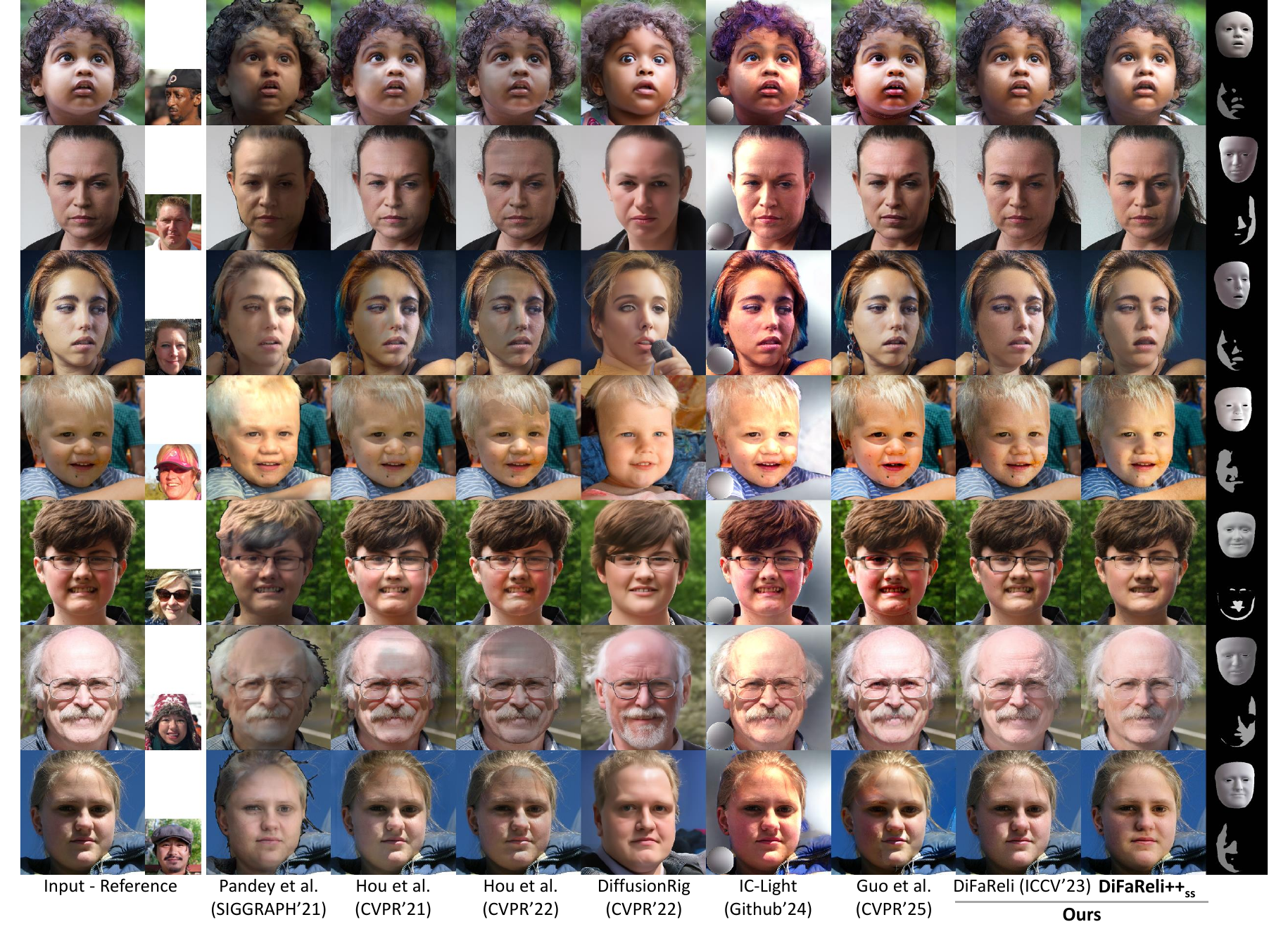}

  \caption{\textbf{Relit results under a given target lighting image on the FFHQ test set \cite{karras2019style}.}}
  \label{fig_app:ffhq_cs_app3}
\end{figure*}

\begin{figure*}[]
  \centering
  \vspace{-1.0cm}
  \includegraphics[width=\textwidth]{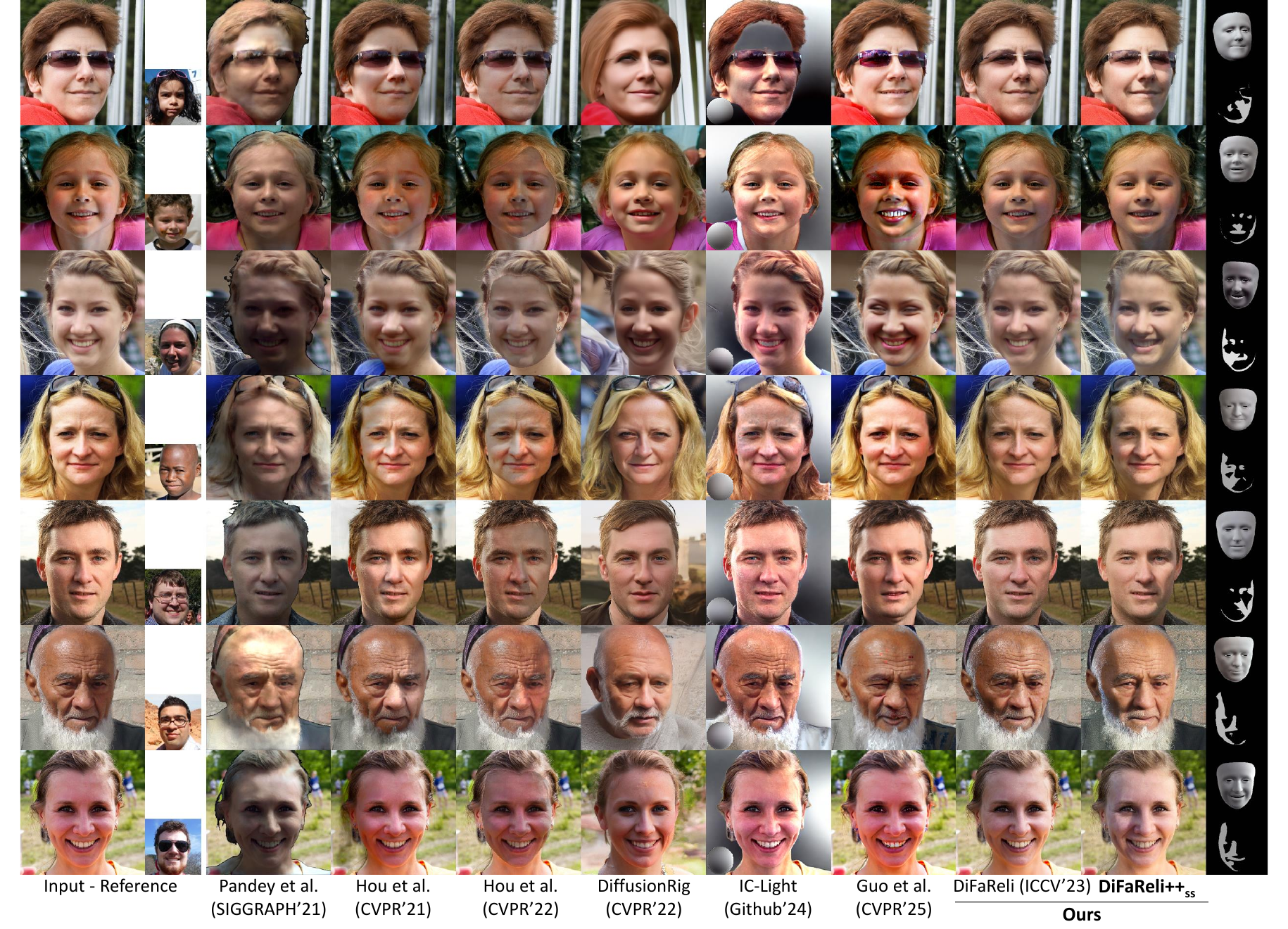}

  \caption{\textbf{Relit results under a given target lighting image on the FFHQ test set \cite{karras2019style}.}}
  \label{fig_app:ffhq_cs_app4}
\end{figure*}

\begin{figure*}[]
  \centering
  \vspace{-1.0cm}
  \includegraphics[width=\textwidth]{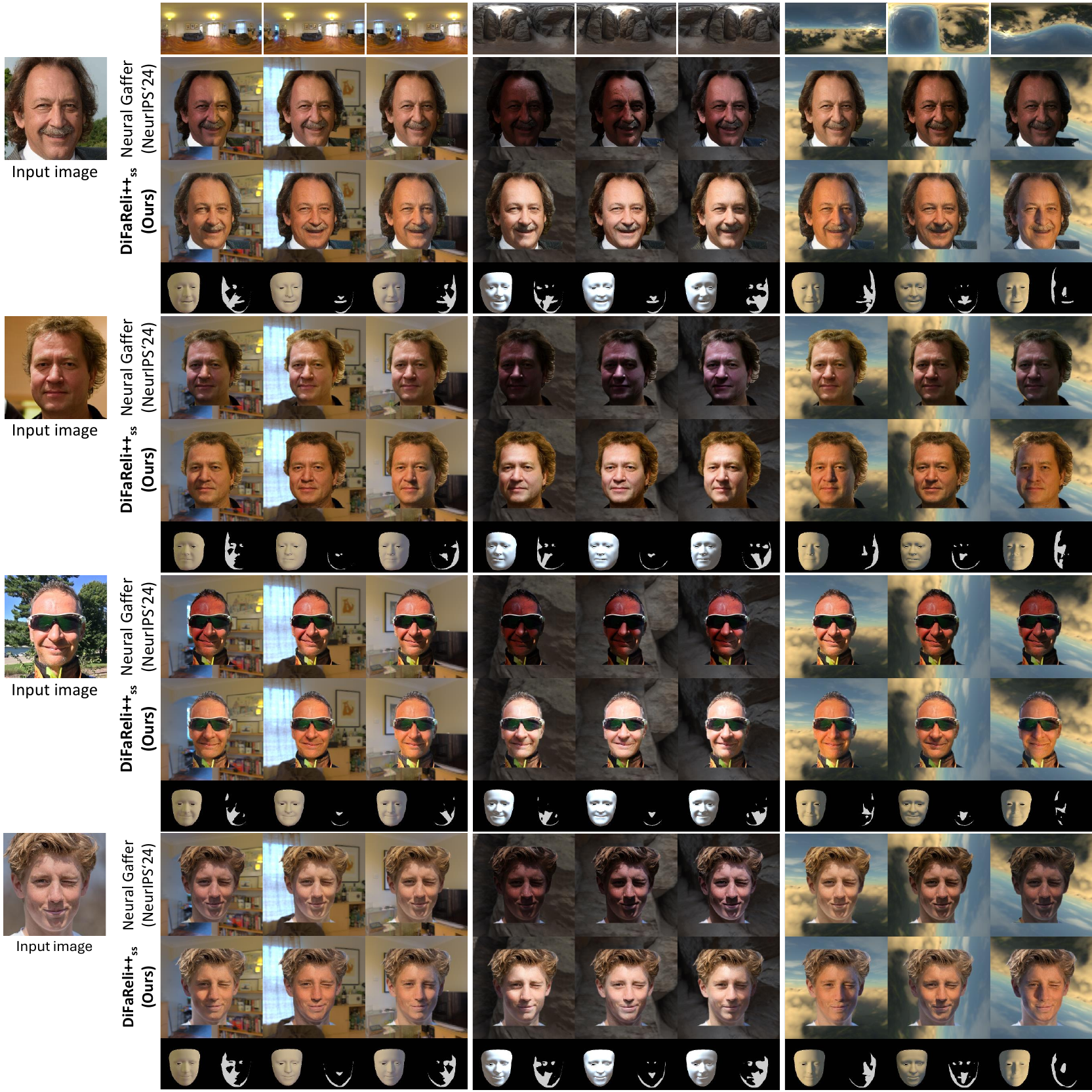}

  \caption{\textbf{Relit results under rotating HDR environment maps on the FFHQ test set \cite{karras2019style}.}}
  \label{fig_app:ffhq_cs_hdr_app1}
\end{figure*}

\begin{figure*}[]
  \centering
  \vspace{-1.0cm}
  \includegraphics[width=\textwidth]{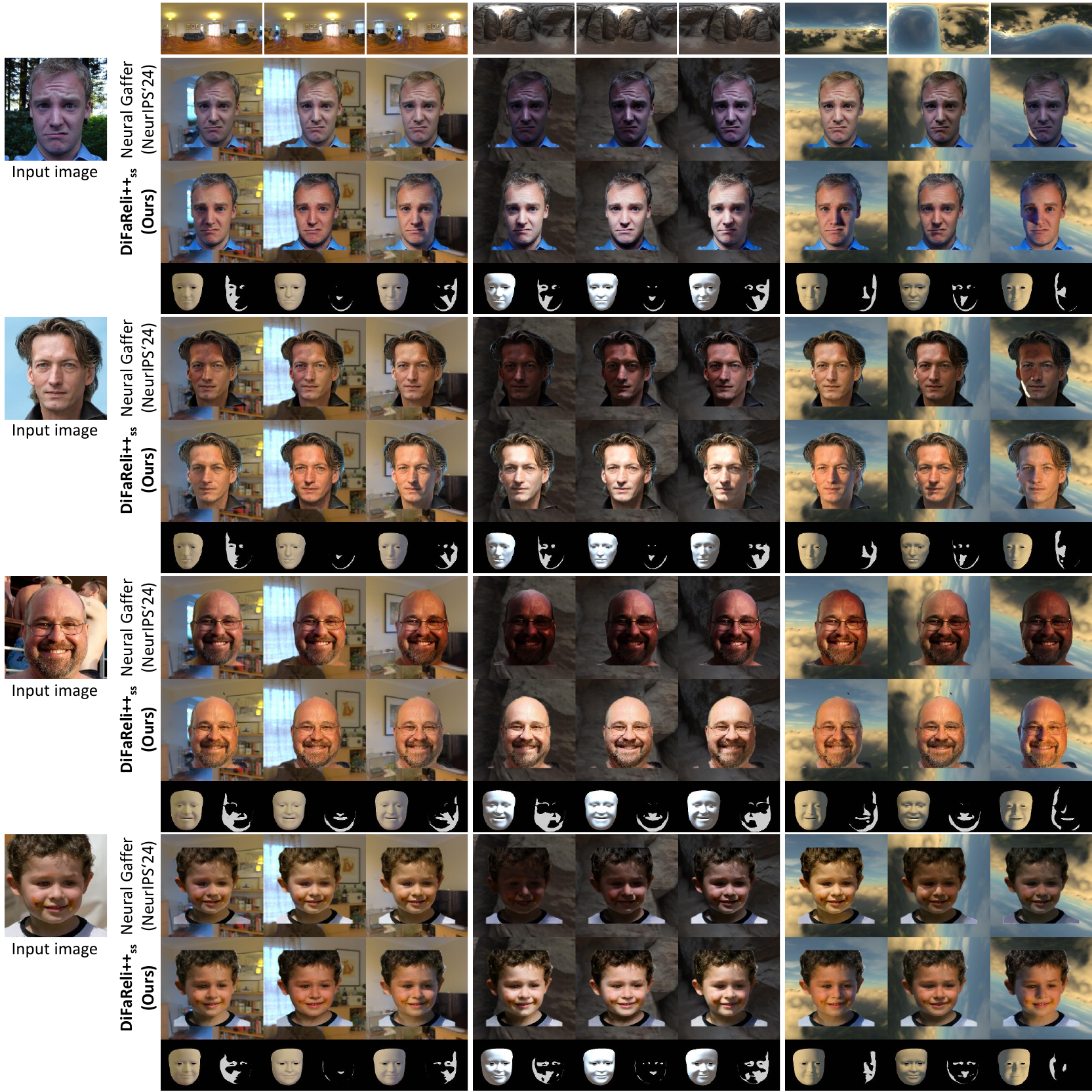}

  \caption{\textbf{Relit results under rotating HDR environment maps on the FFHQ test set \cite{karras2019style}.}}
  \label{fig_app:ffhq_cs_hdr_app2}
\end{figure*}

\begin{figure*}[]
  \centering
  \vspace{-1.0cm}
  \includegraphics[width=0.9\textwidth]{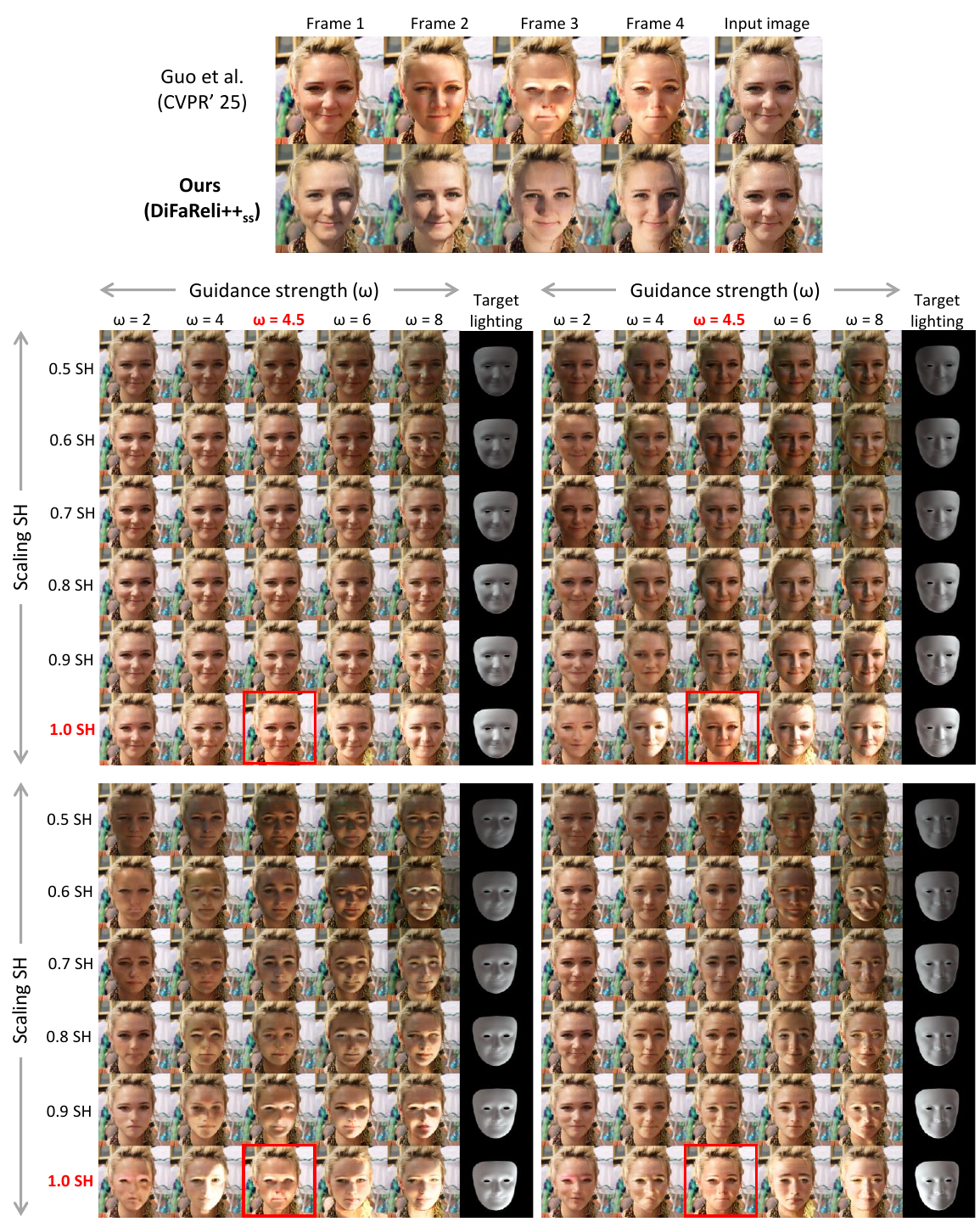}
  \caption{\textbf{Tuning the relighting hyperparameters of Guo et al.~\cite{Guo_2025_CVPR} (under moving lights).}
  We performed a grid search over the guidance strengths ($\omega$) proposed by \cite{Guo_2025_CVPR} (see Table~5 in their supplementary material), along with scaling the SH values, to ensure that the oversaturation artifacts were not due to suboptimal hyperparameters. Results obtained with the default hyperparameters are highlighted with a red box.}
  \label{fig_app:tuninggrid_relipa_rot_sj1}
\end{figure*}

\begin{figure*}[]
  \centering
  \vspace{-1.0cm}
  \includegraphics[width=0.9\textwidth]{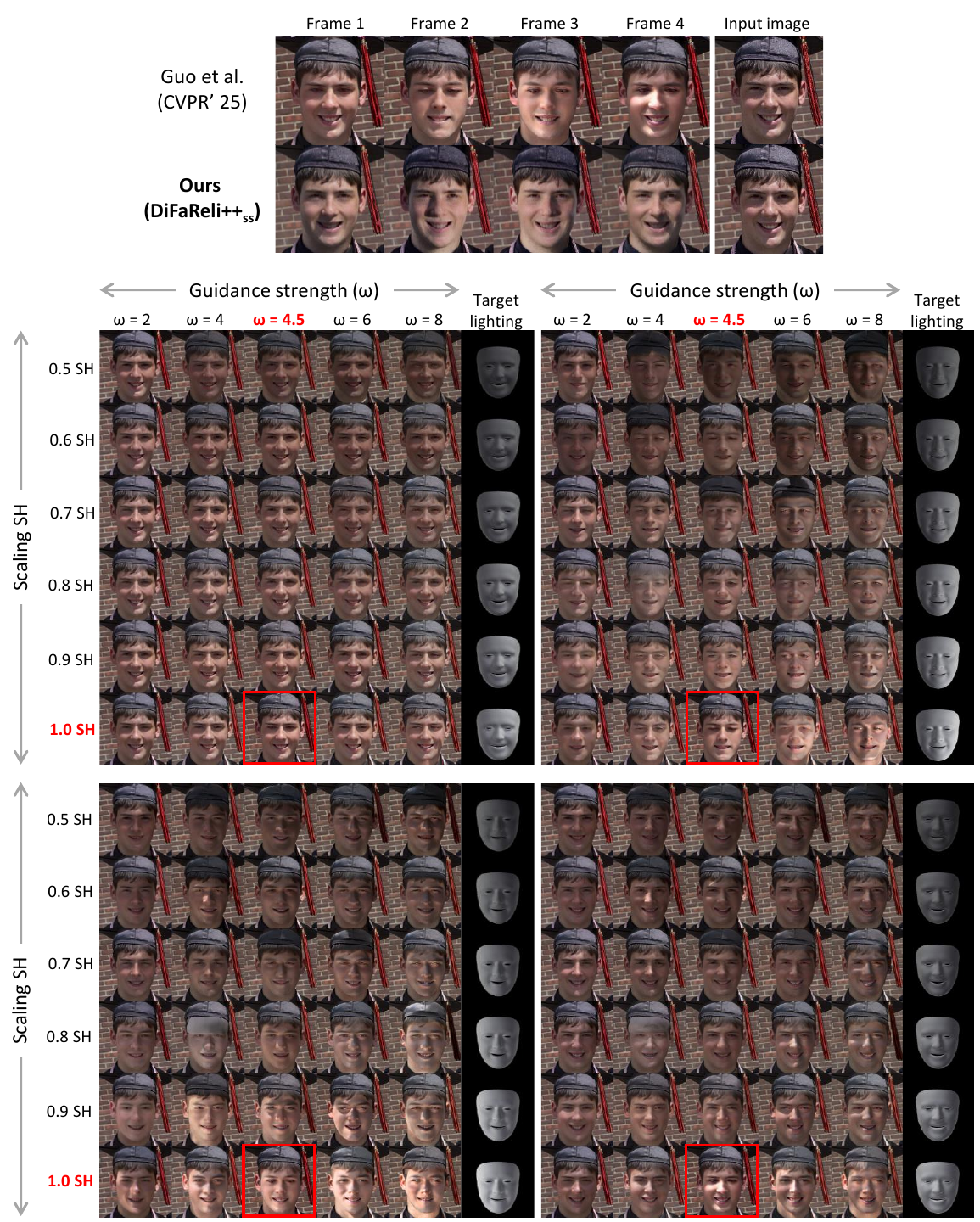}
  \caption{\textbf{Tuning the relighting hyperparameters of Guo et al.~\cite{Guo_2025_CVPR} (under moving lights).}
  %We applied the same relighting hyperparameter tuning procedure as described in Figure~\ref{fig_app:tuninggrid_relipa_rot_sj1}. Results obtained with the default parameters reported in their paper are highlighted with a red box.
 We performed a grid search over the guidance strengths ($\omega$) proposed by \cite{Guo_2025_CVPR} (see Table~5 in their supplementary material), along with scaling the SH values, to ensure that the oversaturation artifacts were not due to suboptimal hyperparameters. Results obtained with the default hyperparameters are highlighted with a red box. 
  }
  \label{fig_app:tuninggrid_relipa_rot_sj2}
\end{figure*}

\begin{figure*}[]
  \centering
  \vspace{-1.0cm}
  \includegraphics[width=0.7\textwidth]{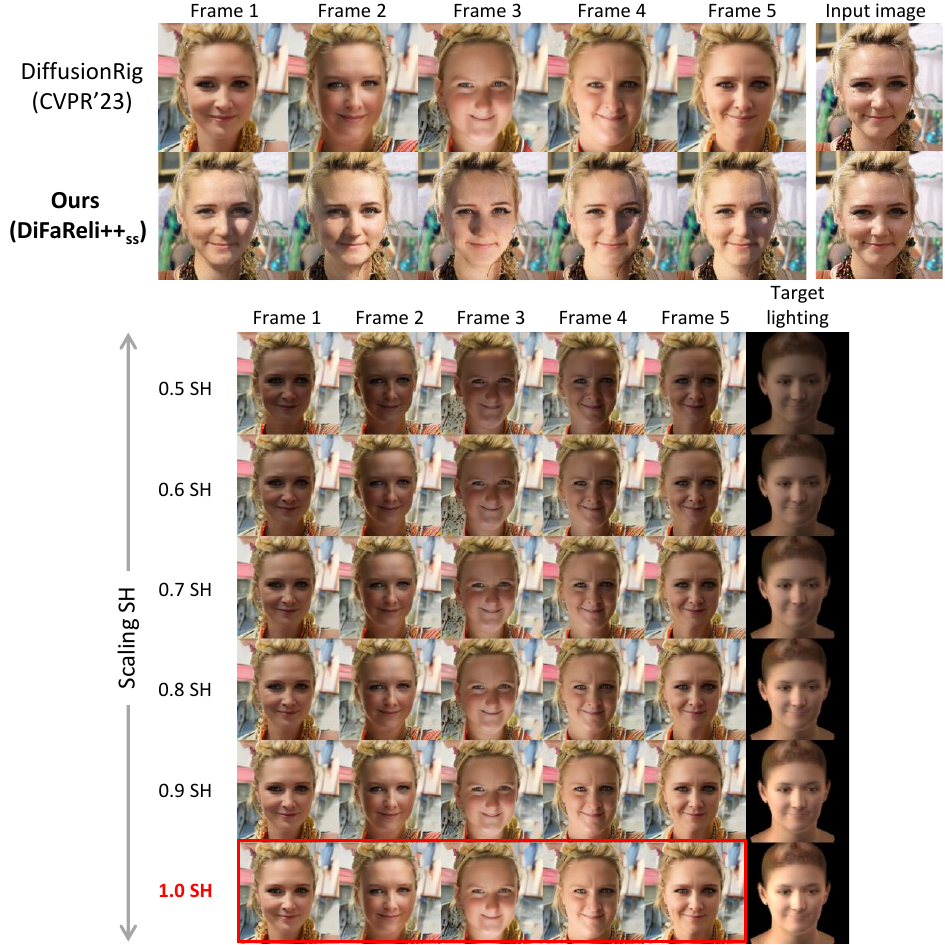}
  \caption{\textbf{Tuning the relighting hyperparameters of DiffusionRig~\cite{ding2023diffusionrig} (under moving lights).} 
  We performed a grid search over different scaled SH values to ensure that the oversaturation artifacts were not due to suboptimal tuning. Results obtained with the default parameters are highlighted with a red box.}
  \label{fig_app:tuninggrid_diffusionrig_rot_sj1}
\end{figure*}

\begin{figure*}[]
  \centering
  \vspace{-1.0cm}
  \includegraphics[width=0.7\textwidth]{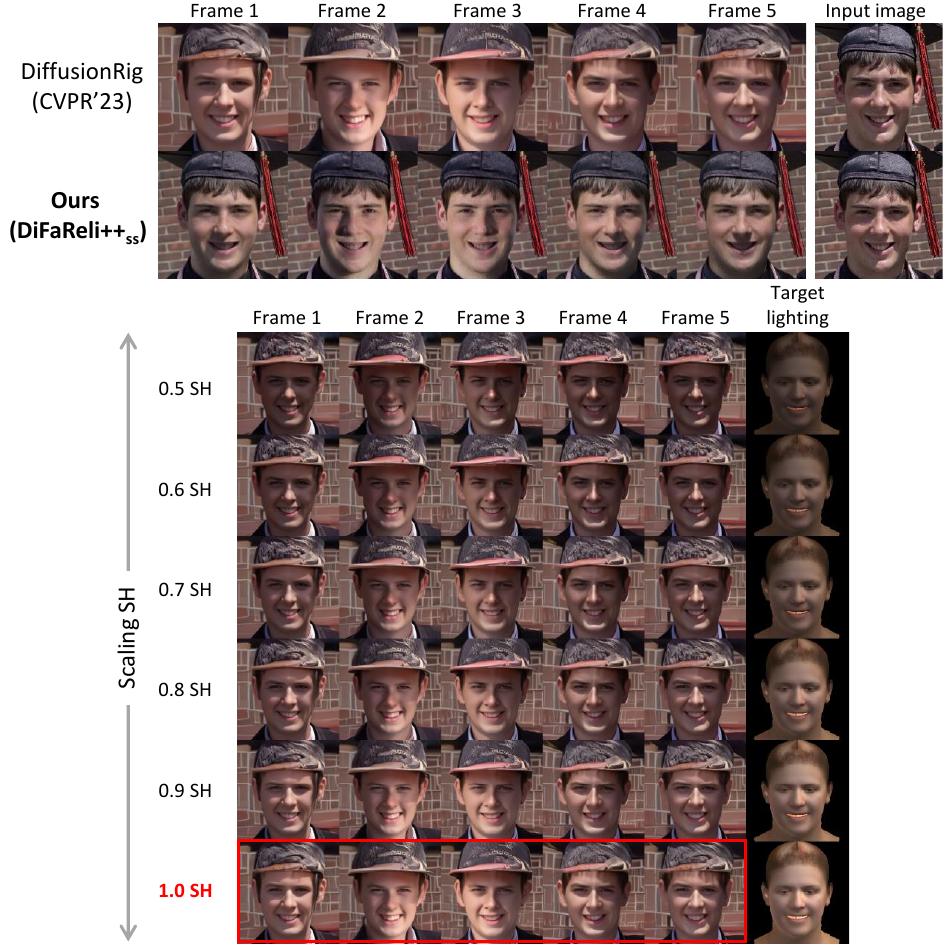}
  \caption{\textbf{Tuning the relighting hyperparameters of DiffusionRig~\cite{ding2023diffusionrig} (under moving lights).}
  We applied the same relighting hyperparameter tuning procedure as described in Figure~\ref{fig_app:tuninggrid_diffusionrig_rot_sj1}. Results obtained with the default parameters are highlighted with a red box.
  }
  \label{fig_app:tuninggrid_diffusionrig_rot_sj2}
\end{figure*}

\begin{figure*}[]
  \centering
  \vspace{-1.0cm}
  \includegraphics[width=0.85\textwidth]{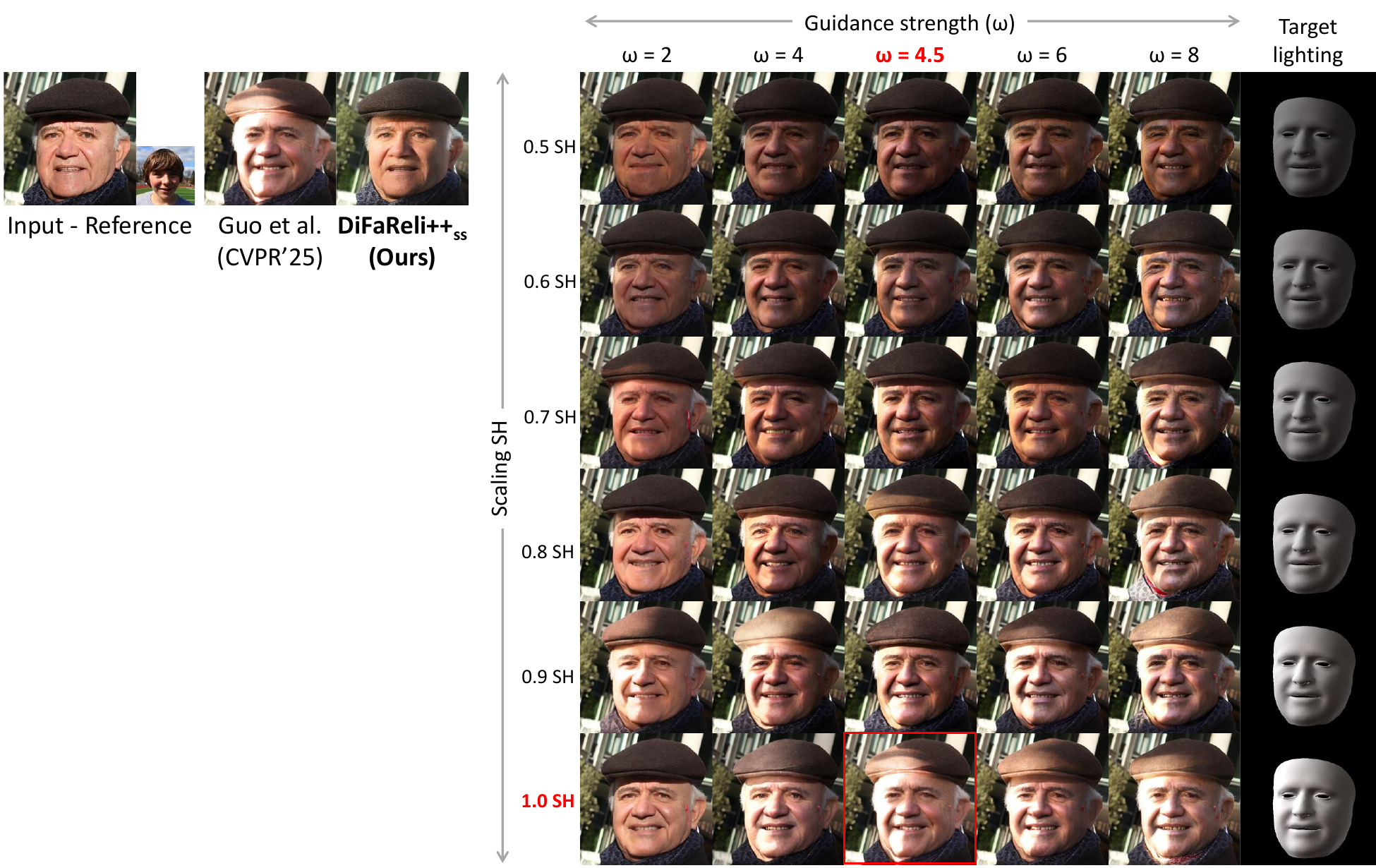}
  \caption{\textbf{Tuning the relighting hyperparameters of Guo et al.~\cite{Guo_2025_CVPR} (under a target lighting image).}
  We applied the same relighting hyperparameter tuning procedure as described in Figure~\ref{fig_app:tuninggrid_relipa_rot_sj1}. Results obtained with the default parameters reported in their paper is highlighted with a red box.
  }
  \label{fig_app:tuninggrid_relipa_targetSH_sj1}
\end{figure*}

\begin{figure*}[]
  \centering
  \vspace{-1.0cm}
  \includegraphics[width=0.85\textwidth]{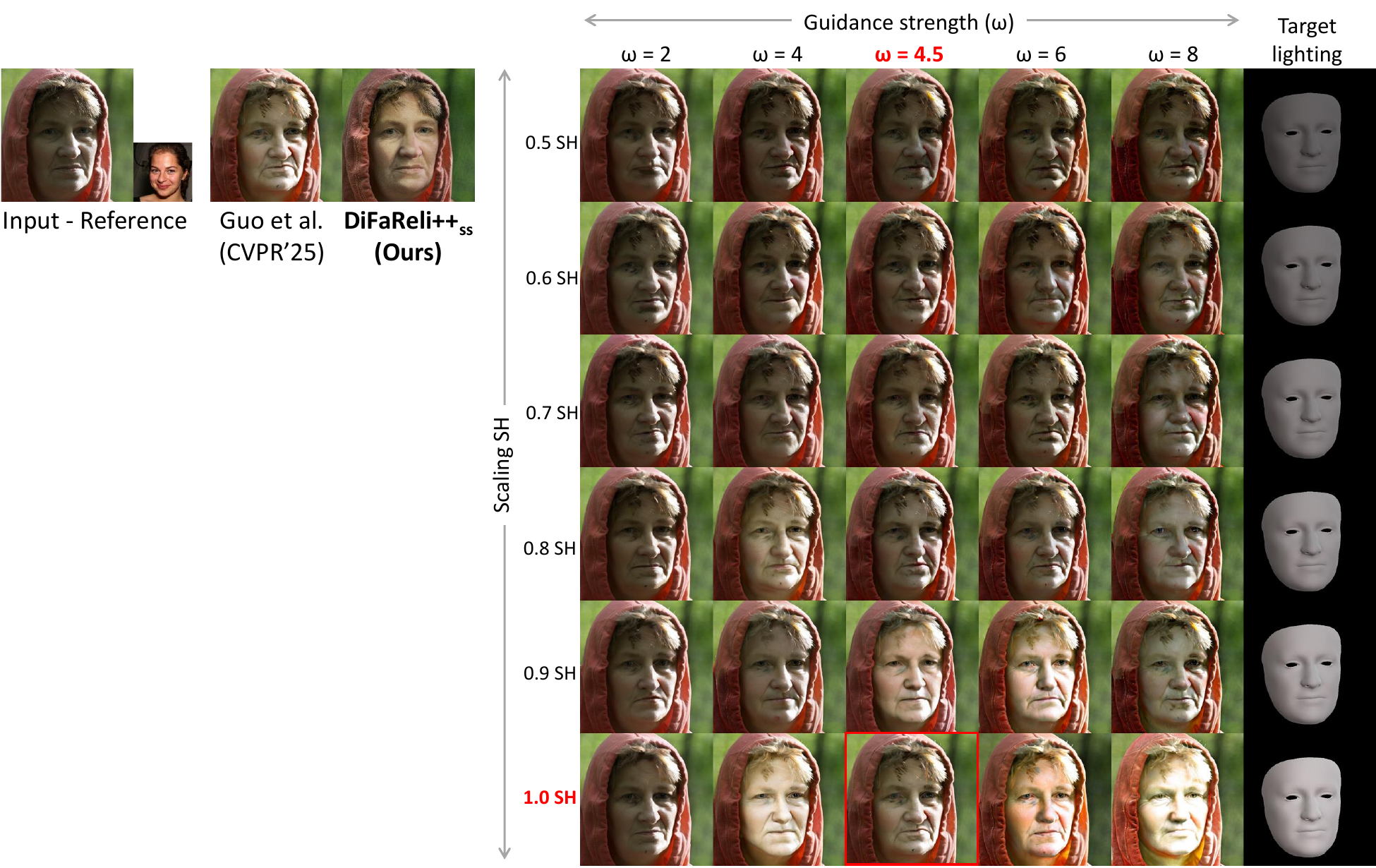}
  \caption{\textbf{Tuning the relighting hyperparameters of Guo et al.~\cite{Guo_2025_CVPR} (under a target lighting image).}
  We applied the same relighting hyperparameter tuning procedure as described in Figure~\ref{fig_app:tuninggrid_relipa_rot_sj1}. Results obtained with the default parameters reported in their paper is highlighted with a red box.
  }
  \label{fig_app:tuninggrid_relipa_targetSH_sj2}
\end{figure*}

\begin{figure*}[]
  \centering
  \vspace{-1.0cm}
  \includegraphics[width=0.85\textwidth]{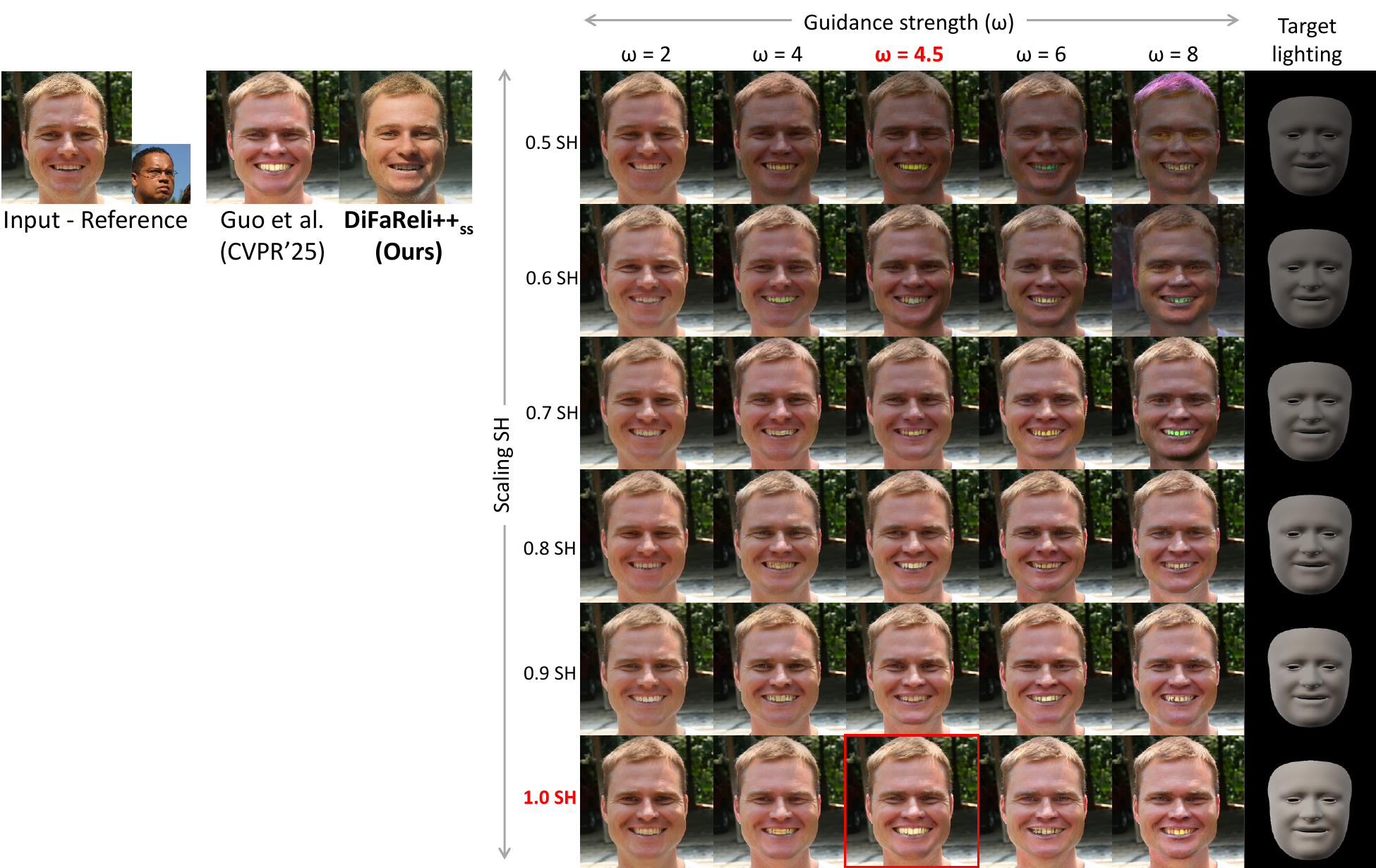}
  \caption{\textbf{Tuning the relighting hyperparameters of Guo et al.~\cite{Guo_2025_CVPR} (under a target lighting image).}
  We applied the same relighting hyperparameter tuning procedure as described in Figure~\ref{fig_app:tuninggrid_relipa_rot_sj1}. Results obtained with the default parameters reported in their paper is highlighted with a red box.
  }  
  \label{fig_app:tuninggrid_relipa_targetSH_sj3}
\end{figure*}

\begin{figure*}[]
  \centering
  \vspace{-1.0cm}
  \includegraphics[width=0.75\textwidth]{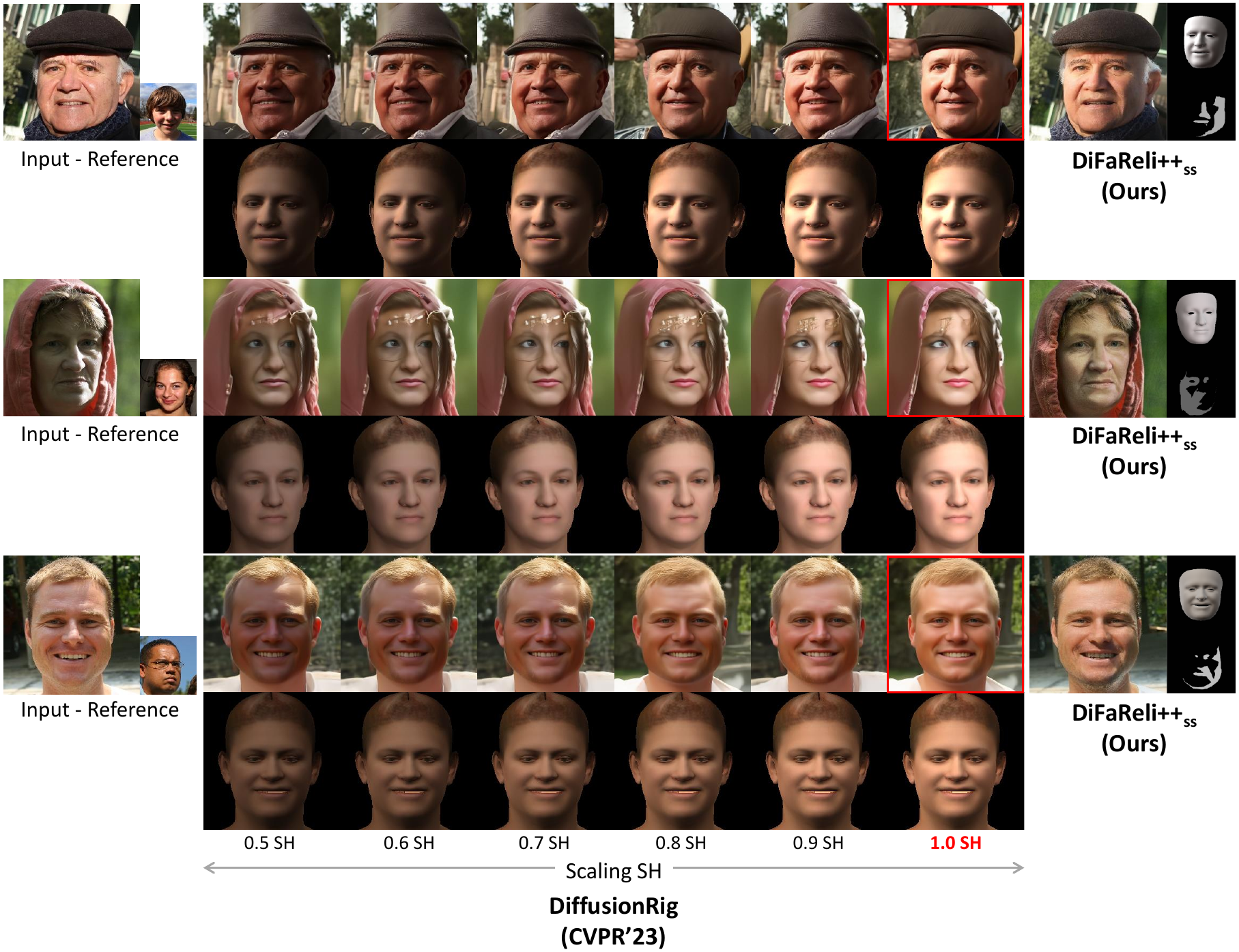}
  \caption{\textbf{Tuning the relighting hyperparameters of DiffusionRig~\cite{ding2023diffusionrig} (under a target lighting image).}
  We applied the same relighting hyperparameter tuning procedure as described in Figure~\ref{fig_app:tuninggrid_diffusionrig_rot_sj1}. Results obtained with the default parameters are highlighted with a red box.
  }
  \label{fig_app:tuninggrid_diffusionrig_targetSH}
\end{figure*}

\iffalse
\begin{figure*}[]
  \centering
  \vspace{-1.0cm}
  \includegraphics[scale=0.73]{./figures/ffhq_app_1.pdf}
  \caption{\textbf{Relit results on the FFHQ \cite{karras2019style}.}}
  \label{fig:ffhq_app1}
\end{figure*}

\begin{figure*}[]
  \centering
  \vspace{-1.0cm}
  \includegraphics[scale=0.73]{./figures/ffhq_app_2.pdf}
  \caption{\textbf{Relit images from FFHQ \cite{karras2019style}.}}
  \label{fig:ffhq_app2}
\end{figure*}
\fi

\begin{figure*}[]
  \centering
  \includegraphics[scale=0.69]{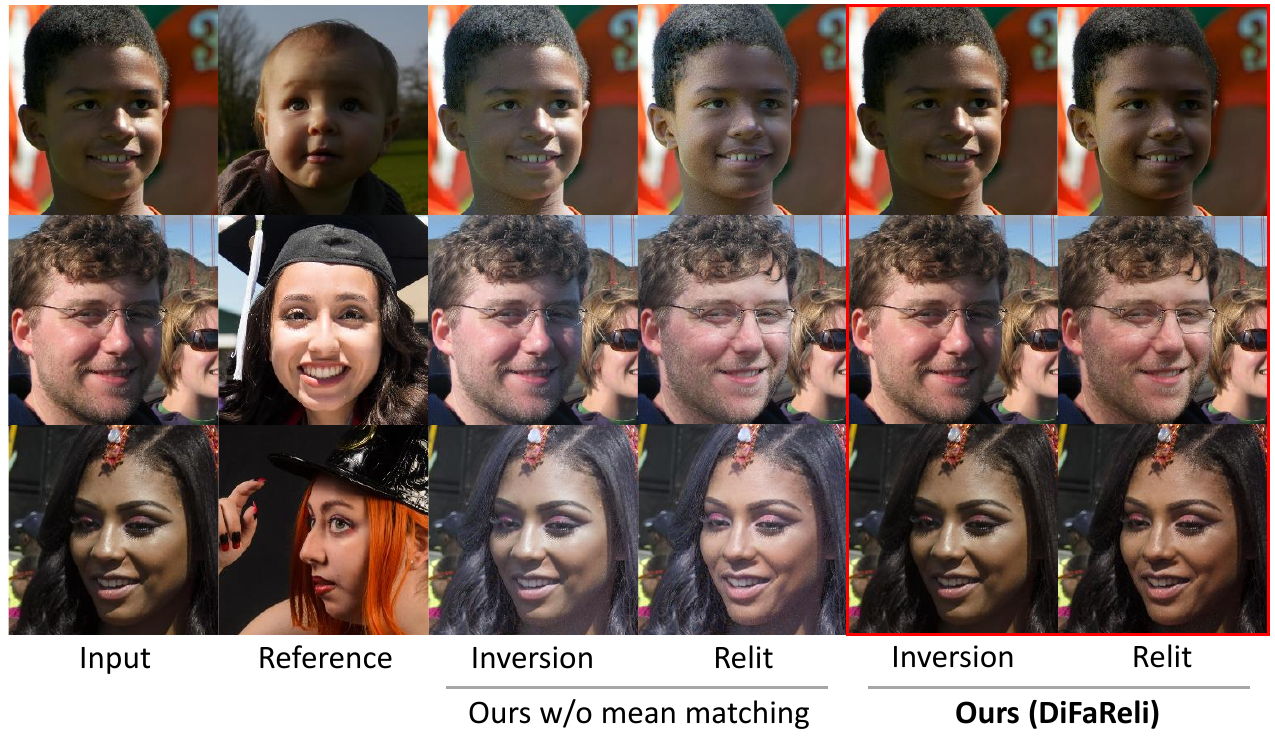}
  \caption{\textbf{Improved DDIM sampling with mean-matching.} We show a qualitative comparison between``with'' and ``without'' mean-matching. 
  Our mean-matching technique helps correct the overall brightness in both the inversion output and relit image.}
  \label{fig:mm}
\end{figure*}

\begin{figure*}[]
  \centering
  \includegraphics[scale=0.56]{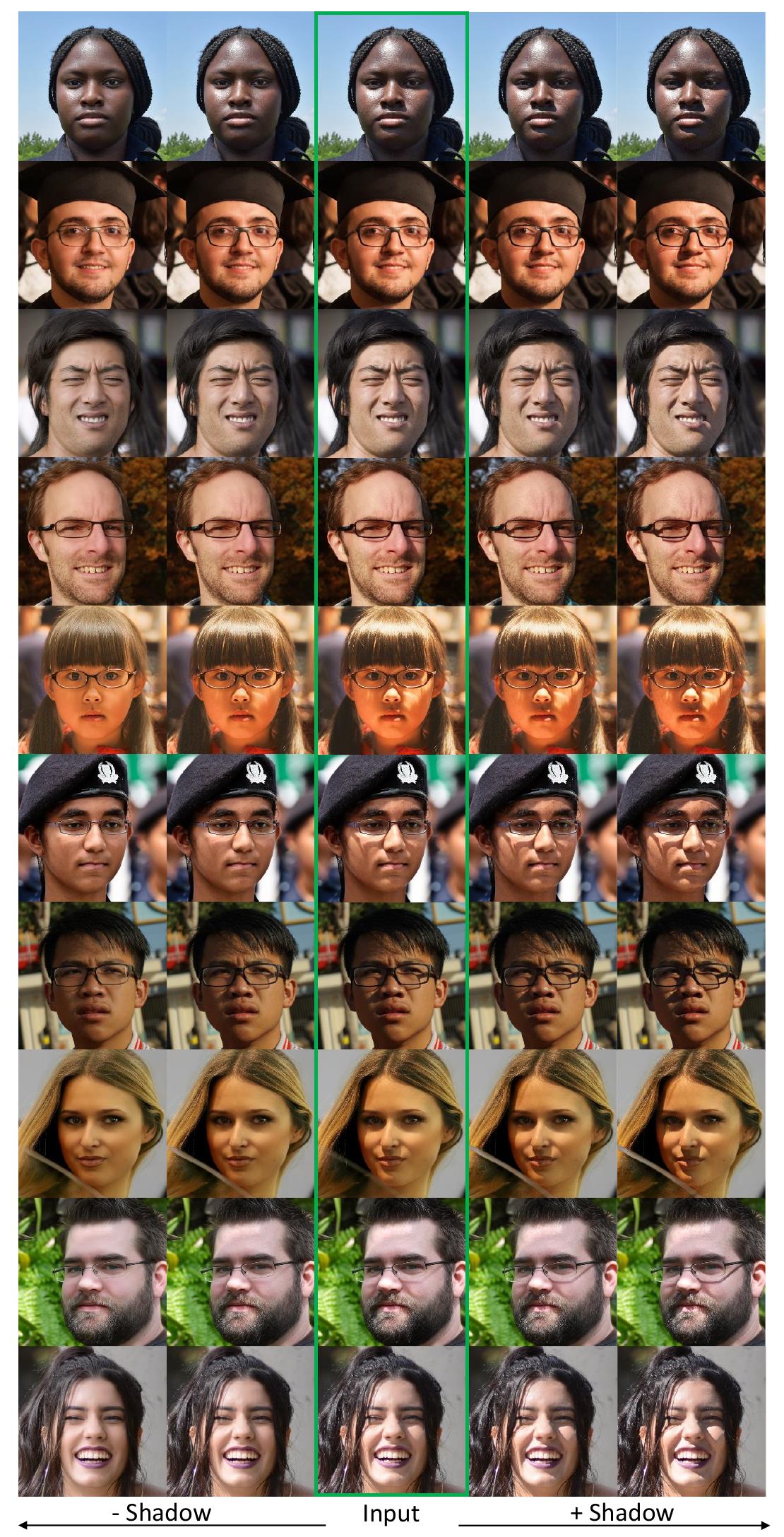}
  \caption{\textbf{Varying the intensities of cast shadows on FFHQ \cite{karras2019style}.}}
  \label{fig:shad}
\end{figure*}

\begin{figure*}[ht!]
\centering
  \includegraphics[scale=0.62]{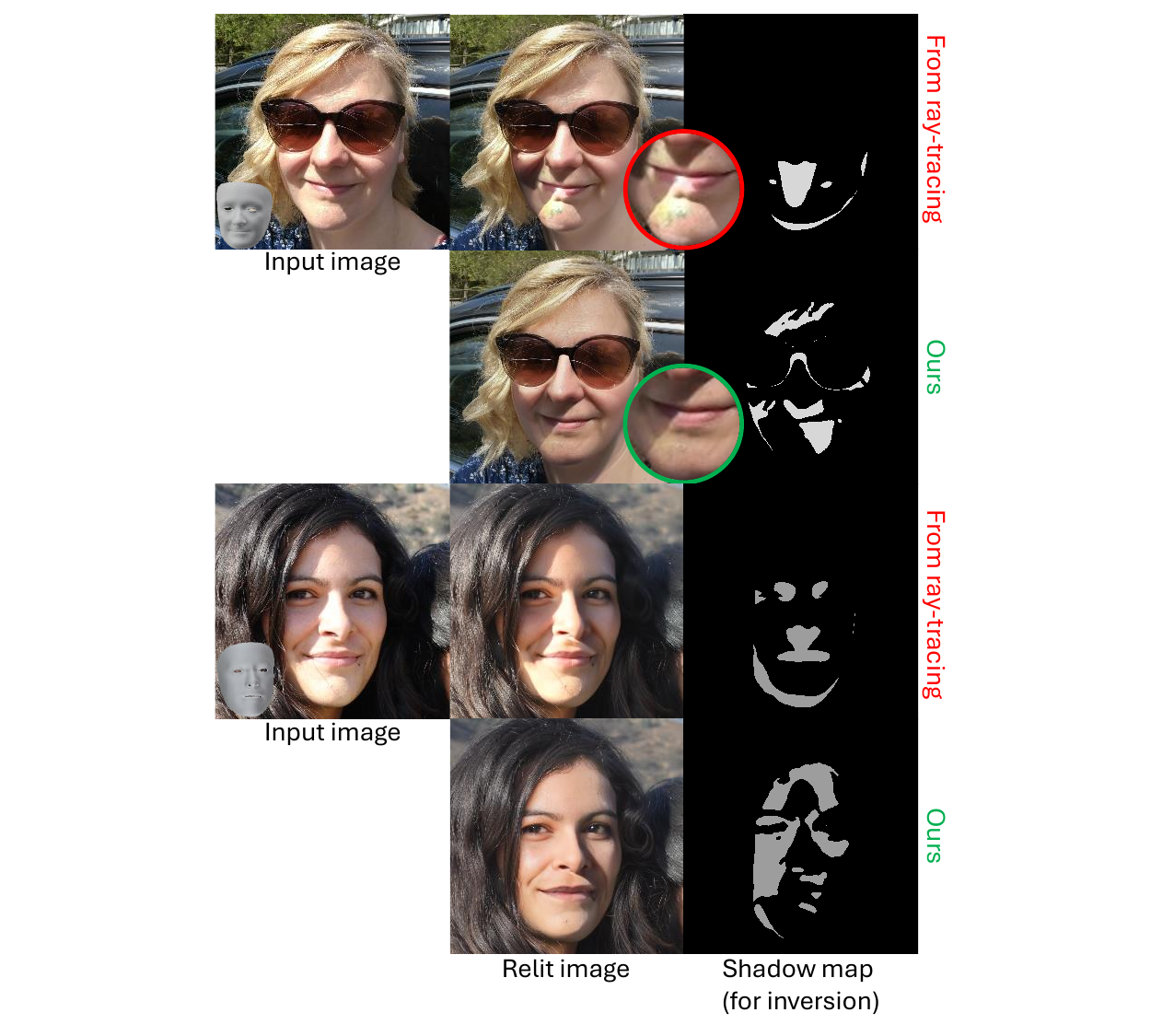}
  \caption{\textbf{Poor results from using ray-traced shadow maps for inversion.} Using ray-traced shadow maps for DDIM inversion, the top result shows that non-shadow areas are over-brightened (highlighted with a red circle), while the bottom result shows a failure to remove shadows and closely follow the conditioning shadow map.
  %(Top) Over-brightening artifacts (highlighted in circle) occur when using a ray-traced shadow map during inversion. 
  %This happens because the network attempts to brighten non-shadowed areas, resulting in the bright spot artifact.
%(Bottom) An additional result shows a case where the shadow cannot be removed, and the relit image does not closely follow the shadow map during relighting. This occurs because the ray-traced shadow map fails to provide useful information about the shadowed areas during inversion.
}
  \label{fig_app:overbright_results}
\end{figure*}

\begin{figure*}[ht!]
\centering
  \includegraphics[scale=0.52]{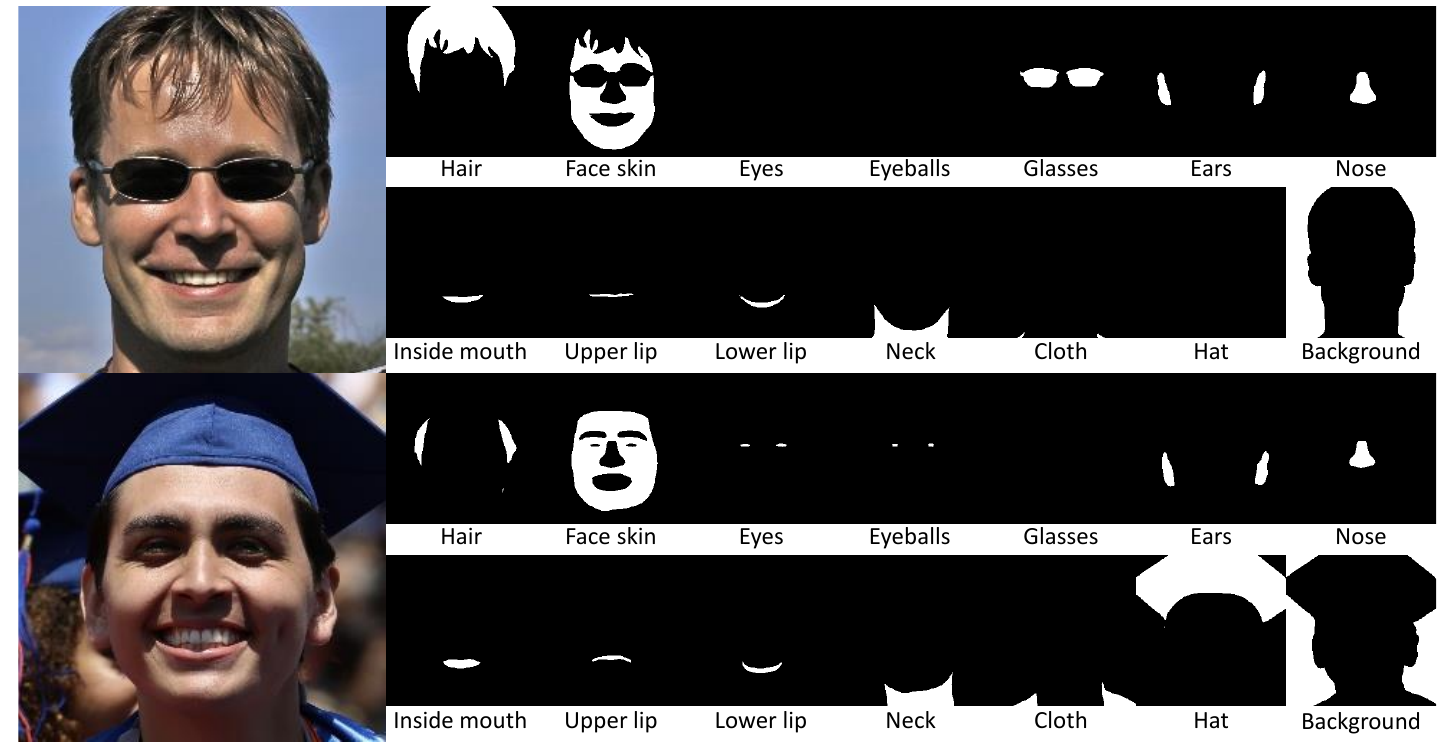}
  \caption{\textbf{All segmentation masks} used as conditioning inputs in DiFaReli++ (Section \ref{sec:relit_bg} in the main text).}
  \label{fig_app:seg_mask}
\end{figure*}

\begin{figure*}[ht!]
\centering
  \includegraphics[scale=0.55]{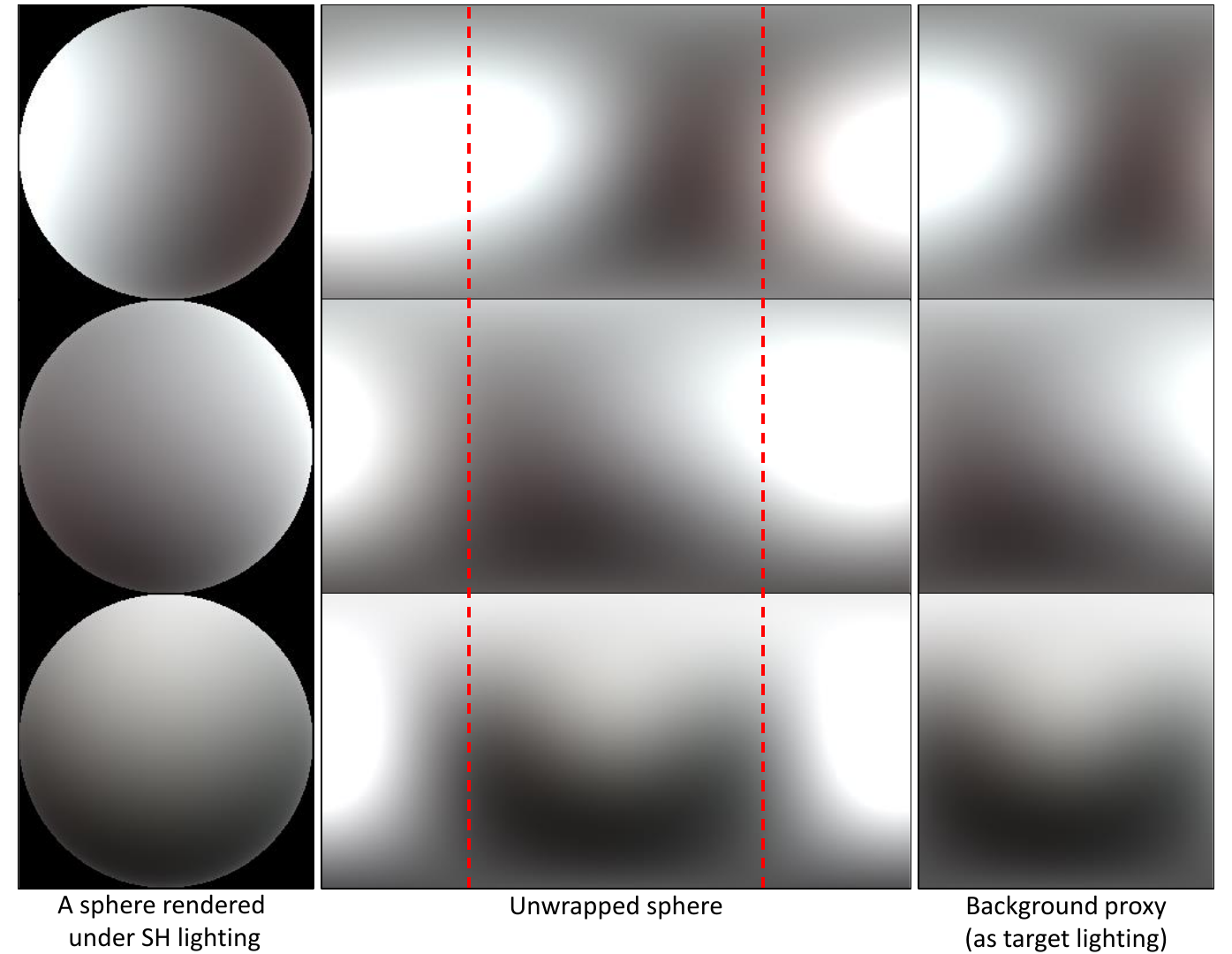}
  \caption{Examples of proxy background images that serve as target lighting for IC-light \cite{iclight}.}
  \label{fig_app:proxy_lighting}
\end{figure*}

\begin{figure*}[ht!]
\centering
  \includegraphics[scale=0.57]{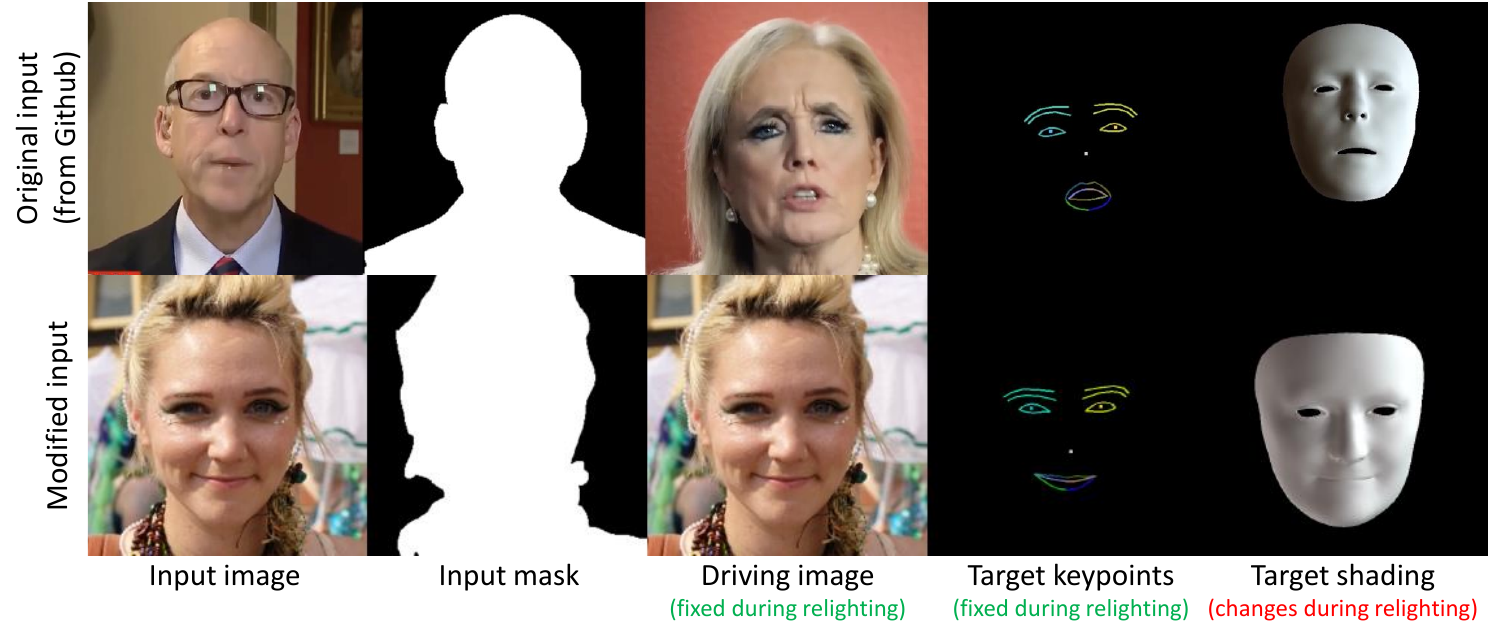}
    \caption{\textbf{Example of original and modified inputs for \cite{Guo_2025_CVPR}.}
    We show the original inputs from the official GitHub repository of \cite{Guo_2025_CVPR} (\url{https://github.com/MingtaoGuo/Relightable-Portrait-Animation}) and our modified versions adapted for the relighting task.  In our modification, we fixed the driving image and target keypoints and changed only the target shading for religthing, thereby limiting control to relighting without animating the face.}
  \label{fig_app:relipa_input}
\end{figure*}

\section{Potential negative societal impacts}
\label{app:negative}
Our method can be used for changing the lighting condition of an existing image and producing the so-called DeepFake, which can deceive human visual perception.
%Lighting is a key component that enhances the realism of a face image. It is possible to use our method for generating synthetic contents, such as DeepFake, that can deceive human visual perception. 
%Our technique is currently limited to changing the lighting condition. 
Our manipulation process is based on conditional DDIM \cite{dhariwal2021diffusion}, and a study from \cite{preechakul2022diffusion}, which uses the same architecture, shows that certain artifacts from DDIM can be currently detected using a CNN with about 92\% accuracy. We developed our work with the intention of promoting positive and creative uses, and we do not condone any misuse of our work.

\section{User interface for relighting user study}
\label{app:user_study_ui}
\begin{figure*}[]
  \centering
  \includegraphics[scale=0.7]{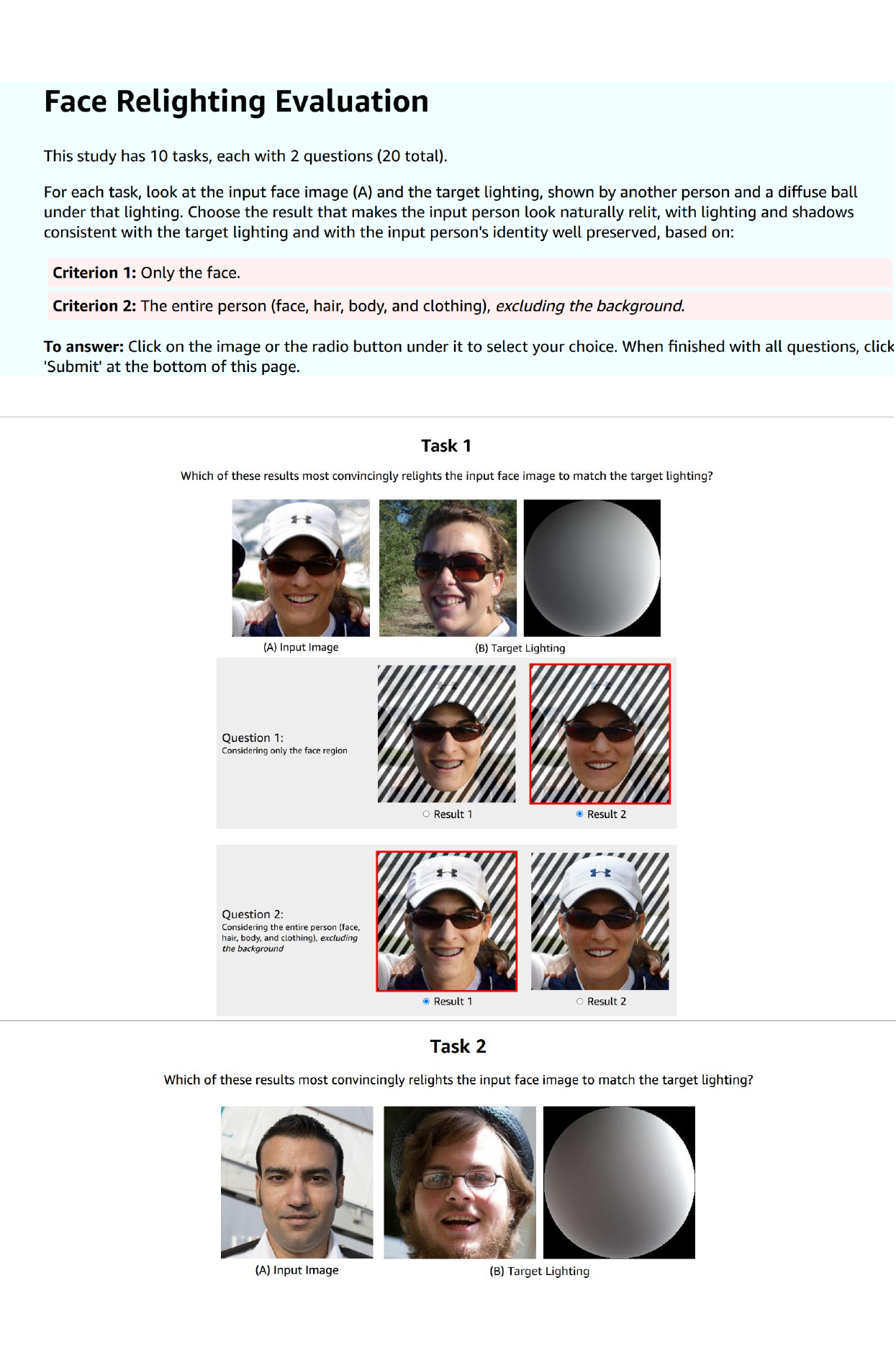}
  \caption{\textbf{User interface for the relighting user study of facial
and non-facial parts} (Section \ref{sec:user_study_targetSH} in the main text).}
  \label{fig_app:user_study_ui_image}
  % \vspace{-1.0em}
\end{figure*}

\begin{figure*}[]
  \centering
    \includegraphics[scale=0.7]{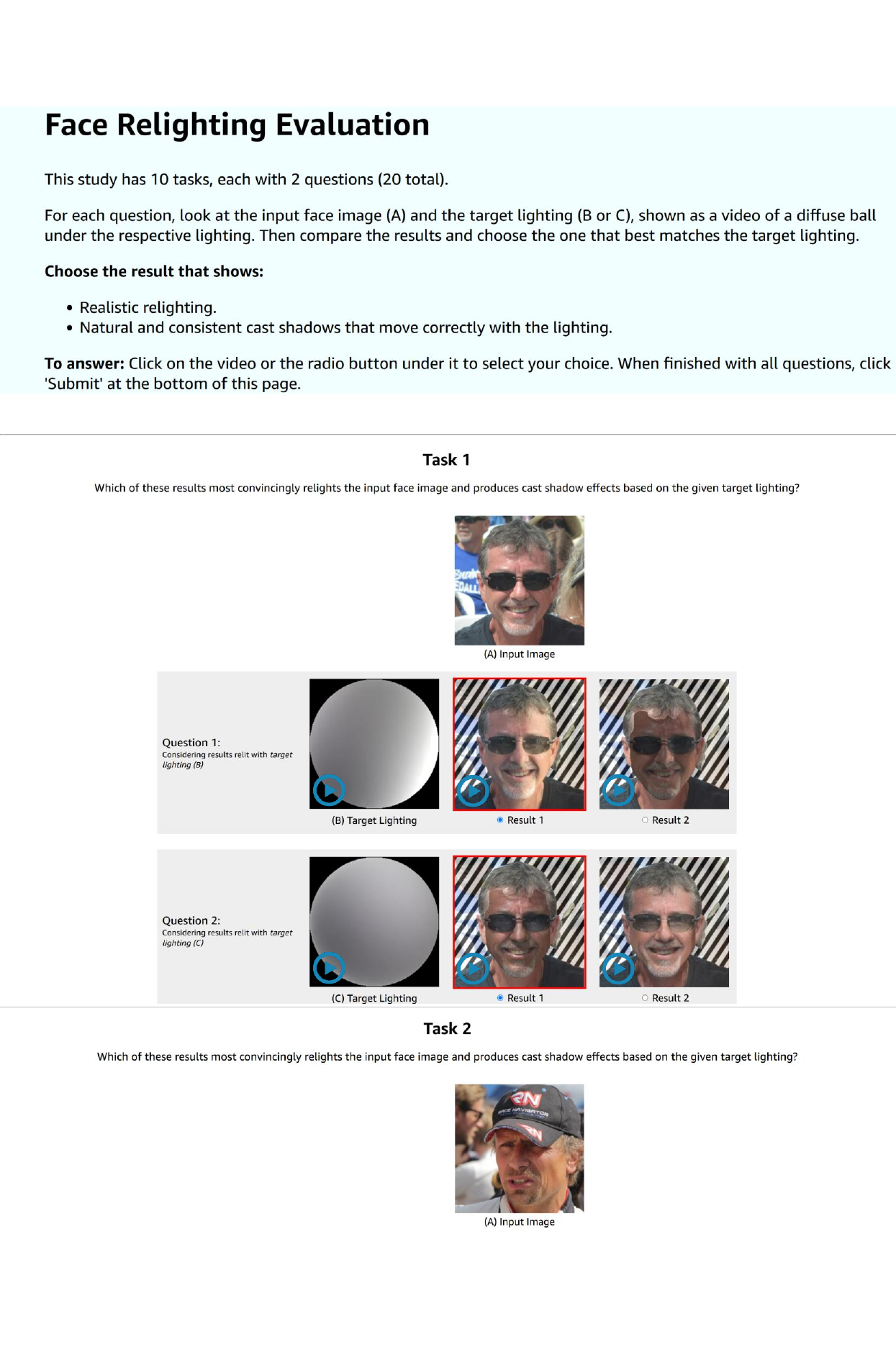}
  \caption{\textbf{User interface for the relighting user study on relighting quality under moving lights} (Section \ref{sec:user_study_cs} in the main text). In the interface, these results are videos that play simultaneously.
  %All videos of target lighting and relit results are shown to the user.
  }
  \label{fig_app:user_study_ui_video}
  % \vspace{-1.0em}
\end{figure*}

\begin{figure*}[]
  \centering
  \includegraphics[scale=0.7]{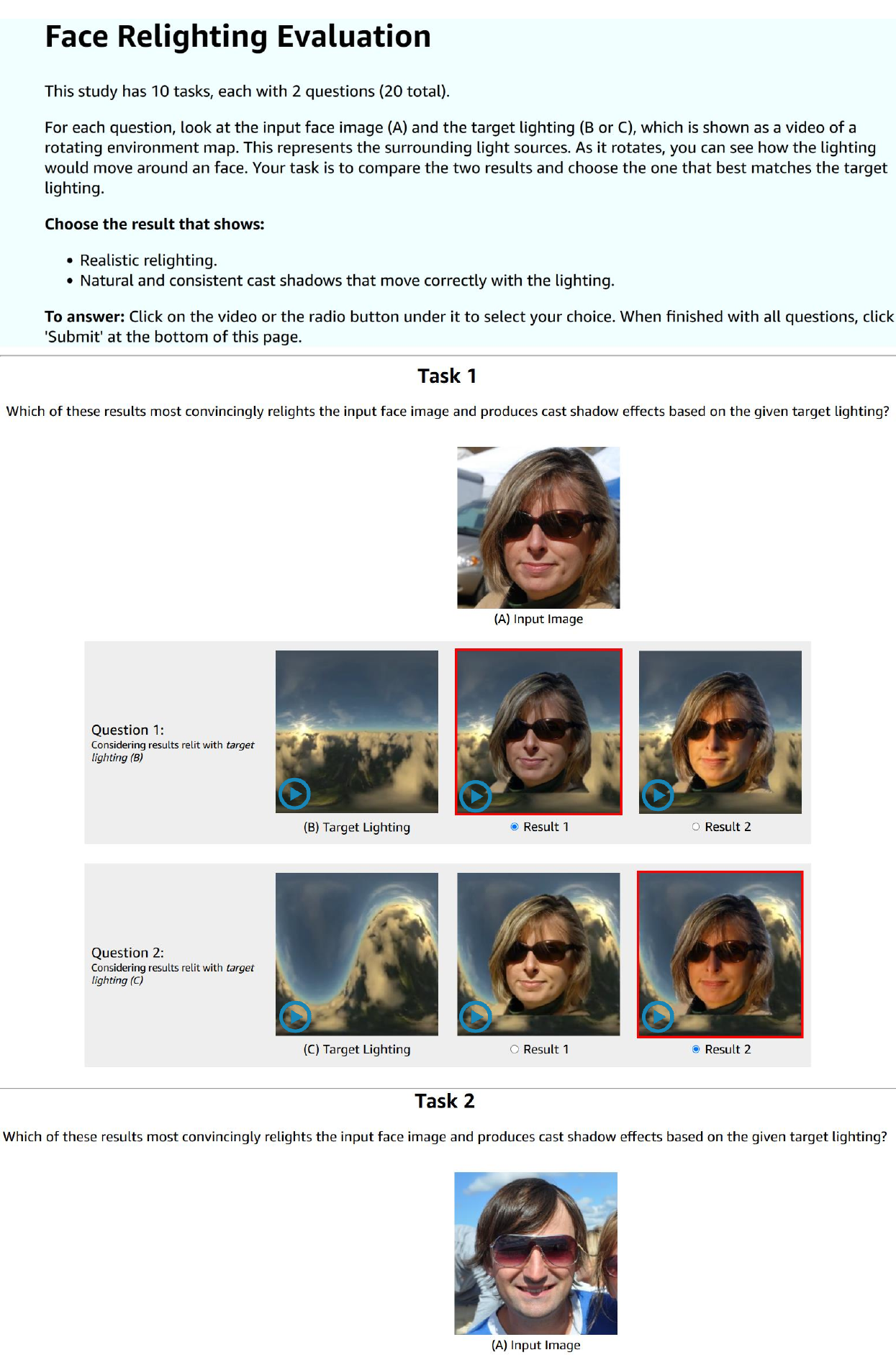}
  \caption{\textbf{User interface for the relighting user study on relighting quality under rotating HDR environment maps} (Section \ref{sec:user_study_cs_hdr} in the main text). In the interface, these results are videos that play simultaneously.
  %All videos of target lighting and relit results are shown to the user.
  }
  \label{fig_app:user_study_ui_hdr}
  % \vspace{-1.0em}
\end{figure*}

Figure \ref{fig_app:user_study_ui_image}, \ref{fig_app:user_study_ui_video}, and \ref{fig_app:user_study_ui_hdr} show the user interfaces used in our user studies, as detailed in Section \ref{sec:user_study} of the main paper. Each page contains 10 tasks, each with 2 questions (a total of 20 questions per page). 

For each task, an input image is displayed along with the target lighting condition, with irrelevant areas masked out to help users focus on either the face region or the entire person. The order of results for each task is shuffled when displayed to each participant. Instructions and criteria for making selections are provided at the top of the page.

%For each questions, each image has been generated using a different relighting method (including ours), and 
%The irrelevant areas have been masked out to help users focus on the face region or the entire person. We shuffle the order of the results for each task. 
%Users are asked to pick the one image that satisfies the criteria.
%The instructions and criteria for making selections are provided at the top of the page.
% }

\clearpage

\end{document}